\documentclass{article} 
\usepackage{collas2023_conference,times}
\usepackage{easyReview}
\usepackage{wrapfig}
\usepackage{xspace}
\usepackage{caption}
\usepackage{booktabs}
\usepackage{multirow}
\usepackage{makecell}
\usepackage{xcolor}
\definecolor{darkgreen}{RGB}{119,185,0}
\definecolor{DarkCoral}{rgb}{0.8, 0.36, 0.27}

\usepackage[utf8]{inputenc}
\usepackage{pgfplots}
\DeclareUnicodeCharacter{2212}{−}
\usepgfplotslibrary{groupplots,dateplot}
\usetikzlibrary{patterns,shapes.arrows}
\pgfplotsset{compat=newest}

\usepackage{amsmath,amsfonts,bm}

\newcommand{\base}{\mathtt{b}}
\newcommand{\data}{\mathcal{D}}

\newcommand{\task}{WSCI\xspace}

\newcommand{\ths}{\textsuperscript{th}\;}
\newcommand{\isup}{\textsuperscript{\textit{i}}\;}
\newcommand{\voc}{VOC\xspace}
\newcommand{\ours}{RaSP\xspace}
\newcommand{\wil}{WILSON\xspace}
\newcommand{\ctov}{COCO-to-VOC\xspace}

\newcommand{\vocfs}{PASCAL-5\isup\xspace}
\newcommand{\cocofs}{COCO-20\isup\xspace}

\newcommand{\up}{\textcolor{darkgreen}{$\uparrow$}}
\newcommand{\down}{\textcolor{red}{$\downarrow$}}

\def\eg{\emph{e.g.}}

\def\ie{\emph{i.e.}}

\def\wrt{w.r.t.\,}

\newcommand{\bfl}{{\bf l}}
\newcommand{\bfm}{{\bf m}}

\newcommand{\bfp}{{\bf p}}
\newcommand{\bfq}{{\bf q}}

\newcommand{\bfs}{{\bf s}}

\newcommand{\bfx}{{\bf x}}
\newcommand{\bfy}{{\bf y}}
\newcommand{\bfz}{{\bf z}}

\newcommand{\bfS}{{\bf S}}

\newcommand{\calC}{{\mathcal C}}

\newcommand{\calI}{{\mathcal I}}
\newcommand{\calL}{{\mathcal L}}

\newcommand{\calY}{{\mathcal Y}}

\def\eqref#1{equation~\ref{#1}}

\def\1{\bm{1}}

\DeclareMathAlphabet{\mathsfit}{\encodingdefault}{\sfdefault}{m}{sl}
\SetMathAlphabet{\mathsfit}{bold}{\encodingdefault}{\sfdefault}{bx}{n}

\newcommand{\R}{\mathbb{R}}

\newcommand{\softmax}{\mathrm{softmax}}

\DeclareMathOperator*{\argmax}{arg\,max}

\usepackage{hyperref}
\hypersetup{
    colorlinks=true,
    linkcolor=red,
    filecolor=magenta,
    urlcolor=blue,
    citecolor=purple,
    pdftitle={collas2023},
    pdfpagemode=FullScreen,
    }

\definecolor{brown}{RGB}{150, 75, 0}
\definecolor{oldcolor}{RGB}{0,100,0}
\definecolor{newcolor1}{RGB}{255, 0, 0}
\definecolor{newcolor2}{RGB}{203,96,21}
\definecolor{newcolor3}{RGB}{25, 25, 112}
\definecolor{newcolor4}{RGB}{221, 160, 221}
\definecolor{newcolor5}{RGB}{48, 213, 200}

\newcommand{\rebuttal}[1]{{\textcolor{black}{#1}}}
\newcommand{\oldclass}[1]{\textcolor{oldcolor}{#1}}
\newcommand{\newclassone}[1]{\textcolor{newcolor1}{#1}}
\newcommand{\newclasstwo}[1]{\textcolor{newcolor2}{#1}}
\newcommand{\newclassthree}[1]{\textcolor{newcolor3}{#1}}
\newcommand{\newclassfour}[1]{\textcolor{newcolor4}{#1}}
\newcommand{\newclassfive}[1]{\textcolor{newcolor5}{#1}}

\newcommand{\method}{RaSP\xspace}

\newcommand{\myparagraph}[1]{\noindent\textbf{#1}}

\usepackage[capitalise]{cleveref}

\Crefname{table}{Tab.}{Tabs.}
\Crefname{figure}{Fig.}{Figs.}
\Crefname{section}{Sec.}{Secs.}

\creflabelformat{equation}{#2\textup{(\textcolor{black}{#1})}#3}
\creflabelformat{table}{#2\textcolor{black}{#1}#3}
\creflabelformat{figure}{#2\textcolor{black}{#1}#3}
\creflabelformat{section}{#2\textcolor{black}{#1}#3}

\title{RaSP: Relation-aware Semantic Prior \\ for Weakly Supervised Incremental Segmentation}

\author{Subhankar Roy\\  
LTCI, Télécom-Paris \\Intitute Polytechnique de Paris\\
\texttt{subhankar.roy@telecom-paris.fr} \\
\And 
Riccardo Volpi, Gabrela Csurka, Diane Larlus  \\
NAVER LABS Europe \\
Meylan, France \\
\texttt{\{name.lastname\}@naverlabs.com} \\
}

\collasfinalcopy 

\begin{document}

\maketitle

\begin{abstract}

Class-incremental semantic image segmentation assumes multiple model updates, each enriching the model to segment new categories. This is typically carried out by providing
expensive pixel-level annotations to the training algorithm for all new objects, limiting the adoption of such methods in practical applications. Approaches that solely require image-level labels offer an attractive alternative, yet, such coarse annotations lack precise information about the location and boundary of the new objects. In this paper we argue that, since classes represent not just indices but semantic entities, the conceptual relationships between them can provide valuable information that should be leveraged. 
We propose a weakly supervised approach that exploits such semantic relations to transfer objectness prior from the previously learned classes into the new ones, complementing the supervisory signal from image-level labels. We validate our approach on a number of continual learning tasks, and show how even a simple pairwise interaction between classes can significantly improve the segmentation mask quality of both old and new classes. We show these conclusions still hold for longer and, hence, more realistic sequences of tasks and for a challenging few-shot scenario.

\end{abstract}
\section{Introduction}
\label{sec:introduction}

When working towards the real-world deployment of artificial intelligence systems, two main challenges arise: such systems should possess the ability to continuously learn, and this learning process should only require limited human intervention. 
While deep learning models have proved
effective in tackling tasks for which large amounts of curated data as well as abundant computational resources are available,
they still struggle to
learn over continuous and potentially heterogeneous sequences of tasks, especially if supervision is limited. 

\begin{wrapfigure}[17]{R}{0.55\linewidth}
    \vspace{-14pt}
    \begin{center}
        \includegraphics[width=\linewidth]{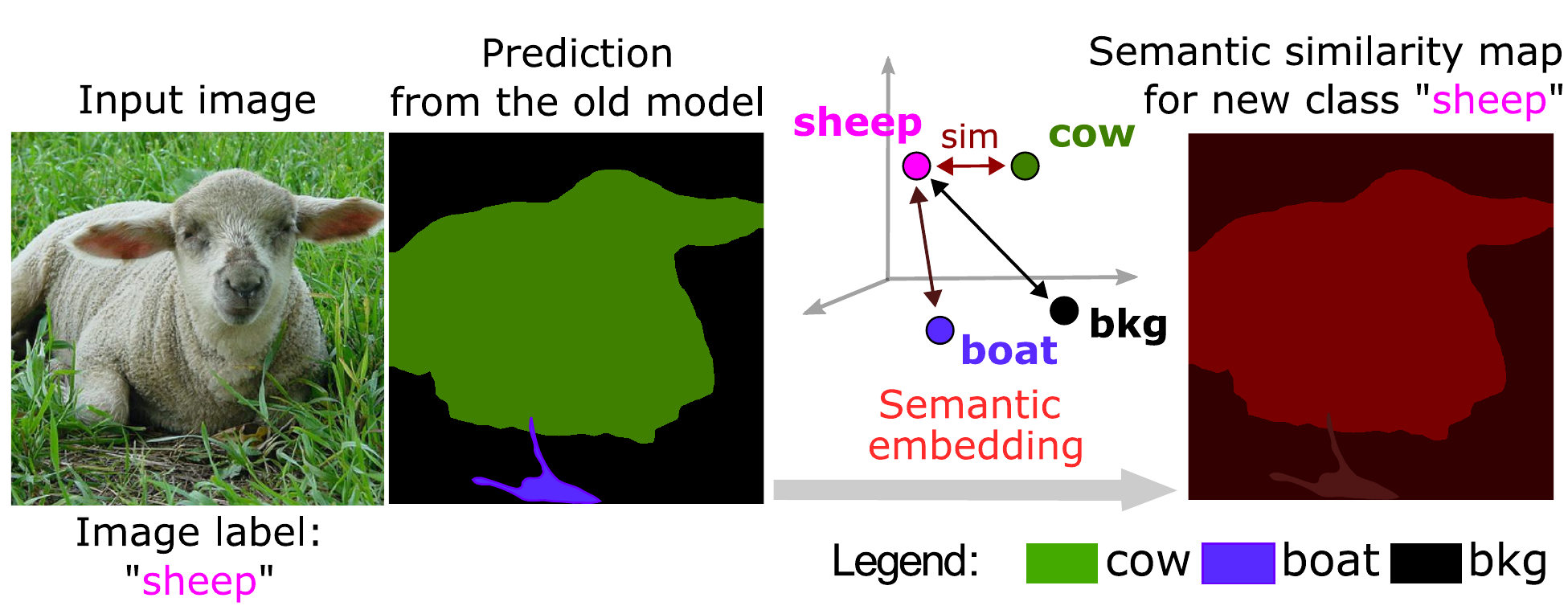}
    \end{center}
    \vspace{-8pt}
     \caption{Our proposed Relation-aware Semantic Prior (\method) loss is based on the intuition that old class predictions from an existing model provide valuable cues for the segmentation of unseen, semantically related classes. Based on the semantic relatedness of the image label (\eg, sheep) and the model predictions (\eg, cow), our model derives denser maps for the new class and leverages them during training}
    \label{fig:eye_catching}
\end{wrapfigure}

In this work, we focus on the task of semantic image segmentation (SIS)~\citep{csurka2022semantic}, where the goal is predicting the class label of each pixel in an image.
A reliable and versatile SIS model should be able to seamlessly 
add new categories to its repertoire without forgetting about the old ones.
Considering for instance a house robot or a self-driving vehicle with such segmentation capability, we would like it to extend its knowledge to new classes without having to retrain the segmentation model from scratch on the old ones. 
Such ability is at the core of continual learning research, the main challenge being to mitigate catastrophic forgetting of what has been previously learned~\citep{Parisi2019ContinualLifelongLearningWithNNAReview}.

Most learning algorithms for SIS assume training samples with dense pixel-level annotations, an expensive and tedious operation.
We argue that this is cumbersome and severely hinders continual learning;
adding new classes over time should be an annotation friendly process.
This is why, here, we focus on the case where only \textit{image-level} labels are 
provided 
(\eg, adding the `sheep' class comes as easily as only providing images guaranteed to contain at least a 
sheep).
However, this  task, denoted as Weakly Supervised Class-Incremental (WSCI) SIS,
is an extremely challenging problem in itself and very few attempts have been made to tackle it in the context of continual learning~\citep{CermelliCVPR22IncrementalLearninginSemSegmfromImageLabels}.

We argue that weakly supervised SIS, despite being a harder problem due to the lack of dense supervision, can efficiently be addressed in an \textit{incremental} scenario,
by exploiting the model's own prior about the objects that have already been learned in the past. This stems from our observation that a SIS model trained to segment, \eg,~the class `cow' often misclassifies the pixels of an unseen class `sheep'
and assign those to the
`cow' class that it knows (see Fig.~\ref{fig:eye_catching}, left). This
is caused by the
visual similarity between the \textit{bovines} (\eg, cow) and \textit{ovines} (\eg, sheep), both being furry four-legged species. In this work, we take advantage of this behavior 
by equipping the model with the ability to take into account the \textit{semantic relationship} between the old and new classes, while localizing new objects -- as humans do. In other words, the localization cues offered by the model on semantically similar objects (\eg, cow and sheep) is
used to approximately convert the coarse image-level supervision of the new class into dense pixel-level supervision (thanks to the semantic similarity maps, as shown in Fig.~\ref{fig:eye_catching}, right). Learning a new class then simply becomes optimizing a pixel-level supervised objective that is erstwhile not available in weakly supervised SIS. Given the similarity maps are derived by leveraging the semantic relationship between the class label names, we term the proposed objective as \textbf{R}elation-\textbf{a}ware \textbf{S}emantic \textbf{P}rior (\method) loss.

The \method~loss has been designed with the goal of improving \textit{forward transfer} in incremental learning scenarios by converting weaker image-level supervision into pixel-level.
It can be seen as a general-purpose plug-and-play module, suitable for any weakly supervised class-incremental SIS framework, as all
it needs is a segmentation network that outputs pixel-level predictions and the class label \textit{names} of the previously seen classes. In our experiments we show that \method---when integrated with the state-of-the-art method WILSON~\citep{CermelliCVPR22IncrementalLearninginSemSegmfromImageLabels}---leads to performance improvements, sometimes by large margins, and especially in longer incremental scenarios.

To summarize, our contributions are threefold: (\textbf{i}) 
We propose the \method loss to facilitate class-incremental SIS when only image-level labels are available as supervision.
It treats class labels as semantic entities and exploits what the model knows about previous classes it has been trained on, to learn new ones at each increment; 
(\textbf{ii}) We broaden the benchmarks previously used for weakly supervised class-incremental SIS and consider longer sequences of tasks (prior work is limited to 2, we extend to up to 11 tasks) and few-shot incremental settings,
in both cases with image-level annotations only; (\textbf{iii}) We empirically validate that the steady improvement brought by \method is also visible in an extended version of our approach that uses an episodic memory, filled with either past samples or web-crawled images for the old classes.
We show that, in this context, the memory does not only mitigate catastrophic forgetting, but also and most importantly fosters the learning of new categories.
\section{Related Work}
\label{sec:related}

This work lies at the intersection of
weakly supervised
and class-incremental learning of SIS models. Due to the nature of our semantic prior loss, it also relates to 
text-guided computer vision.

\myparagraph{Weakly supervised SIS.} 

To circumvent the need for expensive pixel-level annotations when learning SIS models, weakly supervised SIS~\citep{BorensteinECCV04LearningToSegment} approaches 
training SIS models using
cheaper and
lesser constrained forms of annotations such as image captions~\citep{XuCVPR22GroupViTSiSEmergesFromTextSupervision}, bounding boxes~\citep{DaiICCV15BoxSupExploitingBoundingBoxesSuperviseSiS,JiBMVC21WeaklySupSiSBox2TagAndBack,SongCVPR19BoxDrivenClassWiseRegionMaskingFilling}, scribbles~\citep{LinCVPR16ScribbleSupScribbleSupConvNetsForSemSegm,TangCVPR18NormalizedCutLossForWeaklySupCNNSemSegm}, points~\citep{BearmanECCV16WhatsThePointSemSegmWithPointSup,QianAAAI19WeaklySupSceneParsingWithPointBasedDistanceMetricLearning} and image labels~\citep{KolesnikovECCV16SeedExpandConstrainWeaklySupervisedImgSegm,AhnCVPR18LearningPixelLevelSemanticAffinityWSSS,AraslanovCVPR20SingleStageSemSegmFromImageLabels,xu2022multi}. 
Out of these, learning to segment with only image labels is the most attractive alternative, as the annotation cost is arguably the lowest. Our work falls under the family of methods using image-level supervision, but jointly uses the ground truth image-level labels and the predictions of the old model to provide denser supervision to the new classes. Opposed to several of the previous works, our \method loss is simple by design and can be integrated with any SIS model.

\myparagraph{Class-incremental SIS.} Under the 
hood of continual learning~\citep{Parisi2019ContinualLifelongLearningWithNNAReview}, 
class-incremental learning 
consists in exposing a model to sequences of tasks, in which
the goal is learning new classes without having access to data from the previous classes.
While most class-incremental learning methods have  
focused on image classification (see~\cite{masana2020class} for a survey), 
some recent works have started focusing on
SIS~\citep{CermelliCVPR20ModelingBackgroundIncrementalLearningSemSegm,MichieliCVIU21KnowledgeDistillationIncrementalLearningSemSegm, DouillardCVPR21PLOPLearningWithoutForgettingContinualSemSegm,MaracaniICCV21RECALLReplayBasedContinualLearningSemSegm,ChaNeurIPS21SSULSemSegmwithUnknownLabelforExamplarClassIncLearning}. 
Yet, all aforementioned methods assume pixel-level annotations for all the new classes, 
which requires a huge, often prohibitively expensive amount of manual work.
Therefore, weakly-supervised class-incremental SIS has emerged as a viable alternative 
in the pioneering work of \cite{CermelliCVPR22IncrementalLearninginSemSegmfromImageLabels}, which formalizes the WSCI task, and proposes the WILSON method 
to tackle it. 
In details, the WILSON framework builds on top of standard weakly supervised SIS techniques~\citep{AraslanovCVPR20SingleStageSemSegmFromImageLabels}, and explicitly tries to mitigate forgetting using knowledge distillation, akin to the pseudo-labeling approach of PLOP~\citep{DouillardCVPR21PLOPLearningWithoutForgettingContinualSemSegm}. Orthogonal to the components introduced in WILSON that mostly deal with fortifying backward transfer, our \method improves the forward transfer aspect of the WSCI task by providing better supervision to the weakly supervised localizer for segmenting new classes.

\myparagraph{Language-guided computer vision.}
Vision and language have a long history 
of benefiting from each other, and language, a modality that is inherently more semantic, has often been used as a source of supervision to guide computer vision tasks, such as learning visual representations~\citep{quattoni2007learning,gomez2017self,sariyildiz2020icmlm,radford2021learning}
object detection~\citep{ShiICCV17WeaklySupObjectLocalizationUsingThingsAndStuffTransfer},
zero-shot segmentation~\citep{ZhouCVPR16LearningDeepFeaturesForDiscLoc,BucherNIPS19ZeroShotSemanticSegmentation,XianCVPR19SemanticProjectionNetworkZeroAndFewLabelSS,LiNIPS20LConsistentStructuralRelationLearning4ZeroShotSiS,BaekICCV21ExploitingJointEmbeddingSpace4GeneralizedZeroShotSiS},
language-driven segmentation~\citep{ZhaoICCV17OpenVocabularySceneParsing,LiICLR22LanguageDrivenSemSegm,GhiasiECCV22ScalingOpenVocabularySiSImageLevelLabels,XuCVPR22GroupViTSiSEmergesFromTextSupervision} or referring image segmentation~\citep{HuECCV16SegmentationFromNaturalLanguageExpressions,LiuICCV17RecurrentMultimodalInteraction4ReferringImageSegmentation,DingCVPR21VisionLanguageTransformerQueryGenerationReferringSegmentation,WangCVPR22CRISCLIPDrivenReferringImageSegmentation}, among others. 
One of the core ingredients behind the success of language and vision models is the ability to embed the natural language (\eg, captions, class names, etc.) into semantically meaningful spaces using Word2Vec~\citep{mikolov2013efficient}, GloVe~\citep{pennington2014glove} or BERT~\citep{DevlinNAACL19BERT}---to name a few.
Similarly, 
our \method loss assumes the availability of such similarity metrics to be used between textual pairs consisting of the name of the predicted class label and the name of the ground truth image label, where the strength of the similarity map is determined by the semantic closeness. 
Contrary to the open-vocabulary segmentation methods that use large-scale datasets with descriptive image captions~\citep{XuCVPR22GroupViTSiSEmergesFromTextSupervision}, at times alongside ground truth pixel-level annotations~\citep{WangCVPR22CRISCLIPDrivenReferringImageSegmentation} or class-agnostic segmentation annotations~\citep{GhiasiECCV22ScalingOpenVocabularySiSImageLevelLabels}, our \method just requires class label names and predictions of the model itself.

\section{Methods}
\label{sec:method}

We 
develop
a method for the task of Weakly Supervised Class-Incremental SIS (\task), where the goal is incrementally learning to segment objects from new classes by
using
image-level labels only, and thus avoiding the need for pixel-level annotations. 
Before 
detailing
our method, we formalize our setting.

\myparagraph{Problem setup and notations.} Following the \task setting established 
by~\citet{CermelliCVPR22IncrementalLearninginSemSegmfromImageLabels} to evaluate WILSON, 
we likewise assume access to pixel-level annotations for an initial set of categories, followed by incrementally learning on a sequence of new classes using image-level labels only. This can be regarded
as well-aligned with practical scenarios
for which dense annotations are available for entry-level \textit{primitive} classes, whereas the less frequently occurring objects or specialized variants of the generic classes incrementally come with image labels only.

Let $\data^\base = \{(\bfx^\base_k, \bfy^\base_k)\}^{N^\base}_{k=1}$ be a dataset for SIS, where $\bfx^\base \in \R^{H \times W \times 3}$ represents an input image and $\bfy^\base$ is a tensor 
containing
the $|\calC^\base|$-dimensional one-hot label
vectors 
for each pixel,
in a $H \times W$ spatial grid, corresponding to a set of $\calC^\base$ semantic classes.
As typical in SIS,
objects that do not belong to any of the foreground classes
are annotated as a special background class (`\textit{bkg}')---included in $\calC^\base$.
We refer to $\data^\base$ as the \textit{base} task and do not make assumptions on its cardinality.
$\data^\base$ is used to train a base model,
generally defined by an encoder $E^{\base}$ and a decoder $F^{\base}$, $(E^{\base} \circ F^{\base}) \colon \bfx \to \R^{|\calI| \times |\calC^\base|}$,
where ${|\calI|}=H' \times W'$ is a spatial grid---corresponding to the input image size or some resized version of it---and $\bfp = (E^{\base} \circ F^{\base})(\bfx)$ 
is the set of   class prediction maps,  where  $p_i^c$ is the  probability 
for the spatial location $i \in {\calI} $ in the input image $\bfx$ to belong to the class $c$.

After this base training, we assume the model 
undergoes a sequence of learning steps, as training sets for new tasks become available.
Specifically, at each learning step $t$, 
the model is exposed to 
a new set
$\data^{t} = \{(\bfx^{t}_k, \bfl^{t}_k)\}^{N^{t}}_{k=1}$ containing $N^{t}$ instances 
labeled for previously unseen $\calC^{t}$ classes, where $\bfl^{t} \in \R^{|\calC^{t}|}$ 
is the vectorized \textit{image-level} label corresponding to an image $\bfx^{t}$.
Note that in each incremental step, only weak annotations (image-level labels) are provided for the new classes.
This is in sharp contrast with the base task, in which the model is trained with 
pixel-level annotations.

The goal of \task is to update the segmentation model at each incremental step $t$ in a weakly supervised way, without \textit{forgetting} any of the previously learned classes.
We learn the function
$(E^t \circ F^t) \colon \bfx \to \R^{|\calI| \times |\calY^t|}$,
where 
$\calY^t = \bigcup^{t}_{k=1} \{\calC^{k}\} \cup \calC^\base$ 
is the set of labels at step $t$ (old and new ones).
Note that, in general, we assume that data from previous tasks cannot be stored---that is, there is no episodic memory. We relax this assumption for some of our experiments: see Sec.~\ref{sec:rehearsal} for results related to the setting that includes an episodic memory.

\subsection{The Relation-aware Semantic Prior Loss}
\label{sec:contrib}

\begin{figure*}[t!]
    \centering
    \includegraphics[width=\textwidth]{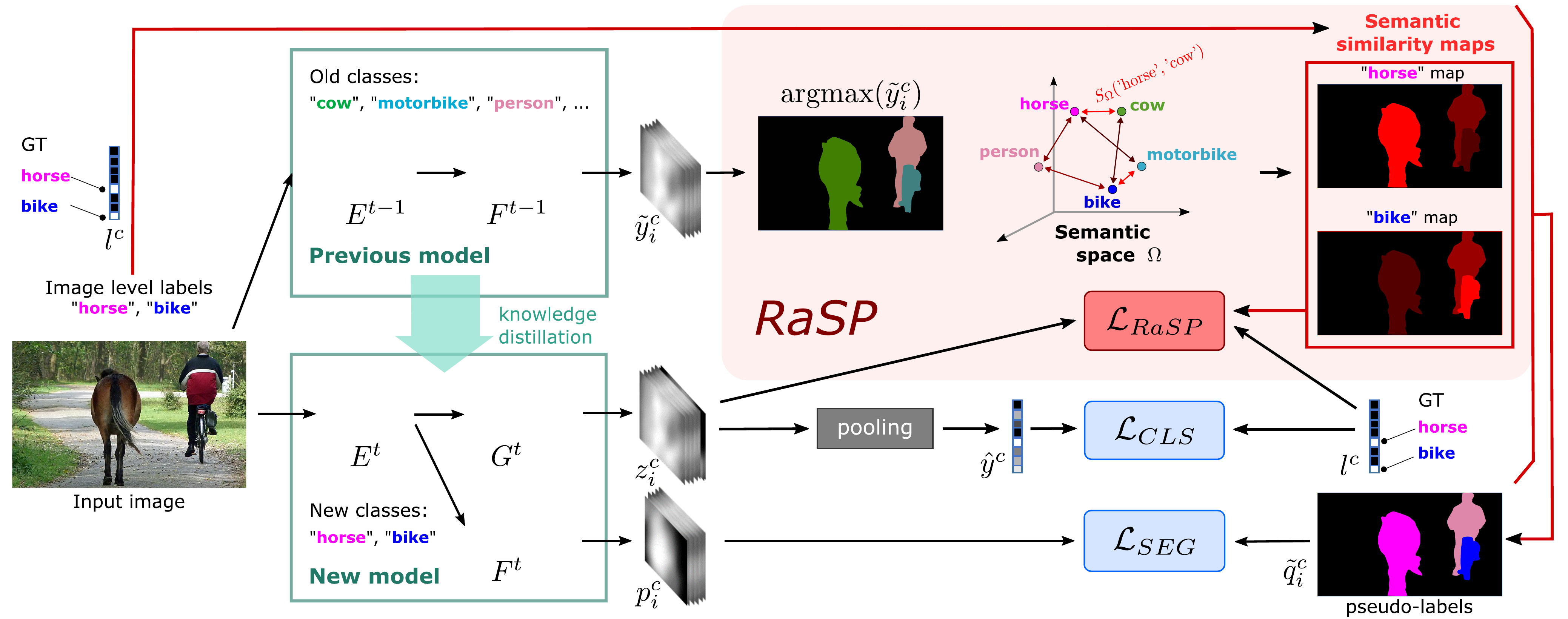}
    \caption{
    Overview of the \textbf{\method} loss integrated in a generic WSCI framework. Given the snapshot of a segmentation model $(E \circ F)^{t-1}$, trained to segment `cow', `motorbike', and `person', the training step $t$ is tasked with learning new classes `horse' and `bike' using image-level labels only. The \method loss uses the old model predictions $\tilde{y}^c_i$ and image labels $l^c$ to generate semantic maps, where the intensity of the semantic map at pixel location $i$ is proportional to the semantic distance between the embeddings of the two class names. These dense semantic maps are then used as pseudo supervision to train the localizer $(E \circ G)^{t}$. $\mathcal{L}_\text{\method}$ can be seamlessly combined with
    any generic weakly supervised $\mathcal{L}_\text{CLS}$ and not-forgetting losses}
    \label{fig:RaspModel}
\end{figure*}

\begin{figure}[!t]

    \centering
    \setlength{\tabcolsep}{1.7pt}
    \begin{tabular}{cccccc}
    
     \small Input & \small GT & \small $(F\circ E)^{t-1}(\bfx_t)$   & \small $\bf{s}^{\bf{l}_t}$ &  \small \ours   & \small \wil \\
        \includegraphics[width=0.15\columnwidth]{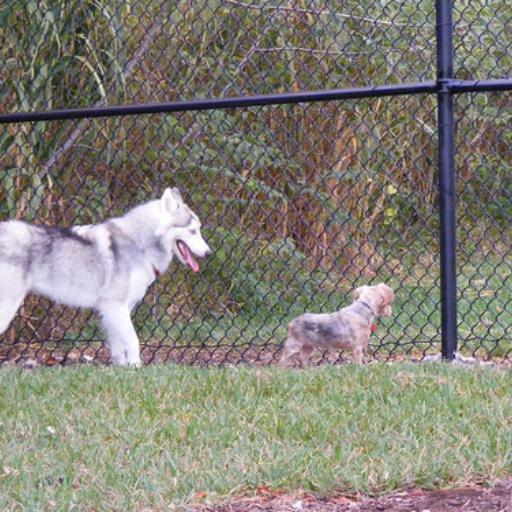} &
       \includegraphics[width=0.15\columnwidth]{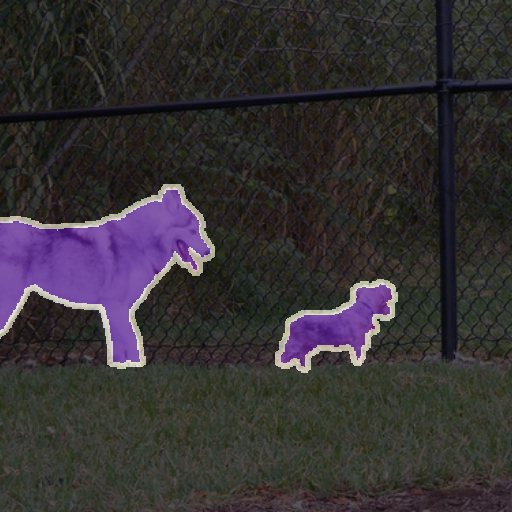} & \includegraphics[width=0.15\columnwidth]{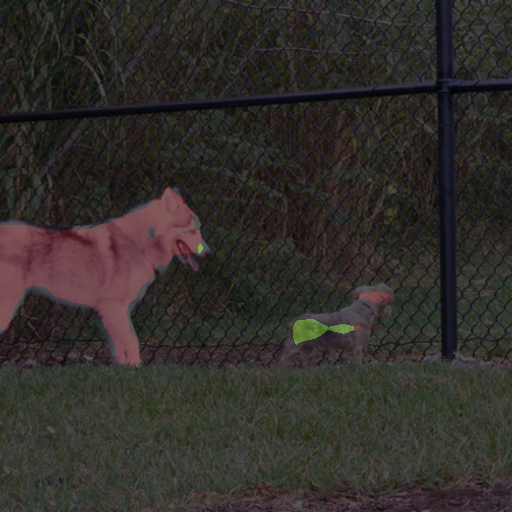} & \includegraphics[width=0.15\columnwidth]{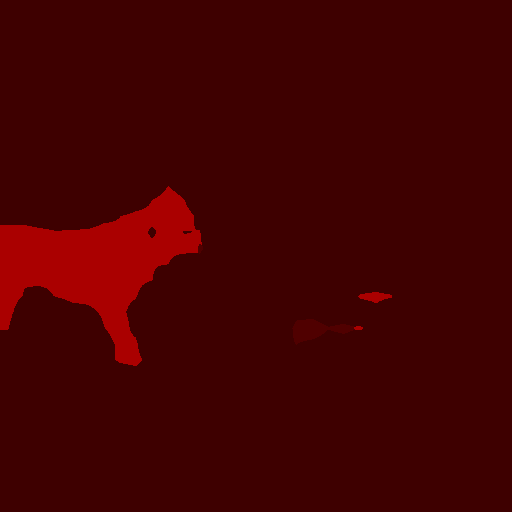} & \includegraphics[width=0.15\columnwidth]{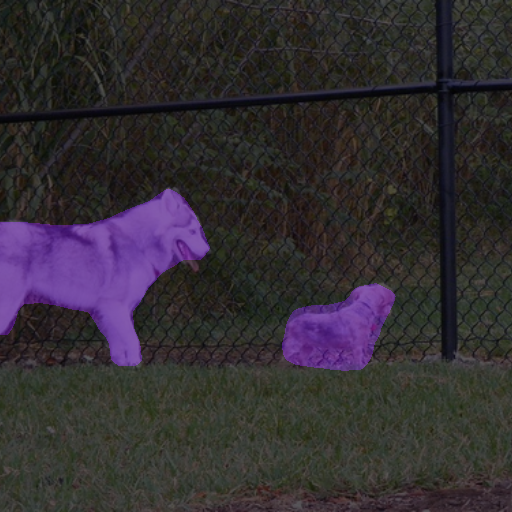} & \includegraphics[width=0.15\columnwidth]{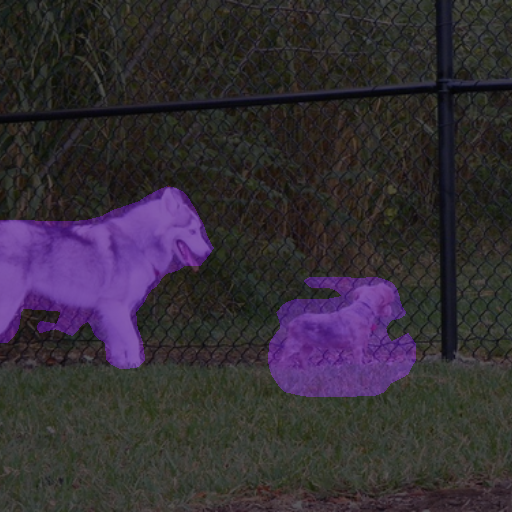} \\
       \
        \includegraphics[width=0.15\columnwidth]{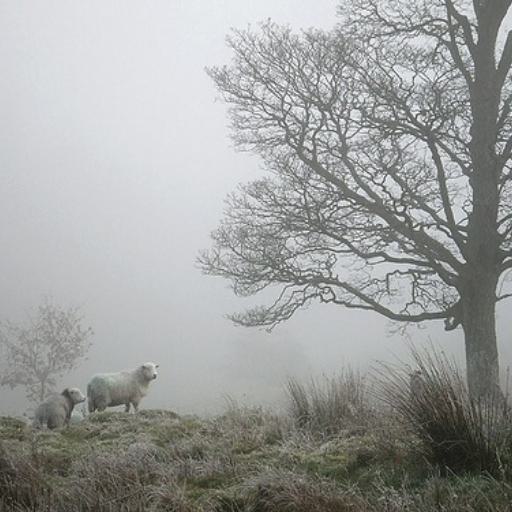} &
       \includegraphics[width=0.15\columnwidth]{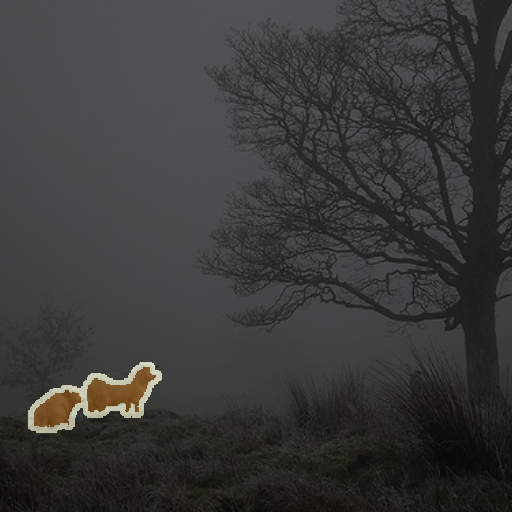} & \includegraphics[width=0.15\columnwidth]{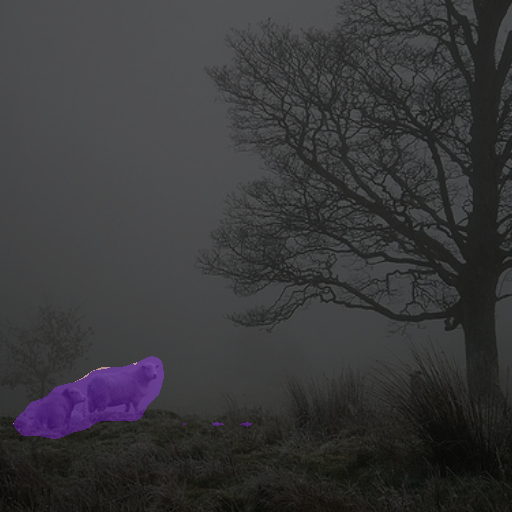} & \includegraphics[width=0.15\columnwidth]{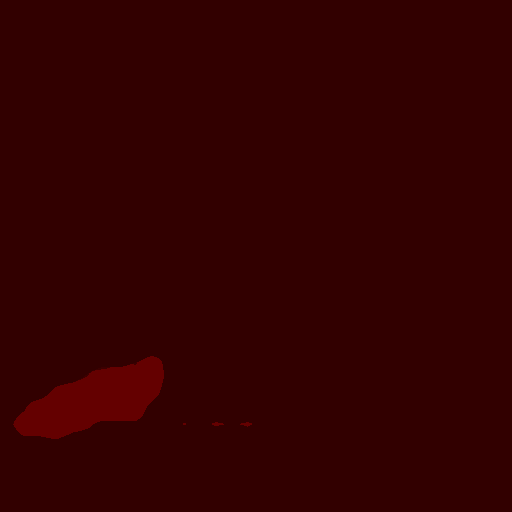} & \includegraphics[width=0.15\columnwidth]{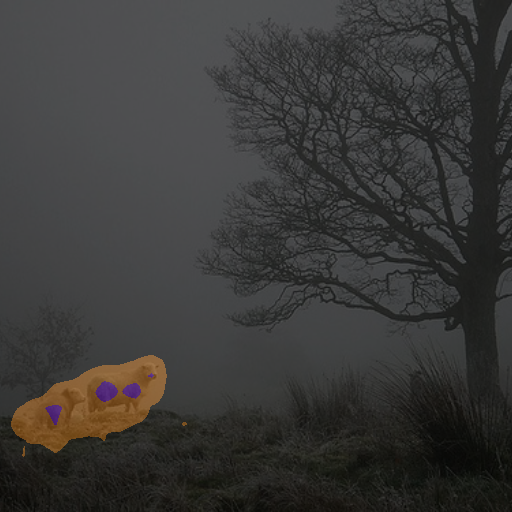} & \includegraphics[width=0.15\columnwidth]{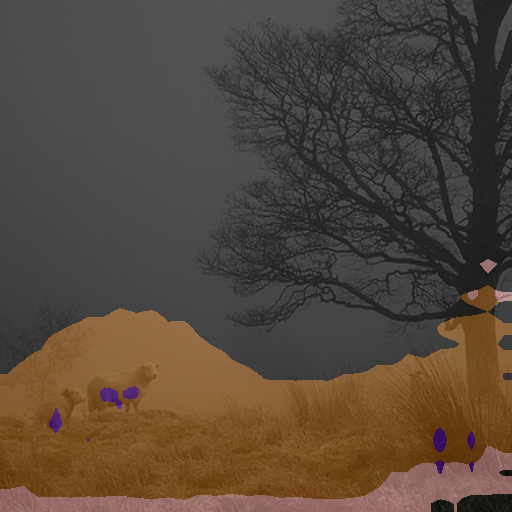} \\
    \end{tabular}
    \captionof{figure}{\textbf{Visualizations.} Qualitative figures from  the \textit{multi-step} \textbf{overlap} incremental protocol  on 10-2 VOC. From left to right: 
    input image, 
    GT  segmentation overlayed, 
    predicted segmentation from old model, 
    semantic similarity map corresponding to the image label (
    {\color{violet}dog} / \textcolor{brown}{sheep}) computed between this label and  old classes, 
    predicted segmentation  obtained with \ours and with \wil. 
    Semantic similarity maps displayed in OpenCV  colormap HOT
    (low~\includegraphics[width=36pt,height=7pt]{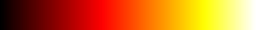}~high similarity)}
    
    \label{fig:qualitative_visualization}
\end{figure}

In this paper, we propose to leverage the semantic relationship between the new and old classes to improve the segmentation results when only image labels are available.
We argue that semantic object categories are not independent, \ie, the new classes $\calC^t$ that are being learned at step $t$ may bear semantic resemblance with the old classes from  $\calY^{t-1}$, seen by the model during previous training steps. 
Going back to our initial example, the network may have been trained to segment instances of the `cow' class
with dense supervision during the base training, and at any arbitrary incremental step $t$ the segmentation network can be tasked with learning to segment the `sheep' class 
from weak-supervision. Since cow and sheep
are closely related species sharing similar attributes (such as being four-legged, 
furry mammals), 
the old snapshot of the model $E^{t-1} \circ F^{t-1}$ (or, for brevity, $(E \circ F)^{t-1}$) can provide valuable cues to localize the `sheep'
regions in an image labeled as sheep,
despite having never seen this animal before
(see Fig.~\ref{fig:eye_catching}). 
Guided by this insight, 
instead of using the old model predictions to solely 
obtain cues about 
the old classes (if present), as done \eg, in WILSON, we propose a semantically-guided 
prior
that uses the old model predictions to discover more precise object boundaries for the new classes, in the form of semantic similarity maps.
Note that the class-incremental SIS methods are often based on the popular background-shift~\citep{CermelliCVPR20ModelingBackgroundIncrementalLearningSemSegm} assumption that unseen objects are \textit{always} classified as background by an old model. Our prior loss challenges this assumption, and is based on our observation that the old model tends to misclassify foreground regions from unseen objects as closely related old classes. We believe that both phenomenona are prevalent in incremental learning, and we root our method on the latter. We qualitatively validate our motivation through extensive visualizations in Fig.~\ref{fig:qualitative_visualization}.

Concretely, at step $t$ and using the old model $(E \circ F)^{t-1}$, for each pixel $\bfx_i^t$  we assign 
the most probable class label $y_i^* = \argmax_{c \in \calY^{t-1}}  \tilde{y}_i^c$ from old classes, yielding the label map $\bfy^*$. Note that our method expects $y_i^*$ to be a class label \textit{name} instead of a class \textit{index} (\eg, say `cow' instead of an index 5 for the class cow). 
Then,
 given the set of ground truth image-level label names $\calL(\bfx^t)=\{c| \bfl^t_c=1 \}$ associated with image $\bfx^t$, we estimate a similarity map $\bfs^c$ between each class $l^c$ in $\calL(\bfx^t)$ and the predicted label map  $\bfy^*$: 
\begin{equation}
\label{eqn:semantic-similarity}
    \bfs^c = \{\bfS_\Omega\left(\omega(y^*_i), \omega(l^c)\right)\}_{i \in \calI},
\end{equation}
where $\omega(c)$ is a vectorial embedding  of 
the semantic class $c$ in a semantic embedding
space $\Omega$ and $\bfS_\Omega$ is a semantic similarity measure defined between the classes in 
$\Omega$ (see~\cref{sec:app-sem-sim} for details). 
Different semantic embeddings 
can
be considered,
such as Word2Vec~\citep{mikolov2013efficient}, GloVe~\citep{pennington2014glove} or 
BERT~\citep{DevlinNAACL19BERT}.
These language models were trained such that 
the dot product between a pair of embedding vectors, $\bfS_\Omega$,
reflects the semantic similarity between their corresponding text. 
For example in Fig.~\ref{fig:eye_catching}, $\bfS_\Omega(\omega(\textrm{`sheep'}),\omega(\textrm{`cow'})) \gg \bfS_\Omega(\omega({`\textrm{sheep}'}),\omega(\textrm{`bkg'}))$, as `sheep' lies closer to `cow' in the semantic space than the `background' class. Intuitively, stronger the similarity between the predicted class label $y^*_i$ at pixel location $i$ and the ground truth image label $l^c$, higher 
the likelihood of the pixel $i$ belonging to the new class $l^c$. 
In this work, we use BERT~\citep{DevlinNAACL19BERT} for all the experiments (see comparisons with other embeddings in the Tab.~\ref{tab:sem-sim-type}).

Note that the similarity maps in~\cref{eqn:semantic-similarity} are computed exhaustively for every pixel location in a given image with respect to all the previous classes in $\calY^{t-1}$, which also includes the \textit{bkg} class. As the background can not reliably provide objectness cues for new object classes, we ensure not to alter the original predictions made on the background class by normalizing the similarity map such that the score for the `\textit{bkg}' class is equal to 1: 
\begin{equation}
\label{eqn:binary-tree-sim}
    s^c_i =
    \frac{\exp(S_\Omega(\omega(y^*_i), \omega(l^c)) / \tau)}{\exp(S_\Omega(\omega(\text{`}bkg\text{'}), \omega(l^c)) / \tau)},  
\end{equation}
where $\tau$ is a scaling hyperparameter. 
By exploiting the similarity maps we convert the image labels $l^c$ into pixel-level label maps $\bfs^c$, one per new class $c$ (see Fig.~\ref{fig:eye_catching} and Fig.~\ref{fig:RaspModel}). We provide more details in the Appendix (Sec.~\ref{sec:app-normalization}).

The generality of our proposed \method loss is evident from the fact that the similarity maps $\bfs^c$ are derived using the same segmentation model from the 
previous
step $(E \circ F)^{t-1}$ and the image-level labels of the image $\bfx^t$ from the current step.
While in many cases the dense similarity maps offered by \method might be sufficient
(when all the new classes in $\calY^{t}$ have strong resemblance to the old classes $\calY^{t-1}$), they can fall short in situations where completely unrelated classes appear in $\calY^{t}$ or the new class region is predicted as background.

To enable learning in all possible scenarios, including the \textit{edge-cases} where \textit{all} the new classes are dissimilar to the previous ones, we couple our method with approaches from the weakly supervised SIS literature. The most common weakly supervised SIS approach is to exploit the classification activation maps (CAM)~\citep{ZhouCVPR16LearningDeepFeaturesForDiscLoc} from a standard classifier (dubbed as \textit{localizer} head $G^t$), trained for predicting image-level class labels, to obtain the most discriminative regions as pixel-level pseudo-labels~\citep{KolesnikovECCV16SeedExpandConstrainWeaklySupervisedImgSegm,AraslanovCVPR20SingleStageSemSegmFromImageLabels}. 
In our formulation, where we
learn with weaker image labels, we bootstrap the training of the aforementioned localizer~\citep{AraslanovCVPR20SingleStageSemSegmFromImageLabels} using the proposed semantic similarity maps in the form of the following binary cross-entropy (BCE) loss:
\begin{equation}
\label{eqn:semantic-loss}
    \calL_{\text{\method}} (\bfz, \bfs) = - \frac{1}{|\calC^t||\calI|}\displaystyle \sum_{i \in \calI} \displaystyle \sum_{c \in \calC^{t}} \sigma(s^c_i) \log (\sigma(z^c_i)) + (1 - \sigma(s^c_i)) \log (1 - \sigma(z^c_i)),
\end{equation}
where $z_i^c$ is the logit corresponding to the class $c$ assigned by the localizer at location $i$,  $\sigma(\cdot)$ is the sigmoid function.
Given a generic loss for a localizer $\calL_{\text{CLS}}$ (an instance of this loss will be detailed in the next section), we can combine the two terms as $\calL = \calL_{\text{CLS}} + \lambda \calL_{\text{\method}}$. Intuitively, our proposed loss serves as a regularizer that encourages forward transfer from the old classes to the new ones. 

\subsection{Full integration of \method}
\label{sec:wilson}

Without loss of generality, we implement our \method loss on top of the WILSON framework~\citep{CermelliCVPR22IncrementalLearninginSemSegmfromImageLabels}. 
We chose WILSON since it is the state of the art and, since it relies on the localizer module introduced by~\citet{AraslanovCVPR20SingleStageSemSegmFromImageLabels} to tackle \task, represents a good fit to test our loss.

\myparagraph{Background.}
WILSON is an end-to-end method for \task that incrementally learns 
to segment new classes
with the supervision of 
pseudo-labels
generated by 
the localizer
trained with image-level supervision.
More specifically, at step $t$,
WILSON is composed of a
shared encoder $E^t$, a main segmentation head $F^t$---which is incrementally extended to accommodate new classes---and a localizer head $G^t$, trained from scratch for every task. It also stores a copy of the model from the previous task, $(E \circ F)^{t-1}$.

Given an image $\bfx$ from the current task,
$\tilde{\bfy} = \sigma( (F \circ E)^{t-1}(\bfx)) \in \R^{|\calI| \times |\calY^{(t-1)}|}$
is the output produced by the old model.
The scores obtained by the localizer, 
 $\bfz = (G \circ E)^t(\bfx) \in \R^{|\calI| \times |\calY^t|}$, 
are aggregated into a one-dimensional vector $\hat{\bfy}  \in \R^{|\calY^{t}|}$ by using normalized Global Weighted Pooling (see~\cref{sec:app-ngwp} for details).
The score $\hat{y}_c$, for each class $c$, can be seen as the likelihood for image $\bfx$ to contain semantic class $c$. This allows training the model with 
image-level labels using the multi-label soft-margin loss: 
\begin{equation}
\label{eqn:cam-loss}
    \calL_{\text{CLS}} (\hat{\bfy}, \bfl) = - \frac{1}{|\calC^t|}\displaystyle \sum_{c \in \calC^t} l^c\log (\sigma(\hat{y}^c)) + \displaystyle \sum_{c \in \calC^t} (1 - l^c)\log (1 - \sigma(\hat{y}^c)).
\end{equation}

Note that, although the localizer outputs a $|\calY^t|$-dimensional vector,
at task $t$ we are only provided with images and their image-level annotations for the new classes. 
Therefore, the sum in \cref{eqn:cam-loss} is computed only over the new classes.
In order
to train the localizer for the old classes and prevent the encoder from shifting towards the new classes and forgetting the old ones,
WILSON distills knowledge from the old model, by adding two knowledge distillation losses---at intermediate feature and output space.
The first one, $\calL_{\text{KDE}}$, computes the mean-squared error between the features extracted by the current encoder $E^t$ and those extracted by the previous one $E^{t-1}$. The second distillation loss $\calL_{\text{KDL}}$ 
encourages consistency between the pixel-wise scores for old classes predicted by the localizer $(E \circ G)^t$ and those predicted by the old model $(E \circ F)^{t-1}$ (see details in~\cref{sec:app-wilson-kd}).

Finally, 
\wil combines
the localizer output with the old model to generate 
the pseudo-supervision scores $\tilde{q}^c$ that are  used to update the main segmentation module $(E \circ F)^t$, following
\begin{equation}
\label{eqn:wilson-final}
    \calL_{\text{SEG}} (\hat{\bfp}, \tilde{\bfq}) = - \frac{1}{|\calY^t||\calI|}\displaystyle \sum_{i \in \calI} \displaystyle \sum_{c \in \calY^{t}} \tilde{q}^c_i \log (\sigma(\hat{p}^c_i)) + (1 - \tilde{q}^c_i) \log (1 - \sigma(\hat{p}^c_i)),
\end{equation}
where  $\hat{\bfp} = (E \circ F)^t(\bfx)$ are the predictions from the main segmentation head and $\tilde{\bfq}$ is the supervisory signal containing: i) the old model's predictions for the old classes, ii) the localizer's refined scores for the new classes and iii) the minimum between the old model and the localizer scores for the background.
The final objective optimized by WILSON is the non-weighted sum of the
different loss terms defined above, $\calL_{\text{W}} = \calL_{\text{CLS}} + \calL_{\text{KDL}} + \calL_{\text{KDE}}+\calL_{\text{SEG}}$. See~\cref{sec:app-wilson-details} for more details.

\myparagraph{Extending WILSON with \method.}
Since \wil exploits a localizer-based approach designed for weakly supervised SIS, it constitutes a good starting point to integrate and test our proposed semantic prior---without the need for any ad hoc architectural changes.
Therefore, we
complement 
\wil's losses
with our loss introduced in Eq~(\ref{eqn:semantic-loss}),
which
simply requires as input: (i) the output from the localizer $\bfz = (E \circ G)^t(\bfx)$, and (ii) the semantic similarity maps between new and old classes, obtained via~\cref{eqn:semantic-similarity,eqn:binary-tree-sim}.
Endowed with these, our prior loss can be applied together with WILSON losses by simply optimizing
the joint loss $\calL_{\text{J}}=\calL_{\text{W}} + \lambda \calL_{\text{\method}}$. The hyperparameter $\lambda$ controls the strength
of our prior loss, which acts as a regularizer fostering forward transfer from the old to the new classes. 
\vspace{-3mm}

\section{Experiments}
\label{sec:exp}

\myparagraph{Datasets.}
Following 
 \citet{CermelliCVPR22IncrementalLearninginSemSegmfromImageLabels},
we run
experiments on two standard weakly supervised SIS benchmarks: Pascal VOC~\citep{everingham2010pascal} and MS-COCO~\citep{lin2014microsoft}. 
Note that being the \task task more challenging
than both supervised SIS~\citep{xie2021segformer} and class-incremental SIS~\citep{CermelliCVPR20ModelingBackgroundIncrementalLearningSemSegm}, state-of-the-art weakly supervised methods~\citep{xu2022multi} do not experiment on ADE20K~\citep{zhou2017scene} or Cityscapes~\citep{cordts2016cityscapes}, which contain \textit{`stuff'} classes, but  restrict themselves to VOC and MS-COCO, focusing on \textit{`thing'} classes. Similarly, we follow suit with the relevant literature. The \voc benchmark consists of $10,582$ training  and $1,449$ validation images covering 20 semantic categories.
MS-COCO is much larger scale and contains 164k training and 5k validation images from 80 `thing' categories. We adopt the same train and testing splits as used in WILSON. 

\myparagraph{Incremental settings.} We provide empirical evaluation under several incremental learning scenarios, which differ in their splitting between the base and new classes. 
We name the settings
following the notation $N_b$-$N_t$ to indicate
that we first learn with \textit{pixel-level supervision} from $N_b$ \textit{base}  classes, and
then learn sets of $N_t$ \textit{new} classes at a time, with \textit{image-level supervision} only. 
Given a total number of $N$ classes, the number of tasks is ${(N - N_b)}/{N_t}+1$. All the new classes can either be added in a single step, the only scenario explored so far by the WSCI literature,
or can be added in multiple learning steps, a more challenging yet more realistic scenario.

As the name suggests, the \textit{Single-step} settings comprise only one incremental learning phase. 
For instance, in the {\bf15-5 VOC} setting (see Tab.~\ref{tab:voc-overlap-sota}), we
first train the model on 15 base classes from VOC
and then learn the remaining 5 (new) classes 
in a single incremental step 
(bringing the total number of classes to 20).
The newly introduced \textit{Multi-step} settings add new classes to the model in multiple sequential steps.
The {\bf 10-2 VOC} setting, for instance,
considers 10 base classes
and 5 incremental steps which
each learn 2 new classes at a time.
In each table, we indicate results
for base classes as $1$-$N_b$ and for the new ones as $(N_b+1)$-$N$. Differently, the \textbf{COCO-to-VOC} setting involves using the 60 classes exclusive to COCO
in the base training, and performing the incremental learning step(s) on the 20 classes of VOC (\eg, \textbf{60-5} is a 5-step protocol where the 20 VOC classes are learned in 4 increments).

Each incremental setting can be designed following one of these two protocols:
i) \textit{Overlap}, if all the training images for a given step contain at least one instance of a new class, but they can also contain previous 
or even future classes;
ii) \textit{Disjoint}, if each 
step consists of images containing only new or previously seen classes, but 
never
future classes. 
In both protocols, image-level annotations are available for the \textit{new classes only}.
We argue that the multi-step setting with the overlap protocol is the most realistic
one.
That said, we also consider the original settings of WILSON, since it facilitates fair comparison with the previous work. 
 
\myparagraph{Implementation details.}
\label{sec:impl}
Following \wil~\citep{CermelliCVPR22IncrementalLearninginSemSegmfromImageLabels}, we use DeeplabV3~\citep{chen2017deeplab} with ResNet101~\citep{he2016deep} and Wide-ResNet-38~\citep{wu2019wider} as backbone for the VOC and MS-COCO datasets, respectively. The localizer is composed of 3 convolutional layers, interleaved with BatchNorm~\citep{IoffeNeurIPS17BatchRenormTowardsReducingMinibatchDependenceinBNModels} and Leaky ReLU layers. For each step, we train the model with SGD for 40 epochs using a batch size of 24. Since the localizer can produce noisy outputs early in training, we do not use $\calL_{\text{SEG}}$ for the first 5 epochs. We set $\tau=5$ and $\lambda=1$ and follow the values suggested in WILSON for all other hyperparameters. See~\cref{sec:ablation} for sensitivity to $\tau$ and $\lambda$.

\myparagraph{Evaluation metrics.}
We evaluate all models
using the standard mean Intersection over Union (mIoU)~\citep{everingham2010pascal} metric. We report the 
mIoU scores evaluated after the last incremental step. 
We report 3 values:
for the base task (considering
results on the base classes excluding the background), for the subsequent ones (new classes added during the incremental steps) and finally considering all the classes including the background (All). 

\subsection{Main Results}
\label{sec:main_results}

\myparagraph{Comparison with the state of the art.} We compare our proposed \ours with 
several
state-of-the-art class-incremental learning methods that use either pixel-level or image-level annotations in the incremental steps. 
We mainly focus on \task methods, \ie, \wil, since it is the current state-of-the-art method and 
it allows fair comparisons (for instance, methods like EPS~\citep{lee2021railroad} use saliency maps as extra supervision). Pixel-supervised methods are interesting but not comparable as they use a prodigious amount of extra-supervision.
The best performing method with image-level and pixel-level supervision are respectively bolded and underlined in tables.
Since~\cite{CermelliCVPR22IncrementalLearninginSemSegmfromImageLabels} tested \wil only for single-step incremental settings, 
we ran experiments in the other settings using the official implementation 
provided by the authors (\texttt{\href{https://github.com/fcdl94/WILSON}{https://github.com/fcdl94/WILSON}}). For comparability, we also re-ran experiments on single-task settings.
``WILSON$\dagger$'' indicates our reproduced results while ``WILSON'' corresponds to the original numbers from the paper.
We further report in tables the relative gain/drop in performance (in \%) of our \ours \wrt WILSON$\dagger$, within brackets.

\begin{wrapfigure}[17]{r}{0.5\textwidth}
  \vspace{-18pt}
  \begin{center}
\begin{tikzpicture}

\definecolor{darkslategray38}{RGB}{38,38,38}
\definecolor{lavender234234242}{RGB}{234,234,242}
\definecolor{lightgray204}{RGB}{204,204,204}
\definecolor{steelblue76114176}{RGB}{76,114,176}

\begin{axis}[
axis background/.style={fill=lavender234234242},
axis line style={line width=.1mm},
legend cell align={left},
legend style={
  font=\tiny,
  fill opacity=1,
  draw opacity=1,
  text opacity=1,
  draw=gray,
  fill=white
},
height=0.65\linewidth,
width=\linewidth,
tick align=outside,
x grid style={white},
xlabel=\textcolor{darkslategray38}{Number of tasks \(\displaystyle \longrightarrow\)},
xmajorgrids,
xmajorticks=true,
xtick pos=left,
xmin=1, xmax=12,
xtick style={color=darkslategray38},
y grid style={white},
ylabel=\textcolor{darkslategray38}{\(\displaystyle \Delta\) (in \%) \(\displaystyle \longrightarrow\)},
ymajorgrids,
ymajorticks=true,
ytick pos =left,
ymin=-4, ymax=30,
ytick style={color=darkslategray38},
xtick={1, 2, 3, 4, 5, 6, 7, 8, 9, 10, 11, 12}
]
\addplot [thick, red, dashed, mark=triangle*, mark size=2.6, mark options={solid}]
table {%
2 1.6
3 5.6
6 26.8
};
\addlegendentry{VOC}
\addplot [thick, steelblue76114176, dashed, mark=square*, mark size=2, mark options={solid}]
table {%
2 -0.7
5 -0.7
11 11.6
};
\addlegendentry{COCO-to-VOC}
\draw (axis cs:2.3,0.6) node[
  scale=0.7,
  anchor=base west,
  text=darkslategray38,
  rotate=0.0
]{1.6};
\draw (axis cs:3.3,4.6) node[
  scale=0.7,
  anchor=base west,
  text=darkslategray38,
  rotate=0.0
]{5.6};
\draw (axis cs:6.3,25.8) node[
  scale=0.7,
  anchor=base west,
  text=darkslategray38,
  rotate=0.0
]{26.8};
\draw (axis cs:2,-2.7) node[
  scale=0.7,
  anchor=base west,
  text=darkslategray38,
  rotate=0.0
]{-0.7};
\draw (axis cs:5,-2.7) node[
  scale=0.7,
  anchor=base west,
  text=darkslategray38,
  rotate=0.0
]{-0.7};
\draw (axis cs:11,9.6) node[
  scale=0.7,
  anchor=base west,
  text=darkslategray38,
  rotate=0.0
]{11.6};
\end{axis}

\end{tikzpicture}
  \end{center}
  \vspace{-3mm}
  \caption{Our \method's relative percentage gain ($\Delta$ in \%) over \wil on new class performance for the VOC and COCO-to-VOC tasks}
  \label{fig:sota-summary}
\end{wrapfigure}
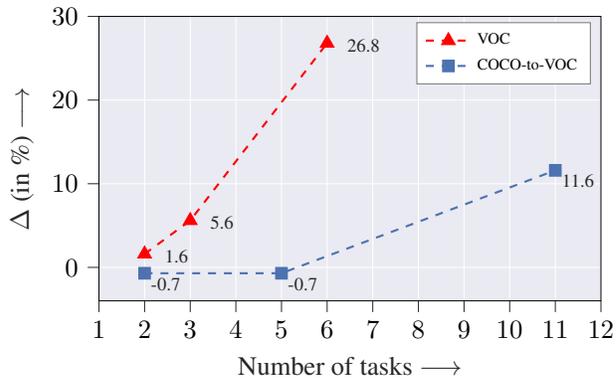

\begin{table}[!t]
    \centering
    \small
    \setlength{\tabcolsep}{3pt}
    \renewcommand{\arraystretch}{.5}
    \begin{tabular}{llc|ccc|ccc}
        \toprule
        & \multirow{2}{*}{\bf{Method}} & \multirow{2}{*}{\bf{Supervision}} & \multicolumn{3}{c|}{\textbf{15-5} (2 tasks)} & \multicolumn{3}{c}{\textbf{10-10} (2 tasks)}  \\
        & & &  1-15~~ & 16-20~~ & All & 1-10 ~~ & 11-20~~ & All  \\
        \midrule
        \multirow{12}{*}{\vspace{-16mm}\rotatebox{90}{{\textbf{single-step}}}} & Fine-Tuning & Pixel & 12.5 & 36.9 & 18.3 & 7.8 & 58.9 & 32.1 \\
        & LWF~\citep{LiECCV16LearningWithoutForgetting} & Pixel & 67.0 & 41.8 & 61.0 & \underline{70.7} & 63.4 & 67.2 \\
       
        & PLOP~\citep{DouillardCVPR21PLOPLearningWithoutForgettingContinualSemSegm} & Pixel & \underline{75.7} & 51.7 & \underline{70.1} & 69.6 & 62.2 & 67.1 \\
        & SDR~\citep{michieli2021continual} & Pixel & 75.4 & 52.6 & 69.9 & 70.5 & \underline{63.9} & \underline{67.4} \\
        
        & RECALL~\citep{MaracaniICCV21RECALLReplayBasedContinualLearningSemSegm} & Pixel & 67.7 & \underline{54.3} & 65.6 & 66.0 & 58.8 & 63.7 \\
        
        \cmidrule{2-9}
        
        & CAM~\citep{ZhouCVPR16LearningDeepFeaturesForDiscLoc} & Image & 69.9 & 25.6 & 59.7 & 70.8 & 44.2 & 58.5 \\
        & SEAM~\citep{wang2020self} & Image & 68.3 & 31.8 & 60.4 & 67.5 & 55.4 & 62.7\\
        & SS~\citep{AraslanovCVPR20SingleStageSemSegmFromImageLabels} & Image & 72.2 & 27.5 & 62.1 & 69.6 & 32.8 & 52.5\\
        & EPS~\citep{lee2021railroad} & Image & 69.4 & 34.5 & 62.1 & 69.0 & 57.0 & 64.3 \\
        & WILSON~\citep{CermelliCVPR22IncrementalLearninginSemSegmfromImageLabels} & Image & 74.2 & 41.7 & 67.2 & 70.4 & 57.1 & 65.0\\
        \cmidrule{2-9}
        & WILSON$\dagger$~\citep{CermelliCVPR22IncrementalLearninginSemSegmfromImageLabels} & Image & \textbf{76.3} & 44.1 & 69.3 & 71.4 & 56.1 & 64.9 \\
        
        & \ours (Ours) & Image & \makecell{76.2\\(\down0.1\%)} & \makecell{\textbf{47.0}\\(\up6.6\%)} & \makecell{\textbf{70.0}\\(\up1.0\%)} & \makecell{\textbf{72.3}\\(\up1.3\%)} & \makecell{\textbf{57.2}\\(\up1.6\%)} & \makecell{\textbf{65.9}\\(\up1.5\%)}  \\
        \midrule
        \midrule
        
        \multirow{5}{*}{\vspace{-8mm}\rotatebox{90}{{\textbf{multi-step}}}} & \multirow{2}{*}{ } & \multirow{2}{*}{ } & \multicolumn{3}{c|}{\textbf{10-5} (3 tasks)} & \multicolumn{3}{c}{\textbf{10-2} (6 tasks)}  \\
        & & &  1-10 & 11-20 &All & 1-10 & 11-20&All  \\
        
        \cmidrule{2-9}
        
         &  WILSON$\dagger$~\citep{CermelliCVPR22IncrementalLearninginSemSegmfromImageLabels} & Image & 66.8 & 46.5 &	58.1 & 38.7 & 22.4 & 32.5 \\
        
        & \ours (Ours) & Image & \makecell{\textbf{68.8}\\(\up3.0\%)} & \makecell{\textbf{49.1}\\(\up5.6\%)} & \makecell{\textbf{60.4}\\(\up4.0\%)} & \makecell{\textbf{44.5}\\(\up15.0\%)} & \makecell{\textbf{28.4}\\(\up26.8\%)} & \makecell{\textbf{38.6}\\(\up18.8\%)}\\
        \bottomrule
    \end{tabular}
    \vspace{-2mm}
    \caption{
    \textbf{VOC results.} The mIoU (in \%) scores for both \textit{single-step} (top half) and \textit{multi-step} (bottom half) \textbf{overlap} incremental settings on \voc. 
    For each experiment, the three different columns indicate performance on 
    base, new and all 21 classes (including background), respectively.
    For \ours (Ours), we further report the relative gain/drop in performance (in \%) \wrt WILSON$\dagger$
    }
    \vspace{-4mm}
    \label{tab:voc-overlap-sota}
\end{table}
\begin{table}[!t]
    \centering
    \small
    \setlength{\tabcolsep}{4.pt}
    \renewcommand{\arraystretch}{.5}
    \resizebox{\textwidth}{!}{%
    \begin{tabular}{llc|ccc|c|ccc|c}
        \toprule
        
        & \multirow{4}{*}{\bf{Method}} & \multirow{4}{*}{\bf{Supervision}} & \multicolumn{8}{c}{\textbf{60-20} (2 tasks)} \\
        
        & & & \multicolumn{6}{c|}{\bf{COCO}} & \multicolumn{2}{c}{\bf{VOC}} \\
        \cmidrule{4-9} \cmidrule{10-11}
        
        & & & \multicolumn{2}{c}{1-60} & \multicolumn{2}{c}{61-80} & \multicolumn{2}{c|}{All} & \multicolumn{2}{c}{All} \\
        \midrule
        
        \multirow{11}{*}{\vspace{-16mm}\rotatebox{90}{{\textbf{single-step}}}} & Fine-Tuning & Pixel & \multicolumn{2}{c}{1.9} & \multicolumn{2}{c}{41.7} & \multicolumn{2}{c|}{12.7} & \multicolumn{2}{c}{\underline{75.0}} \\
        
        & LWF~\citep{LiECCV16LearningWithoutForgetting} & Pixel & \multicolumn{2}{c}{36.7} & \multicolumn{2}{c}{\underline{49.0}} & \multicolumn{2}{c|}{\underline{40.3}} & \multicolumn{2}{c}{73.6}\\
        
        & ILT~\citep{michieli2019incremental} & Pixel & \multicolumn{2}{c}{\underline{37.0}} & \multicolumn{2}{c}{43.9} & \multicolumn{2}{c|}{39.3} & \multicolumn{2}{c}{68.7} \\
        
        & PLOP~\citep{DouillardCVPR21PLOPLearningWithoutForgettingContinualSemSegm} & Pixel & \multicolumn{2}{c}{35.1} & \multicolumn{2}{c}{39.4} & \multicolumn{2}{c|}{36.8} & \multicolumn{2}{c}{64.7} \\
        
        \cmidrule{2-11}
        
        & CAM~\citep{ZhouCVPR16LearningDeepFeaturesForDiscLoc} & Image & \multicolumn{2}{c}{30.7} & \multicolumn{2}{c}{20.3} & \multicolumn{2}{c|}{28.1} & \multicolumn{2}{c}{39.1} \\
        
        & SEAM~\citep{wang2020self} & Image & \multicolumn{2}{c}{31.2} & \multicolumn{2}{c}{28.2} & \multicolumn{2}{c|}{30.5} & \multicolumn{2}{c}{48.0} \\
        
        & SS~\citep{AraslanovCVPR20SingleStageSemSegmFromImageLabels} & Image & \multicolumn{2}{c}{35.1} & \multicolumn{2}{c}{36.9} & \multicolumn{2}{c|}{35.5} & \multicolumn{2}{c}{52.4} \\
        
        & EPS~\citep{lee2021railroad} & Image & \multicolumn{2}{c}{34.9} & \multicolumn{2}{c}{38.4} & \multicolumn{2}{c|}{35.8} & \multicolumn{2}{c}{55.3} \\
        
        & WILSON~\citep{CermelliCVPR22IncrementalLearninginSemSegmfromImageLabels} & Image & \multicolumn{2}{c}{39.8} & \multicolumn{2}{c}{\bf{41.0}} & \multicolumn{2}{c|}{40.6} & \multicolumn{2}{c}{\bf{55.7}} \\
        
        \cmidrule{2-11}
        
        & WILSON$\dagger$~\citep{CermelliCVPR22IncrementalLearninginSemSegmfromImageLabels} & Image & \multicolumn{2}{c}{\bf{41.1}} & \multicolumn{2}{c}{\bf{41.0}} & \multicolumn{2}{c|}{\bf{41.6}} & \multicolumn{2}{c}{54.8} \\
        
        & \ours (Ours) & Image & \multicolumn{2}{c}{\makecell{\bf{41.1}\\(0.0\%)}} & \multicolumn{2}{c}{\makecell{40.7\\(\down0.7\%)}} & \multicolumn{2}{c|}{\makecell{\bf{41.6}\\(0.0\%)}} & \multicolumn{2}{c}{\makecell{54.4\\(\down0.7\%)}}\\
        
        \midrule
        \midrule
        
        \multirow{5}{*}{\vspace{-11mm}\rotatebox{90}{{ \textbf{multi-step}}}} & & & \multicolumn{4}{c|}{\textbf{60-5} (5 tasks)} & \multicolumn{4}{c}{\textbf{60-2} (11 tasks)} \\
        
        & & & \multicolumn{3}{c|}{\bf{COCO}} & \bf{VOC} & \multicolumn{3}{c|}{\bf{COCO}} & \bf{VOC}\\
        
        \cmidrule{4-6} \cmidrule{7-8} \cmidrule{8-10} \cmidrule{10-11}
        
        & & & 1-60 & 61-80 & All & All & 1-60 & 61-80 & All & All\\
        \cmidrule{2-11}
        
        & WILSON$\dagger$~\citep{CermelliCVPR22IncrementalLearninginSemSegmfromImageLabels} & Image & 30.1 & 28.0 & 30.2 & \bf{42.0} & 10.2 & 14.8 & 12.2 & 24.1 \\
        
        & \ours (Ours) & Image & \makecell{\bf{33.0}\\(\up9.6\%)} & \makecell{\bf{28.2}\\(\up0.7\%)} & \makecell{\bf{32.5}\\(\up7.6\%)} & \makecell{41.7\\(\down0.7\%)} & \makecell{\bf{14.6}\\(\up43.1\%)} & \makecell{\bf{16.5}\\(\up11.5\%)} & \makecell{\bf{15.9}\\(\up30.3\%)} & \makecell{\bf{26.9}\\(\up11.6\%)} \\
        
        \bottomrule
    \end{tabular}%
    }
    \caption{
    \textbf{COCO-to-VOC results.} The mIoU (in \%) scores for both \textit{single-step} (top half) and \textit{multi-step} (bottom half) \textbf{overlap} incremental settings on \textbf{COCO-to-VOC}. 
   For each experiment, the first three columns indicate performance on base, new and all classes (81 including background) computed on COCO, respectively; last column indicates performance on all classes for \voc 
    }
    \label{tab:coco-voc-sota}
\end{table}

\myparagraph{Results.}
We summarize the results of the VOC and COCO-to-VOC experiments in Fig.~\ref{fig:sota-summary}. In particular we report \method's relative percentage gain (in \%) over \wil and observe that, as the number of incremental steps increases, the overall gain of \method over \wil for the new classes becomes more and more noteworthy. Next we elaborate the results for each benchmark.

In Tab.~\ref{tab:voc-overlap-sota}, we show results for
both single-step and multi-step incremental settings---on \voc, using the \textit{overlap} protocol. 
We observe that our \ours outperforms \wil in almost all the considered settings. In particular, the relative gain (in \%) \wrt \wil grows wider as the number of incremental steps increases, with \ours achieving +26.8\% relative improvement over WILSON in new class performance, in the 10-2 setting. 
Not only our semantic-prior loss improves new class performance 
but 
also it leads to
15\% lesser forgetting \wrt \wil.
We provide qualitative examples in Fig.~\ref{fig:qualitative_visualization} and in the Fig.~\ref{sec:app-qual}, showing how the semantic maps 
aid the final segmentation.

\begin{figure*}[!t]
    \centering
    \includegraphics[width=0.9\textwidth]{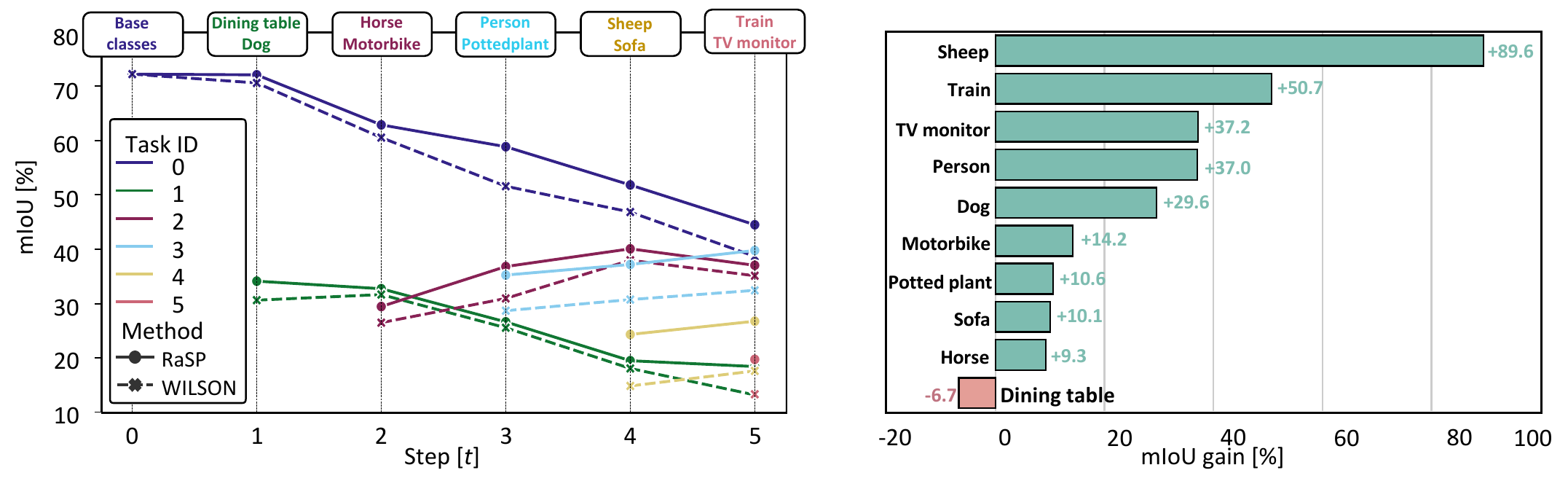}
    \vspace{-0.4cm}
  \caption{\textbf{Left:}  Per-task and per-step  mIoU for the 10-2 VOC \textit{multi-step} \textbf{overlap} incremental setting. \textbf{Right:} Per class gain/drop
  of \method \wrt WILSON, evaluated for each class in the step it was learned. 
We observe that \method encounters lesser forgetting and improved learning of the new classes
  }
\label{fig:main_plots}
\vspace{-3mm}
\end{figure*}

Fig.~\ref{fig:main_plots} (left) shows the mIoU scores per task and per step for the 
10-2 VOC \textit{overlap} setting, indicating which classes are learned at each step (for \wil and \ours). 
This plot shows how our method consistently
improves over \wil throughout the learning sequence, while at the same time \textit{forgetting less}.
In Fig.~\ref{fig:main_plots} (right), we report \ours's per-class relative percentage improvement \wrt \wil,
computed at each step. It is reasonable that due to the semantically relatedness between the classes such as `sheep' (new) and `cow' (old), the relative performance gain obtained by \method on `sheep' is way higher than for `sofa' (new), where no semantically related object is to be found. We show more of such visualizations in Fig.~\ref{fig:app-plots} and Fig.~\ref{fig:app-gains}.

We report the results for 
the \ctov setting (\textit{overlap} protocol) in Tab.~\ref{tab:coco-voc-sota}. 
Our \method performs comparably with \wil in the single-step setting, but outperforms \wil 
when the number of new classes learned in each task decreases and the number of tasks increases---from 60-5 (5 tasks) to 60-2 (11 tasks).
In the longer incremental scenarios, we observe not only improvements for the new classes, but also more limited forgetting of the old ones. This highlights that our \method loss is most effective when the sequence of tasks get longer, \ie, in more plausible settings.

\subsection{Learning one new class at a time}
\label{sec:rehearsal}

A major limitation of the state-of-the-art approaches for \task is that their performance degrades when learning one new class at a time:
the model fails to learn the new class and undergoes drastic forgetting.
This is due to the fact that
\cref{eqn:cam-loss} is optimized for a single positive class: 
the 
lack
of negative classes leads to gross overestimation of the foreground, causing the localizer to provide 
poor
pseudo labels to the 
segmentation head, with a negative effect on old classes as well.
We show in Tab.~\ref{tab:mem-voc-overlap} (top-half) results of \wil and \ours for two single-class incremental settings (15-1 and 10-1), using \voc.
Both
methods
struggle with learning new classes, yielding 
poor performance
compared to pixel-supervised 
methods.
These fully-supervised methods
can learn the new classes better since
their annotations are composed of
both positive foreground-object pixels and negative background pixels.

\myparagraph{A solution: episodic memory.} 
To circumvent this issue we store a small number of images from the previous classes in a fixed-size memory $\mathcal{M}$. Intuitively, the samples in the memory help the localizer by providing negative samples (via pseudo labels on the old objects). We show in Tab.~\ref{tab:mem-voc-overlap} (bottom-half) that storing as little as 100 past samples from the previous classes dramatically improves the learning on new classes for both
\wil and \ours, with \ours + $\mathcal{M}$ outperforming \wil + $\mathcal{M}$ 
(28.3\% \textit{vs} 20.8\% in the 15-1 setting). 
Unsurprisingly, 
it also helps retaining performance on the base classes. Similar observations hold for the 10-1 setting.

\myparagraph{External data as an alternative.}
Inspired by the class-incremental SIS method RECALL~\citep{MaracaniICCV21RECALLReplayBasedContinualLearningSemSegm}, we 
consider the option of retrieving samples of the previous classes from external sources. We define this 
memory as $\mathcal{M}_\text{ext}$. 
Concretely, we retrieve 100 samples per old class from ImageNet (by creating a 
mapping with the \voc classes). As shown in Tab.~\ref{tab:mem-voc-overlap}, this further improves
both
\wil + $\mathcal{M}_\text{ext}$ and \ours + $\mathcal{M}_\text{ext}$ compared to the previous episodic memory $\mathcal{M}$. RECALL performs better on new classes, but 
i) relies on pixel-level supervision and ii) uses significantly more web-crawled images---therefore, it is not directly comparable. 

\begin{table}[t!]
    \centering
    \small
    \resizebox{0.9\textwidth}{!}{%
    \begin{tabular}{ll|c|ccc|ccc}
        \toprule
        & \multirow{2}{*}{\bf{Method}} & \multirow{2}{*}{\bf{Supervision}} & \multicolumn{3}{c|}{{\bf 15-1} (6 tasks)} & \multicolumn{3}{c}{{\bf 10-1} (11 tasks)}\\
        \cmidrule{4-6} \cmidrule{7-9}
         & & & 1-15 & 16-20 & All & 1-10 & 11-20 & All\\
        
        \midrule
        
        \multirow{4}{*}{\rotatebox{90}{{w/o memory}}} & ILT~\citep{michieli2019incremental} & Pixel & 4.9 & 7.8 & 5.7 & 16.5 & 1.0 & 9.1 \\
        
        & MiB~\citep{CermelliCVPR20ModelingBackgroundIncrementalLearningSemSegm} & Pixel & 35.1 & 13.5 & 29.7 & 15.1 & 14.8 & 15.0 \\
        
        \cmidrule{2-9}
        
        & WILSON$\dagger$~\citep{CermelliCVPR22IncrementalLearninginSemSegmfromImageLabels} & Image & 0.0 & 2.3 & 0.6 & 0.0 & 0.2 & 0.1 \\
        
        & \ours (Ours) & Image & 17.7 & 0.9 & 13.2 & 2.0 & 0.7 & 1.3 \\
        
        \midrule
        \midrule
        
        \multirow{5}{*}{\rotatebox{90}{{w/ memory}}} & WILSON$\dagger$ + $\mathcal{M}$ & Image & 61.5 & 20.8 & 52.5 & 33.4 & 24.6 & 30.0\\
        
        & \ours (Ours) + $\mathcal{M}$ & Image & 63.3 & 28.3 & 56.0 & 38.9 & 30.9 & 36.9 \\
        
        \cmidrule{2-9}
        
        & WILSON$\dagger$ + $\mathcal{M}_\text{ext}$ & Image & {\bf 75.7} & 32.9 & 65.9 & {\bf 66.8} & 34.9 & 52.3 \\
        & \ours (Ours) + $\mathcal{M}_\text{ext}$ & Image & {\bf 75.7} & {\bf 35.2} & {\bf 66.6} & {\bf 66.8} & {\bf 39.1} & {\bf 54.4} \\
        
        \cmidrule{2-9}
        
        & RECALL (Web)~\citep{MaracaniICCV21RECALLReplayBasedContinualLearningSemSegm} & Pixel & \underline{67.8} & \underline{50.9} & \underline{64.8} & \underline{65.0} & \underline{53.7} & \underline{60.7} \\
        \bottomrule
    \end{tabular}
    }
    \caption{\textbf{Effect of memory.} Results on single-class \textit{multi-step} \textbf{overlap} incremental setting on \voc. 
    $\mathcal{M}$ and $\mathcal{M}_\text{ext}$ indicate memories of previously seen or external samples, respectively
    }
    \label{tab:mem-voc-overlap}
\end{table}

\subsection{Class-Incremental Few-shot Segmentation}

We compare \ours with \wil on 
the task of
Incremental Few-Shot Segmentation (iFSS)~\citep{CermelliBMVC21PrototypeBasedIncFewShotSemSegm},
where the model learns incrementally from only few images per new class (\eg, 2 or 5 images). 
This is a 
challenging setting, only tested so far with pixel-level supervision for the new classes. Here, we 
add the challenging constraints that
the training images for the new classes are
only weakly annotated, \ie, with image-level labels.
Following \citet{CermelliBMVC21PrototypeBasedIncFewShotSemSegm}, we consider 
4 folds of 5 new classes for PASCAL-5\textsuperscript{\textit{i}} and the 4 folds of 20 new classes for COCO-20\textsuperscript{\textit{i}}, where each fold is used, in turn, as incremental setting 
with the other classes defining the base task. 

\cref{tab:ifss} reports the results averaged over the 4 folds (per-fold results are reported in the~\cref{sec:app-fss}). 
The bottom rows of \cref{tab:ifss} reports the results obtained by the \task methods \wil and \ours.
As expected, in the case of COCO-20\textsuperscript{\textit{i}} both methods perform poorly when compared to the pixel-supervised methods, which is not surprising as even the strongly supervised methods (top rows) have difficulties to learn the new classes.
On the other hand, 
in PASCAL-5\textsuperscript{\textit{i}},
not only \ours consistently outperforms \wil, but in the 5-shot case it also outperforms or performs on par with some of the
strongly supervised  methods. Finally, we can observe that the performance of \ours on the base classes remains comparable and is sometimes higher than most of the strongly supervised methods, where the higher performance on the new classes tends to come with a 
more severe forgetting.  

\begin{table}[t!]
    \centering
    \small
    \setlength{\tabcolsep}{1.pt}
    \resizebox{\textwidth}{!}{%
    \begin{tabular}{l|c|ccc|ccc|ccc|ccc}
    \toprule
         \multirow{2}{*}{\bf{Method}} & \multirow{2}{*}{\bf{Supervision}} & \multicolumn{3}{c|}{\bf{VOC (5-shot)}} & \multicolumn{3}{c|}{\bf{VOC (2-shot)}} & \multicolumn{3}{c|}{\bf{COCO (5-shot) }} & \multicolumn{3}{c}{\bf{COCO (2-shot)}}   \\
        \cmidrule{3-14}
         
          & & 1-15 & 16-20 & HM & 1-15 & 16-20 & HM & 0-60 & 61-80 & HM & 0-60 & 61-80 & HM \\ 
         \midrule
         
         Fine-Tuning & Pixel & 
         55.8 & 29.6 & 38.7 &
         59.1 & 19.7 & 29.5 & 
         41.6 & 12.3 & 19.0 &
         41.5 & 7.3 & 12.4  \\
      \midrule
    
             WI~\citep{QiCVPR18LowShotLearningWithImprintedWeights} & Pixel & 63.3 & 21.7 & 32.3 &
         63.3 & 19.2 & 29.5  &
         43.6 & 8.7 & 14.6 &
         44.2 & 7.9 & 13.5 \\

         AMP~\cite{SiamICCV19AMPAdaptiveMaskedProxies4FewShotSegm} & Pixel	& 51.9 &	18.9 &	27.7  
         & 54.4 & 18.8 & 27.9 & 
         34.6 & 11.0& 16.7 & 
           35.7 & 8.8 & 14.2 \\
         
          MiB~\citep{CermelliCVPR20ModelingBackgroundIncrementalLearningSemSegm} & Pixel & \underline{65.0} &	28.1 &	39.3 &	  
         63.5 & 12.7 & 21.1 & 
         44.7 & 11.9 & 18.8 &
         44.4 & 6.0 & 10.6  \\

          PIFS~\citep{CermelliBMVC21PrototypeBasedIncFewShotSemSegm}  & Pixel & 60.0 &	33.3 &	\underline{42.8} &
         60.5 & \underline{26.4} & \underline{36.8} & 
         42.8 & \underline{15.7} & \underline{23.0} &
         40.9 & \underline{11.1} & \underline{17.5} \\
         \midrule
         
         WILSON$\dagger$~\citep{CermelliCVPR22IncrementalLearninginSemSegmfromImageLabels} & Image & 64.1 &	20.5 &	31.1 & 63.3 & 10.2 &	17.6 & 
         45.0 &	\bf{5.8} &	\bf{10.3} &
          \bf{43.6} &	1.9 &	3.6  \\
         
          \ours (Ours) & Image &	\makecell{\bf{64.4}\\(\up0.5\%)} &	\makecell{\bf{21.3}\\(\up3.9\%)} &	\makecell{\bf{32.0}\\(\up2.9\%)} &	
         \makecell{\bf{63.5}\\(\up0.3\%)} & \makecell{\bf{10.7}\\(\up4.9\%)} & \makecell{\bf{18.3}\\(\up4.0\%)} & 
         \makecell{\bf{45.1}\\(\up0.2\%)} &	\makecell{5.6\\(\down3.4\%)} &	\makecell{10.0\\(\down2.9\%)} &	
         \makecell{43.5\\(\down0.2\%)} &	\makecell{\bf{2.0}\\(\up5.3\%)} &	\makecell{\bf{3.8}\\(\up5.6\%)}  \\
         \bottomrule
    \end{tabular}%
    }
    \caption{\textbf{Few-shot results.} The mIoU (in \%) scores for the \textit{single-step} (2 tasks) incremental few-shot SiS settings on the \vocfs and \cocofs benchmarks, for 5-shot and 2-shot cases. We show the average results over the 4 folds as in \citep{CermelliBMVC21PrototypeBasedIncFewShotSemSegm}.
    For each experiment, columns report performance on old classes, new classes, and the Harmonic-Mean (HM) of the two scores. 
    For \ours (Ours), we also report the relative gain/drop in performance (in \%) \wrt WILSON$\dagger$
    }
    \label{tab:ifss}
    
\end{table}

\paragraph{Limitations.}
In edge cases, where the new classes are very dissimilar to the old classes, our proposed \method loss will not bring tangible improvements. However, in practical applications one can reasonably assume that a model has already learned an array of primitive classes (often leveraging stronger pixel-level supervision), and that the incremental learner will encounter new objects that have some degree of resemblance to those primitive classes. 
We posit that such limitation will become less and less relevant as the model learns a large number of new classes, since the likelihood of finding an old class similar to each new one at hand increases more and more over the model lifetime.

\section{Conclusions}
\label{sec:concl}

We proposed a method for Weakly Supervised Class-Incremental Semantic Image Segmentation, where the model is tasked with incrementally learning to segment new objects using weaker image labels as supervision. Guided by the observation that new classes to be learned by the model often bear resemblance with the old ones that it already knows, we designed a Relation-aware Semantic Prior (\ours) loss that fosters forward transfer. It transfers objectness prior from the past model by leveraging the semantic similarity between old and new class names and aids the model in learning new categories. We validated our proposed method on a wide variety of incremental scenarios derived from standard benchmarks. In particular, we demonstrated that our method is resilient in unexplored and challenging scenarios, where the number of tasks is high (or number of classes in each task is low) and reduced data availability for each task.

\clearpage

\bibliography{collas2023_conference}
\bibliographystyle{collas2023_conference}

\clearpage

\setcounter{table}{0}
\renewcommand{\thetable}{A\arabic{table}}%
\setcounter{figure}{0}
\renewcommand{\thefigure}{A\arabic{figure}}%
\setcounter{equation}{0}
\renewcommand{\theequation}{A\arabic{equation}}%

\appendix

\section*{Appendix}

The Appendix is organized as follows: Sec.~\ref{sec:app-impl-details} provides implementation details of \ours. Sec.~\ref{sec:app-add-exp} includes additional experimental results on ablation study, different class orderings, classwise performance, class-incremental few-shot segmentation. Sec.~\ref{sec:app-wilson-details} lists the details about the \wil framework. Sec.~\ref{sec:app-qual} provides additional qualitative results. Finally, in Sec.~\ref{sec:app-discuss} we conclude with a discussion.

\section{Implementation details of RaSP}
\label{sec:app-impl-details}

\subsection{Semantic Similarity Metric}
\label{sec:app-sem-sim}

The similarity metric ${\bf S}_\Omega$ used in the \cref{eqn:binary-tree-sim} of the main paper is derived from the    cosine distance, which is computed between a pair of class label names as:
    
\begin{equation}
    {\bf S}_\Omega = - (1 - \frac{\omega(c_i) \cdot \omega(c_j)}{||\omega(c_i)||_2 ||\omega(c_j)||_2}).
\end{equation}
where $\omega(c_i)$ and $\omega(c_j)$ represent the vectorial embeddings for the $i$\textsuperscript{th} and $j$\textsuperscript{th} classes. The value of ${\bf S}_\Omega$ is then substituted to the \cref{eqn:binary-tree-sim} of the main paper. Note that higher the semantic similarity between a pair of labels $c_i$ and $c_j$, higher is the $s^c_i$ value.
    
We obtain the vectorial embedding $\omega(c)$ corresponding to a class label name $l_c$ using 
the BERT transformer~\citep{DevlinNAACL19BERT}.
In details, we prompt the transformer with the class label name to obtain a 768-dimensional vector representation $\omega(c) = \texttt{Transformer}(``\texttt{An image of a }\{l_c\}")$. While one could omit the prompt and simply provide the class label name, we do it to give context to the transformer that the class label name is a \texttt{noun}. Please note that our method can work with other semantic mapping functions, \textit{e.g.}, Word2Vec (see Tab.~\ref{tab:sem-sim-type}).

\subsection{Selective backrpopagation of Rasp loss}
\label{sec:app-rasp-impl}

To recap, we compute the semantic similarity maps (described in \cref{eqn:binary-tree-sim} of the main paper) only for the new foreground classes $\mathcal{C}^t$ present in an incremental step $t$. In other words, the semantic map $s^{bkg}$ for the \textit{bkg} class is not computed, and not enforced by the optimization in \cref{eqn:semantic-loss}. Moreover, we selectively backpropagate the RaSP loss $\mathcal{L}_{\text{RaSP}}$ only for those new class channels of the localizer $G_t$ for which ground truth \textit{image labels} are available. As an example, in an incremental step $t$ if there are five new classes, $|\mathcal{C}^t| = 5$, and if for a given image only the new class ``dog'' is present, then we simply backpropagate the gradients of the RaSP loss for the ``dog'' channel only. All the other channels, including the \textit{bkg} channel, are ignored during the backpropagation. Given the fact that the old model does not perfectly predict the new classes as \textit{bkg} and is spuriously activated as foreground for the new classes (see the $(F \circ  E)^{t-1}({\bf x}_t)$ column in Fig.~\ref{tab:qualitative_visualization2} where new class objects are not \textit{bkg}), the RaSP loss in practice does not largely suppress the CAM loss. We hope that our new findings will encourage future WSCI works to tackle overconfident model predictions on unseen classes.

\subsection{Role of normalization in semantic similarity}
\label{sec:app-normalization}

Here we expand on the role of normalization introduced in \cref{eqn:binary-tree-sim}. Our proposed RaSP first computes the semantic similarity between a given new class $c$ and all the old classes (including the ``bkg'' class) in the image following Eq.~\ref{eqn:semantic-similarity}. It means that for the pixel locations predicted as ``bkg'' by the old model $(E \circ F)^{t-1}$, each new class can have a non-zero semantic similarity with different scale, which can be detrimental for learning the new class. To this end, we enforce that the semantic similarity $s^c_i$ is always lower-bounded by 1 (\textit{a.k.a} normalization) for the pixels corresponding to the ``bkg'' class for every new class $c$ (using \cref{eqn:binary-tree-sim}). Then, as shown in the RaSP loss of \cref{eqn:semantic-loss}, we apply a squashing function (sigmoid) such that the network activations corresponding to such pixels are suppressed, and only foreground pixels with high semantic similarities are encouraged by the network.

\section{Additional experiments}
\label{sec:app-add-exp}

\subsection{Hyperparameter Sensitivity Analysis}
\label{sec:ablation}

\begin{figure*}[h!]
    \centering
    \includegraphics[width=\textwidth]{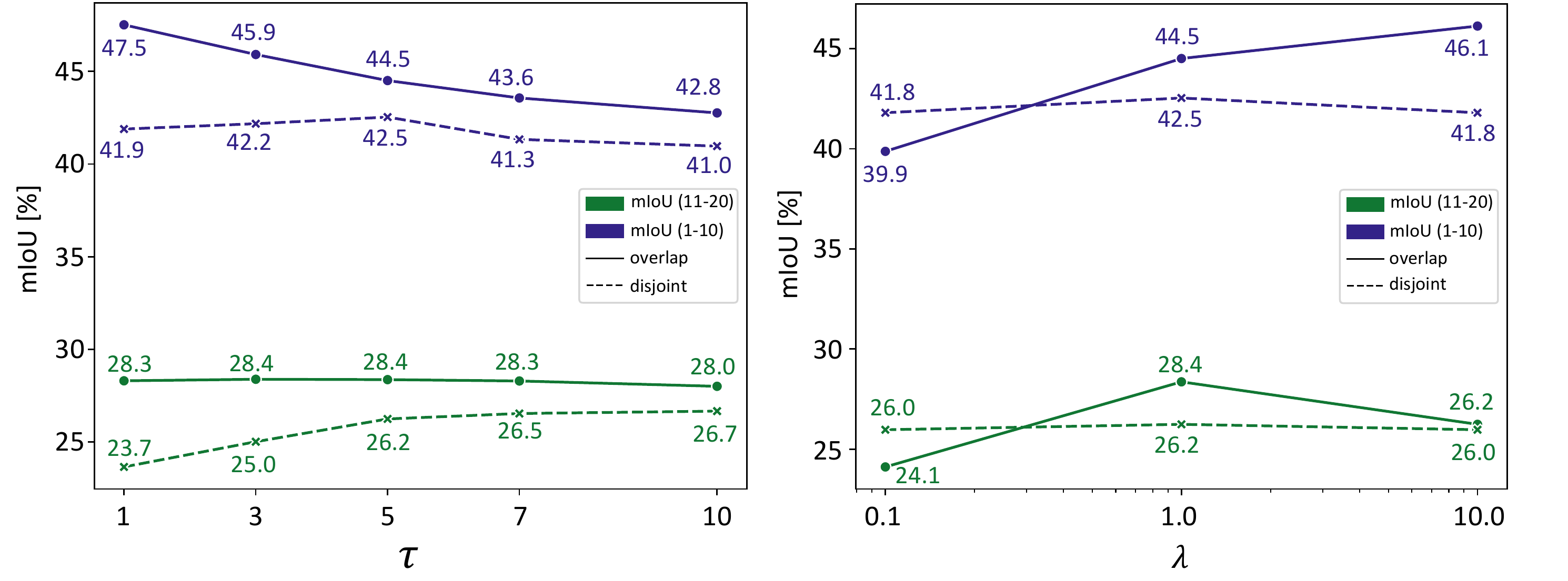}
  \caption{\textbf{Ablating $\tau$ (left) and $\gamma$ (right).} \ours results on \voc, using the 10-2 setting (6 tasks). Solid and dashed lines indicate overlap and disjoint results, respectively. Blue and green lines indicate performance on old and new classes, respectively}
\label{fig:ablations_tau_gamma}
\end{figure*}

In Fig.~\ref{fig:ablations_tau_gamma} we show 
how results are affected when we vary the hyperparameters $\tau$ and  $\lambda$ in the case of 10-2 VOC \textit{ multi-step} \textbf{overlap} (solid) and \textbf{disjoint} (dashed)
incremental settings, reporting both performance on old and new classes (in blue and green, respctively). To recap, $\tau$ plays a role in computing the similarity maps via \cref{eqn:binary-tree-sim}; in particular, it is a scaling factor that controls how steep the decay is, as two semantic entities are more or less similar. Instead, $\gamma$ controls the strength of our \ours loss: the larger the $\gamma$, the higher the impact of our prior over the other terms.

For our experiments, we have selected $\tau=5$ and $\gamma=1$,  the former by observing that it provides a sufficiently steep decay, the latter following \wil's approach of not assigning different weights to the different terms, since they operate at similar scales. We can observe in Fig.~\ref{fig:ablations_tau_gamma} that i) \ours is satisfactorily robust against the choice of these hyperparameters and ii) better results than the ones proposed in the main paper can be obtained. Notice that using $\lambda=0$ nullifies the effect of \ours,  making the method equivalent to \wil; for comparison, \wil's mIoU performance 
for the \textbf{disjoint} setting is $36.4$ and $20.8$ 
(see Tab.~\ref{tab:voc-disjoint-sota}, bottom-right) and
for the \textbf{overlap} setting is $38.7$ and $22.4$ (see Tab.~\ref{tab:voc-overlap-sota}, bottom-right) for old and new classes, respectively. In both cases, significantly below performance of \ours, regardless of the hyperparameters selected.

Please note that annotated test sets of \voc and COCO are not available. Thus, all the performances are reported on the validation set. Because of this reason, it is hard to tune the hyperparameters without overfitting the validation set. For this reason, we did not spend computational resources into hyperparameter validation and based our decisions on the aforementioned heuristics.

\begin{table}[h]
\centering
\begin{tabular}{lccc}\\
   \toprule  
     \multirow{2}{*}{\makecell{Semantic\\Similarity}} & \multicolumn{3}{c}{\bf 10-2 VOC} \\
    \cline{2-4}
     & 1-10 & 11-20 & All \\
    \midrule
    WordNet & 47.6 & 27.9 & 39.7 \\
     GloVe & 43.1 & 26.8 & 37.2 \\
     BERT & 44.5 & 28.4 & 38.6 \\
     \midrule 
     WILSON$\dagger$ & 38.7 &   22.4 & 32.5 \\
    \bottomrule
\end{tabular}
\caption{\textbf{Ablating semantic similarity} on VOC 10-2  \textit{multi-step} \textbf{overlap} incremental setting}\label{tab:sem-sim-type}

\end{table}

Next, we compare different semantic embedding methods for building the similarity between the semantic classes
defined in \cref{eqn:semantic-loss}. While by default we used BERT~\citep{DevlinNAACL19BERT} in our experiments,  we can also consider other alternatives such as 
GloVe~\citep{pennington2014glove} or a WordNet sub-tree. 
In the latter case, to compute the similarities
between two class, we used 1 over the number of hops (edges) between the two classes (nodes) in the sub-tree.  
As  we can see, 
using a different semantic embedding yields to relatively similar performance, with a slight drop when we use GLoVE, and  significant gain on old classes when we use the WordNet sub-tree. Still, all three methods outperform \wil: this result further validates the idea that leveraging semantic similarity between old and new classes
can improve the localizer and, hence, the final model.

\subsection{Results on VOC using the disjoint protocol}

We report in Tab.~\ref{tab:voc-disjoint-sota} the \voc results in the \textbf{disjoint} settings. This table complements the analysis of the Tab.~\ref{tab:voc-overlap-sota}, which focused on the \textbf{overlap} setting. We can draw similar conclusions:
\ours's improvements over WILSON$\dagger$ increase as we increase the number of tasks.

\begin{table}[!h]
    \centering
    \small
    \setlength{\tabcolsep}{1pt}
    \renewcommand{\arraystretch}{.5}
    \begin{tabular}{llc|ccc|ccc}
        \toprule
        & \multirow{2}{*}{\bf{Method}} & \multirow{2}{*}{\bf{Supervision}} & \multicolumn{3}{c|}{\textbf{15-5} (2 tasks)} & \multicolumn{3}{c}{\textbf{10-10} (2 tasks)}  \\
        & & &  1-15 & 16-20 &All & 1-10 & 11-20 &All  \\
        \midrule
        & Joint & Pixel & 75.5 & 73.5 & 75.4 & 76.6 & 74.0 & 75.4 \\
        \midrule
        \multirow{13}{*}{\vspace{-16mm}\rotatebox{90}{{\textbf{single-step}}}} & FT & Pixel & 8.4 & 33.5 & 14.4 & 7.7 &60.8 &33.0 \\
        & LWF~\cite{LiECCV16LearningWithoutForgetting} & Pixel & 39.7 & 33.3 & 38.2 & 63.1 & 61.1 & 62.2 \\
        & ILT~\cite{michieli2019incremental} & Pixel & 31.5 & 25.1 & 30.0 & \underline{67.7} & \underline{61.3} & \underline{64.7} \\
        
        & PLOP~\cite{DouillardCVPR21PLOPLearningWithoutForgettingContinualSemSegm} & Pixel & 71.0 & 42.8 & 64.3 & 63.7 & 60.2 & 63.4 \\
        & SDR~\cite{michieli2021continual} & Pixel & \underline{73.5} & 47.3 & \underline{67.2} & 67.5 & 57.9 & 62.9\\
        & RECALL~\cite{MaracaniICCV21RECALLReplayBasedContinualLearningSemSegm} & Pixel & 69.2 & \underline{52.9} & 66.3 & 64.1 & 56.9 & 61.9 \\
        \cmidrule{2-9}
        & CAM~\cite{ZhouCVPR16LearningDeepFeaturesForDiscLoc} & Image & 67.5 & 25.5 & 57.8 & 64.8 & 41.2 & 54.2 \\
        & SEAM~\cite{wang2020self} & Image & 68.9 & 32.5 & 61.1 & 61.5 & 52.3 & 58.3\\
        & SS~\cite{AraslanovCVPR20SingleStageSemSegmFromImageLabels} & Image & 68.9 & 25.9 & 60.2 & 60.3 & 27.2 & 45.5 \\
        & EPS~\cite{lee2021railroad} & Image & 70.7 & 36.8 & 63.6 & 64.3 & 53.8 & 60.5 \\
        & WILSON~\cite{CermelliCVPR22IncrementalLearninginSemSegmfromImageLabels} & Image & 72.0 & 44.1 & 66.3 & 64.2 & \textbf{54.5} & \textbf{60.8}\\
        \cmidrule{2-9}
        & WILSON$\dagger$~\cite{CermelliCVPR22IncrementalLearninginSemSegmfromImageLabels} & Image & 75.8 & 45.2 & 69.3 & 63.7 & 51.1 & 59.0 \\
        
        & \ours (Ours) & Image & \makecell{\textbf{75.9} \\ (\up0.1\%)} & \makecell{\textbf{47.5}\\(\up5.1\%)} & \makecell{\textbf{69.9}\\(\up0.9\%)} & \makecell{\textbf{64.5}\\(\up1.3\%)} & \makecell{51.2\\(\up0.2\%)} & \makecell{59.4\\(\up0.7\%)} \\
        \midrule
        \midrule
        
        \multirow{5}{*}{\vspace{-8mm}\rotatebox{90}{{\textbf{multi-step}}}} & \multirow{2}{*}{ } & \multirow{2}{*}{ } & \multicolumn{3}{c}{\textbf{10-5} (3 tasks)} & \multicolumn{3}{c}{\textbf{10-2} (6 tasks)}  \\
        & & &  1-10 & 11-20 &All & 1-10 & 11-20 &All \\
        \cmidrule{2-9}
        & WILSON$\dagger$~\cite{CermelliCVPR22IncrementalLearninginSemSegmfromImageLabels} & Image & 58.6 & 45.3 & 53.6 & 36.4 & 20.8 & 30.6 \\
        & \ours (Ours) & Image &  \makecell{\textbf{60.5}\\(\up3.2\%)} & \makecell{\textbf{46.8}\\(\up3.3\%)} & \makecell{\textbf{55.3}\\(\up3.2\%)} & \makecell{\textbf{42.5}\\(\up16.8\%)} & \makecell{\textbf{26.2}\\(\up26.0\%)} & \makecell{\textbf{36.6}\\(\up19.6\%)}\\
        \bottomrule
    \end{tabular}
    \caption{The m-IoU (in \%) scores for both \textit{single-step} (top half) and \textit{multi-step} (bottom half)  \textbf{disjoint} incremental settings on the \voc. The best numbers for the pixel supervised and image supervised methods are highlighted in underline and bold, respectively 
    }
    \label{tab:voc-disjoint-sota}

\end{table}

Furthermore, we report in Tab.~\ref{tab:mem-voc-disjoint} the \voc results for the memory-based approaches detailed in Sec.~\ref{sec:rehearsal}, for the \textbf{disjoint} setting, to complement the analysis we provided in Tab.~\ref{tab:mem-voc-overlap}, which focused on the \textbf{overlap} setting.

\begin{table}[!h]
    \centering
    \small
    \resizebox{\textwidth}{!}{%
    \begin{tabular}{ll|c|ccc|ccc}
        \toprule
        
        & \multirow{2}{*}{\bf{Method}} & \multirow{2}{*}{\bf{Supervision}} & \multicolumn{3}{c|}{{\bf 15-1} (6 tasks)} & \multicolumn{3}{c}{{\bf 10-1} (11 tasks)}\\
        
        \cline{4-6} \cline{7-9}
        
        & & & 1-15 & 16-20 & All & 1-10 & 11-20 & All\\
        
        \midrule
        
        \multirow{4}{*}{\rotatebox{90}{{\scriptsize w/o memory}}} & ILT~\citep{michieli2019incremental} & Pixel & 6.7 & 1.2 & 5.4 & 14.1 & 0.6 & 7.5 \\
        
        & MiB~\citep{CermelliCVPR20ModelingBackgroundIncrementalLearningSemSegm} & Pixel & 46.2 & 12.9 & 37.9 & 14.9 & 9.5 & 12.3 \\
        
        \cline{2-9}
        
        & WILSON$\dagger$~\citep{CermelliCVPR22IncrementalLearninginSemSegmfromImageLabels} & Image & 0.0 & 1.4 & 0.4 & 0.0 & 0.2 & 0.1  \\
        
        & \ours (Ours) & Image & 16.2 & 1.8 & 12.4 & 1.3 & 1.0 & 1.1 \\
        
        \midrule
        
        \multirow{5}{*}{\rotatebox{90}{{\scriptsize w/ memory}}} & WILSON$\dagger$ + $\mathcal{M}$ & Image & 64.9 & 24.8 & 56.0 & 43.4 & 21.7 & 34.1 \\
        
        & \ours (Ours) + $\mathcal{M}$ & Image & 66.7 & 30.9 & 59.0 & 42.7 & 28.8 & 37.9 \\
        
        \cline{2-9}
        
        & WILSON$\dagger$ + $\mathcal{M}_\text{ext}$ & Image & 74.2 & 30.3 & 64.3 &  {\bf 62.0} & 33.9 & 49.5 \\
        
        & \ours (Ours) + $\mathcal{M}_\text{ext}$ & Image & {\bf 74.7} & {\bf 35.8} & {\bf 66.1} & 61.7 & {\bf 37.4} & {\bf 51.2}  \\
        
        \cline{2-9}
        
        & RECALL (Web)~\citep{MaracaniICCV21RECALLReplayBasedContinualLearningSemSegm} & Pixel & \underline{67.6} & \underline{49.2} & \underline{64.3} & \underline{62.3} & \underline{50.0} & \underline{57.8} \\
        \bottomrule
    \end{tabular}
    }
    \caption{\textbf{Effect of memory.} Results on single-class \textit{multi-step} \textbf{disjoint} incremental setting on \voc. 
    $\mathcal{M}$ and $\mathcal{M}_\text{ext}$ indicate memories of previously seen or external samples, respectively. The best numbers for the pixel supervised and image supervised methods are highlighted in underline and bold, respectively}
    \label{tab:mem-voc-disjoint}
\end{table}

\subsection{\ours performance over tasks}
\label{sec:app-gains}

\begin{figure*}[h]
    \centering
    \includegraphics[width=\textwidth]{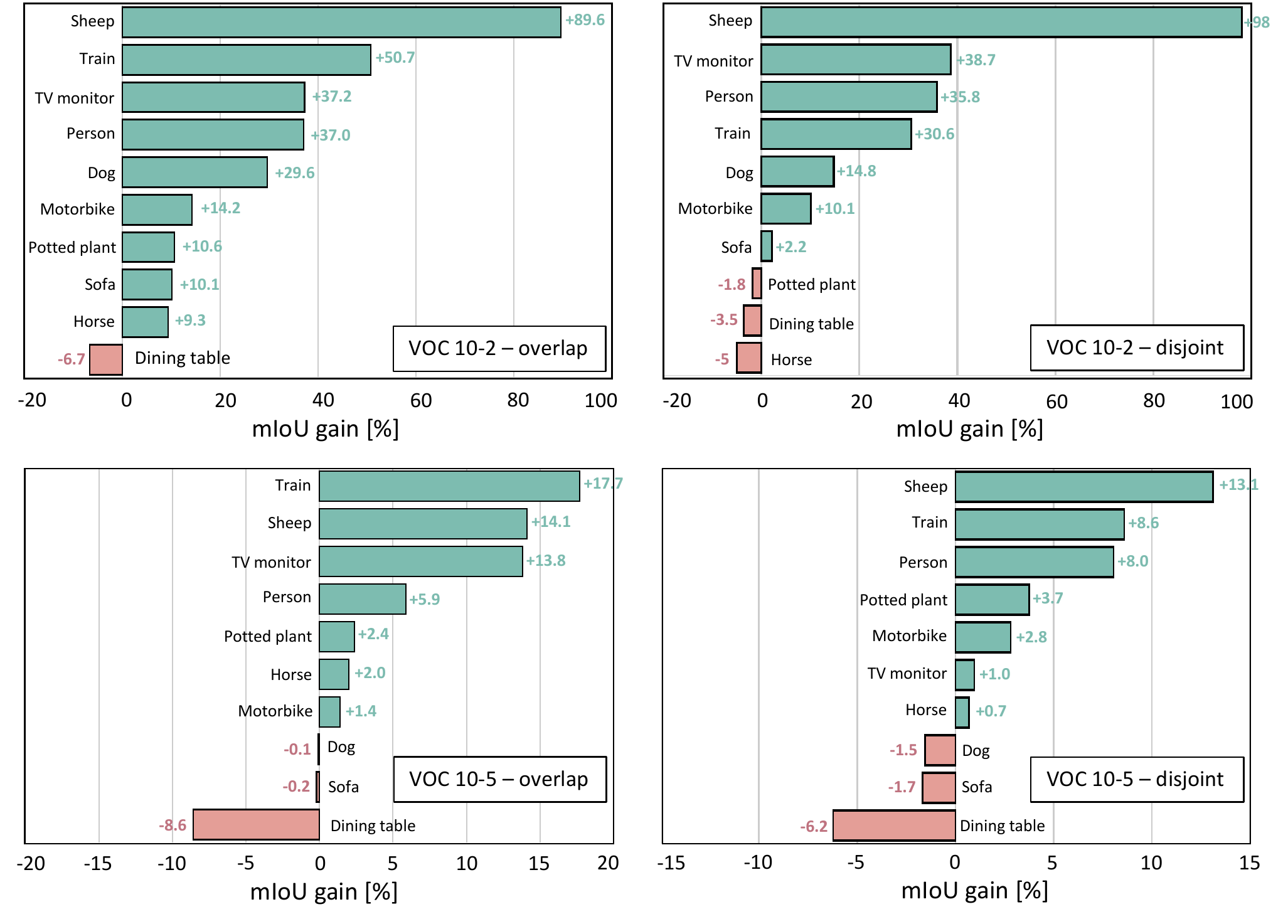}
  \caption{Per class gain/drop
  of \method \wrt WILSON, evaluated for each class in the step it was learned.
  Results computed on \voc. Top plots show \textbf{10-2} settings; bottom plots show \textbf{10-5} settings; leftmost plots show \textbf{overlap} settings; rightmost plots show \textbf{disjoint} settings. Note the different scales 
  }
\label{fig:app-plots}
\end{figure*}

The Fig.~\ref{fig:app-plots} extends the plot shown in Fig.~\ref{fig:main_plots} (right). We report \ours's gains \wrt \wil for different \voc settings.  
As expected, since \ours outperforms \wil more when the number of tasks is larger, the per-class gains are more evident for the 10-2 setting (top) than for the 10-5 one (bottom).

\begin{figure*}[h]
    \centering
    \includegraphics[width=\textwidth]{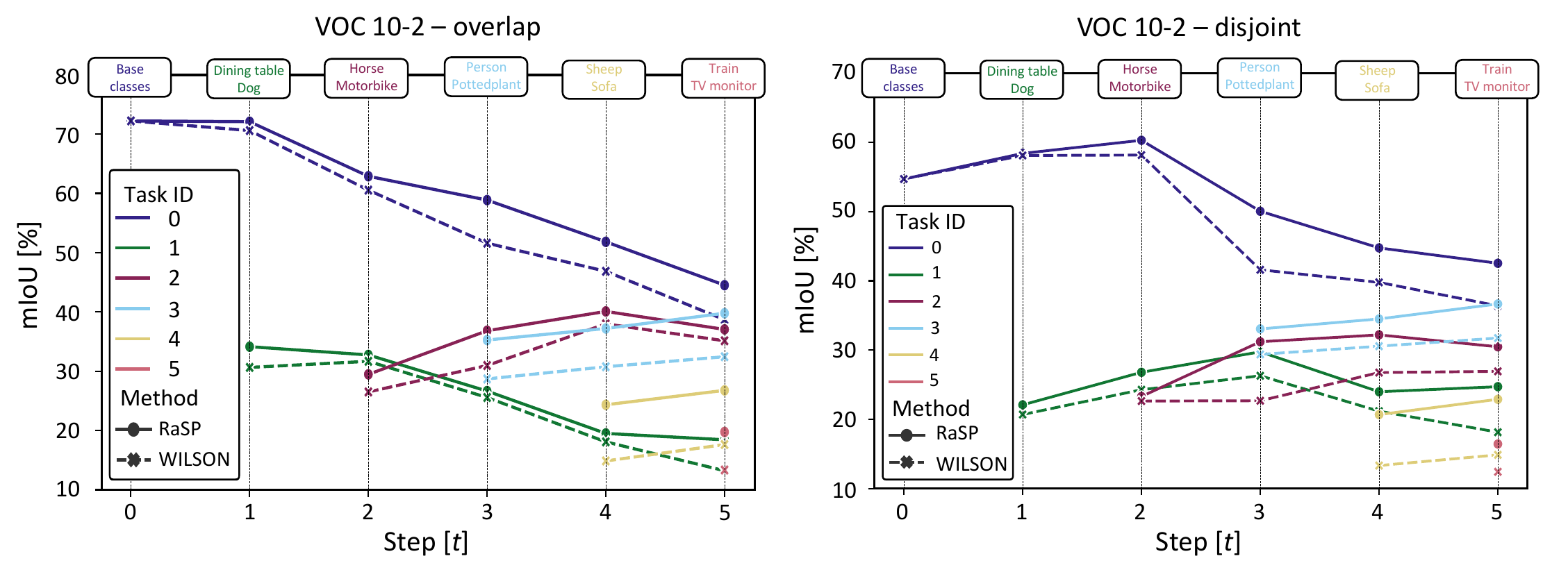}
  \caption{Per-task and per-step mIoU for the 10-2 VOC \textit{multi-step} incremental setting. Leftmost plot shows \textbf{overlap} results; rightmost plot shows \textbf{disjoint} results. Note the different scales}
\label{fig:app-gains}
\end{figure*}

The Fig.~\ref{fig:app-gains} extends the plot shown in Fig.~\ref{fig:main_plots} (left).
We report the evolution of the performance across sequence of tasks in the 10-2 \voc setting (6 tasks), for overlap and disjoint protocols (left and right, respectively).
For these plots, the conclusions made in the 
main paper still hold.

\subsection{Impact of class ordering}
\label{sec:app-class-ordering}

To demonstrate that our proposed semantic prior loss $\mathcal{L}_\text{RaSP}$ is versatile under different class ordering, we chose the 15-5 VOC disjoint setting, having 15 base classes and 5 novel classes, and randomized the old-novel classes splits. We ran experiments on four of such random splits and report the results in the Tab.~\ref{tab:class-ordering}. From the Tab.~\ref{tab:class-ordering} it is evident that RaSP outperforms WILSON on the four randomly chosen base-novel classes split, denoted by 15-5a, 15-5b, 15-5c and 15-5d of VOC, indicating that our improvements are consistently better on all of the class orderings. While the improvement by RaSP varies among the base-novel splits, yet most importantly they do not drop below WILSON. Thus we believe that our proposed method is well suited for real world applications where the classes will appear in random (and unknown) order and yet our incremental learner can perform better than its competitors.

\begin{table}[!h]
        \centering
        \small
        \setlength{\tabcolsep}{2pt}
        \begin{tabular}{l|ccc|ccc|ccc|ccc|ccc}
             \toprule
             Method & \multicolumn{15}{c}{\textbf{15-5} (2 tasks)}  \\
             & \multicolumn{3}{c|}{15-5a} & \multicolumn{3}{c|}{15-5b} & \multicolumn{3}{c|}{15-5c} & \multicolumn{3}{c|}{15-5d} & \multicolumn{3}{c}{Mean}\\
             \cline{2-16}
             & 1-15 & 16-20 & All & 1-15 & 16-20 & All & 1-15 & 16-20 & All & 1-15 & 16-20 & All & 1-15 & 16-20 & All \\
             \midrule
             WILSON$\dagger$ & 75.8 & 45.2 & 69.3 & 71.2 & 48.5 & 66.7 & 68.7 & 42.7 & 63.6 & 66.5 & 56.2 & 65.3 & 70.6 & 48.2 & 66.2\\
             RaSP & \textbf{75.9} & \textbf{47.5} & \textbf{69.9} & \textbf{71.8} & \textbf{53.3} & \textbf{68.4} & \textbf{70.8} & \textbf{44.5} & \textbf{65.5} & \textbf{66.7} & \textbf{57.8} & \textbf{65.9} & \textbf{71.3} & \textbf{50.8} & \textbf{67.4}\\
             \bottomrule
        \end{tabular}
        \caption{Comparison with the state-of-the-art on the 15-5 VOC disjoint incremental setting under different class orderings. The m-IoU (in \%) scores have been reported for the methods}
        \label{tab:class-ordering}
\end{table}

\subsection{Classwise performance}
\label{sec:app-classwise}

To get a complete understanding about the performance of each class, we report the classwise mIoU scores in a couple of settings of VOC for \ours and compare it with \wil. In details, we report the step-wise performance of both \wil and our proposed \ours for the single-step 15-5 VOC and the multi-step 10-2 VOC overlap settings in the Tab.~\ref{tab:classwise-15-5-ov} and Tab.~\ref{tab:classwise-10-2-ov}, respectively. In the Tab.~\ref{tab:classwise-15-5-ov} and Tab.~\ref{tab:classwise-10-2-ov} the incremental step (\ie, learning with weak labels) starts from step 1, with step 0 being the base training. The summarized versions of the Tab.~\ref{tab:classwise-15-5-ov} and Tab.~\ref{tab:classwise-10-2-ov} have been reported in the Tab.~\ref{tab:voc-overlap-sota} of the main paper.

\begin{table}[!h]
    \centering
    \resizebox{\textwidth}{!}{%
    \begin{tabular}{l|c|cccccccccccccccc|ccccc|ccc}
        \toprule
        & & \multicolumn{16}{c|}{Old Classes} & \multicolumn{5}{c|}{New Classes} & \multicolumn{3}{c}{Aggregate}\\
        Method & Step & \rotatebox{90}{bkg} & \rotatebox{90}{\oldclass{aeroplane}} & \rotatebox{90}{\oldclass{bicycle}} & \rotatebox{90}{\oldclass{bird}} & \rotatebox{90}{\oldclass{boat}} & \rotatebox{90}{\oldclass{bottle}} & \rotatebox{90}{\oldclass{bus}} & \rotatebox{90}{\oldclass{car}} & \rotatebox{90}{\oldclass{cat}} & \rotatebox{90}{\oldclass{chair}} & \rotatebox{90}{\oldclass{cow}} & \rotatebox{90}{\oldclass{dinningtable}} & \rotatebox{90}{\oldclass{dog}} & \rotatebox{90}{\oldclass{horse}} & \rotatebox{90}{\oldclass{motorbike}} & \rotatebox{90}{\oldclass{person}} & \rotatebox{90}{\newclassone{pottedplant}} & \rotatebox{90}{\newclassone{sheep}} & \rotatebox{90}{\newclassone{sofa}} & \rotatebox{90}{\newclassone{train}} & \rotatebox{90}{\newclassone{tv-monitor}} & \rotatebox{90}{\textbf{1-15}} & \rotatebox{90}{\textbf{16-20}} & \rotatebox{90}{\textbf{All}} \\
        \midrule
        
        WILSON$\dagger$ & 1 & 90.3 & 89.3 & \textbf{42.6} & 87.0 & 68.2 & \textbf{79.3} & 89.0 & 89.0 & \textbf{92.6} & \textbf{42.0} & \textbf{70.7} & \textbf{58.9} & \textbf{87.9} & \textbf{81.9} & 80.4 & \textbf{86.3} & 25.6 & 52.0 & \textbf{38.4} & 59.7 & 44.7 & \textbf{76.3} & 44.1 &  69.3 \\
        
        \ours & 1 & \textbf{91.4} & \textbf{89.8} & \textbf{42.6} & \textbf{87.5} & \textbf{65.8} & \textbf{79.3} & \textbf{89.5} & \textbf{89.1} & 92.0 & 41.3 & \textbf{70.7} & 58.7 & 87.7 & 81.8 & \textbf{81.7} & \textbf{86.3} & \textbf{26.5} & \textbf{54.6} & 36.8 & \textbf{70.5} & \textbf{46.5} & 76.2 & \textbf{47.0} & \textbf{70.0} \\
        \bottomrule
    \end{tabular}%
    }
    \caption{\textbf{Classwise results.} The mIoU (in \%) scores for the \textit{single-step} 15-5 (2 tasks) \textbf{overlap} incremental setting on VOC. The 15 old classes are denoted by \oldclass{green} and the 5 new classes are denoted in \newclassone{red}. The best numbers are highlighted in bold}
    \label{tab:classwise-15-5-ov}
\end{table}

From the Tab.~\ref{tab:classwise-15-5-ov} we observe that our \ours improves forward transfer by outperforming \wil in four out of the five new classes. In-line with our intuition, \ours's gain over \wil is noticeable in the new class ``train'' (by +10.8 absolute points) since ``train'' can be considered to have high visual similarity with the old class ``bus''. The gain in the other new classes (such as ``pottedplant'' or ``tv-monitor'') is slightly subdued due to the lack of closely resembling old classes. Nevertheless, in terms of new classes (\textbf{16-20}) and \textbf{All} aggregate performance \ours outperforms \wil.

\begin{table}[!h]
    \centering
    \resizebox{\textwidth}{!}{%
    \begin{tabular}{l|c|ccccccccccc|cc|cc|cc|cc|cc|ccc}
        \toprule
        & & \multicolumn{11}{c|}{Old Classes} & \multicolumn{10}{c|}{New Classes} & \multicolumn{3}{c}{Aggregate}\\
        
        Method & Step & \rotatebox{90}{bkg} & \rotatebox{90}{\oldclass{aeroplane}} & \rotatebox{90}{\oldclass{bicycle}} & \rotatebox{90}{\oldclass{bird}} & \rotatebox{90}{\oldclass{boat}} & \rotatebox{90}{\oldclass{bottle}} & \rotatebox{90}{\oldclass{bus}} & \rotatebox{90}{\oldclass{car}} & \rotatebox{90}{\oldclass{cat}} & \rotatebox{90}{\oldclass{chair}} & \rotatebox{90}{\oldclass{cow}} & \rotatebox{90}{\newclassone{dinningtable}} & \rotatebox{90}{\newclassone{dog}} & \rotatebox{90}{\newclasstwo{horse}} & \rotatebox{90}{\newclasstwo{motorbike}} & \rotatebox{90}{\newclassthree{person}} & \rotatebox{90}{\newclassthree{pottedplant}} & \rotatebox{90}{\newclassfour{sheep}} & \rotatebox{90}{\newclassfour{sofa}} & \rotatebox{90}{\newclassfive{train}} & \rotatebox{90}{\newclassfive{tv-monitor}} & \rotatebox{90}{\textbf{1-10}} & \rotatebox{90}{\textbf{11-20}} & \rotatebox{90}{\textbf{All}} \\
        \midrule
        
        \multirow{5}{*}{WILSON$\dagger$} & 1 & 90.6	& 86.5 & 41.3 & 81.4 & 67.4 & 82.8 & 87.8 & 81.7 & 85.3 & 35.1 & 56.4 & 30.7 & 30.6 &  &  &  &  &  &  &  &  & 70.6 & &  \\
        & 2 & 89.1 & 84.0 & 31.8 & 76.0 & 66.2 & 75.5 & 85.7 & 56.5 & 71.5 & 32.7 & 25.6 & 28.4 & 34.9 & 32.3 & 20.7 &  &  &  &  &  &  & 60.5 &  &  \\
        & 3 & 79.4 & 61.0 & 30.2 & 68.7 & 48.1 & 72.9 & 52.6 & 54.5 & 71.2 & 30.6 & 26.2 & 16.2 & 35.0 & 30.6 & 31.3 & 26.6 & 30.8 & & & & & 51.6 & \\
        & 4 & 74.6 & 49.6 &	27.6 & 56.9 & 57.5 & 62.8 &	65.2 & 57.4 & 59.2 & 6.2 &	26.0 & 0.0 & 36.2 &	36.7 & 39.3 & 29.6 & 32.0 &	20.1 &	9.7 & & & 46.8 \\
        
        & 5 & 72.6 & 37.9 & 25.8 & 59.5 & 48.9 & 58.7 & 48.0 & 30.1 & 57.8 & 5.1 & 15.1 & 0.0 & 26.4 & 35.5 & \textbf{34.8} & 32.2 & 32.7 & 21.7 & 13.6 & 21.4 & 5.3 & 38.7 & 22.4 & 32.5 \\
        
        \midrule
        
        \multirow{5}{*}{\ours} & 1 & 92.2 &	86.3 &	40.7 &	83.1 &	69.5 &	83.3 &	88.6 &	82.1 &	87.0 &	35.0 &	65.2 &	28.7 & 	39.6 & & & & & & & & & 72.1 \\
        & 2 & 91.3 &	83.9 &	34.2 &	77.2 &	68.3 &	77.8 &	86.0 &	58.0 &	68.0 &	30.5 &	44.8 &	24.9 &	40.6 &	35.3 &	23.7 & & & & & & & 62.9\\
        & 3 & 86.5 &	64.5 &	32.5 &	73.6 &	59.6 &	74.4 &	79.0 &	62.9 & 69.1 &	27.9 &	45.3 &	15.1 &	38.2 &	38.4 &	35.3 &	36.4 &	34.1 & & & & & 58.9 \\
        & 4 & 84.5 &	52.4 &	29.3 &	66.2 &	62.5 &	62.2 &	80.7 &	60.8 &	53.7 &	7.2 &	43.3 &	0.0 &	39.0 &	41.0 &	39.2 &	36.7 &	37.7 &	38.0 &	10.7 & & & 51.8 \\
        
        & 5 & \textbf{82.7} &	\textbf{44.2} &	\textbf{27.4} &	\textbf{67.1} &	\textbf{53.2} &	\textbf{58.8} &	\textbf{65.3} &	\textbf{34.5} &	\textbf{57.9} &	\textbf{6.5} &	\textbf{30.1} &	\textbf{0.1} &	\textbf{36.8} &	\textbf{39.6} &	34.5 &	\textbf{40.4} &	\textbf{39.2} &	\textbf{39.2} &	\textbf{14.4} &	\textbf{32.3} &	\textbf{7.3} &	\textbf{44.5} & \textbf{28.4} & \textbf{38.6} \\
         \bottomrule
    \end{tabular}%
    }
    \caption{\textbf{Classwise results.} The mIoU (in \%) scores for the multi-step 15-5 (6 tasks) \textbf{overlap} incremental setting on VOC. The 10 old classes are denoted by \oldclass{green} and the remainder new classes at consecutive steps are color coded as \{\newclassone{dinningtable, dog}\}; \{\newclasstwo{horse, motorbike}\}; \{\newclassthree{person, pottedplant}\}; \{\newclassfour{sheep, sofa}\}; and \{\newclassfive{train, tv-monitor}\}. The best numbers at the end of the final incremental step is highlighted in bold  }
    \label{tab:classwise-10-2-ov}
\end{table}

For the multi-step 10-2 VOC setting, reported in the Tab.~\ref{tab:classwise-10-2-ov}, the improvement of \ours over \wil  is even more stark compared to the single-step 15-5 VOC setting. In details, \ours outperforms in 20 out of the 21 classes in the Pascal-VOC benchmark, achieving greatly improved results in both the old (\textbf{1-10}) and the new classes (\textbf{11-20}). Careful scrutiny of the Tab.~\ref{tab:classwise-10-2-ov} reveals that the forward transfer offered by our \ours has a significant positive impact on the new classes such as ``dog'', ``horse'', ``sheep'' and ``train'', improving by +10.4, +4.1, +17.5 and +10.9 absolute points, respectively. Interestingly, the old classes suffer from lesser forgetting w.r.t \wil, with an aggregate improvement of +5.8 absolute points at the end of the final incremental step. We found that in incremental tasks where there are very few new classes (e.g., 2 new classes in the 10-2 VOC) \wil tends to overestimate the foreground (see Fig.~\ref{fig:qualitative_visualization} of the main and Fig. \ref{tab:qualitative_visualization2}), thereby forgetting more on the older classes. Contrarily, our \ours due to the semantic guidance for the foreground objects suffers less from the \textit{recency-bias}. This makes \ours better suited for the real-world incremental settings where the incremental learner will encounter tasks with very few new classes.

\subsection{Class-Incremental Few-shot Segmentation}
\label{sec:app-fss}

To push the limits of the WSCI task we also experiment on the weakly supervised few-shot class-incremental scenarios. Given the results on the few-shot settings greatly depend on the chosen few-shot image instances, we run the methods on four different folds of the \vocfs and \cocofs benchmarks. The Tab.~\ref{tab:ifss_full} is an extended version of Tab.~\ref{tab:ifss} with more pixel-supervised methods. The Tab.~\ref{tab:ifss-voc-ss-5shot}, Tab.~\ref{tab:ifss-voc-ss-2shot}, Tab.~\ref{tab:label} and Tab.~\ref{tab:ifss-coco-ss-2shot} show per-fold results for \voc (5-shot), \voc (2-shot), COCO (5-shot) and COCO (2-shot), respectively. In the per fold tables we only show the results for  the pixel-level supervised methods that performed best in average either on the base, new or the harmonic mean (HM score) of the base and the new classes (underlined in Tab.~\ref{tab:ifss_full}). We can observe from these tables that \ours is perfectly capable of operating in harder incremental scenarios when only few image labelled data are available for the new classes. Despite the overall lower performance of the image-label supervised methods, which is understandable, \ours can provide better and denser supervision on top of \wil.

\begin{table}[h]
    \centering
    \small
    \setlength{\tabcolsep}{0.pt}
    \resizebox{\textwidth}{!}{%
    \begin{tabular}{l|c|ccc|ccc|ccc|ccc}
    \toprule
         \multirow{2}{*}{Method} & \multirow{2}{*}{Supervision} & \multicolumn{3}{c|}{\bf{VOC (5-shot)}} & \multicolumn{3}{c|}{\bf{VOC (2-shot)}} & \multicolumn{3}{c|}{\bf{COCO (5-shot) }} & \multicolumn{3}{c}{\bf{COCO (2-shot)}}   \\
         
          & & 1-15 & 16-20 & HM & 1-15 & 16-20 & HM & 0-60 & 61-80 & HM & 0-60 & 61-80 & HM \\ 
         \midrule
         
         Fine-Tuning & Pixel & 
         55.8 & 29.6 & 38.7 &
         59.1 & 19.7 & 29.5 & 
         41.6 & 12.3 & 19.0 &
         41.5 & 7.3 & 12.4  \\
      \hline
         
         WI~\citep{QiCVPR18LowShotLearningWithImprintedWeights} & Pixel & 63.3 & 21.7 & 32.3 &
         63.3 & 19.2 & 29.5  &
         43.6 & 8.7 & 14.6 &
         44.2 & 7.9 & 13.5 \\
         
          DWI~\citep{GidarisCVPR18DynamicFewShotVisualLearningWithoutForgetting} & Pixel & 64.9 & 23.5 & 34.5  &
         \underline{64.8} & 19.8 & 30.4 &
         44.9 & 12.1 & 19.1 &
          45.0 & 9.4 & 15.6 \\
         
          RT~\citep{TianECCV20RethinkingFewShotImCatGoodEmbeddingIsAllYouNeed} & Pixel & 60.4 & 27.5 & 37.8  &
         60.9 & 21.6 & 31.9 & 
         46.9 & 13.7 & 21.2 &
         \underline{46.7} & 8.8 & 14.8 \\

         AMP~\cite{SiamICCV19AMPAdaptiveMaskedProxies4FewShotSegm} & Pixel	& 51.9 &	18.9 &	27.7  
         & 54.4 & 18.8 & 27.9 & 
         34.6 & 11.0& 16.7 & 
           35.7 & 8.8 & 14.2 \\
         
          SPN~\citep{XianCVPR19SemanticProjectionNetworkZeroAndFewLabelSS} & Pixel &	58.4 &	\underline{33.4} &	42.5 &
         60.8 &	26.3 & 36.7 & 
         43.7 & 15.6 & 22.9 &
        43.7 & 10.2 & 16.5  \\

          LWF~\citep{LiECCV16LearningWithoutForgetting} & Pixel &	59.7 &	30.9 &	40.8 &         63.6 & 18.9 & 29.2 &         44.6 & 12.9 & 20.1 &         44.3 & 7.1 & 12.3   \\
         
          ILT~\citep{michieli2019incremental} & Pixel &	61.4 &	32.0 &	42.1 &	          \underline{64.2} & 23.1 & 34.0 &         \underline{47.0} & 11.0 & 17.8 &         46.3 & 6.5 & 11.5    \\
         
          MiB~\citep{CermelliCVPR20ModelingBackgroundIncrementalLearningSemSegm} & Pixel & \underline{65.0} &	28.1 &	39.3 &	  
         63.5 & 12.7 & 21.1 & 
         44.7 & 11.9 & 18.8 &
         44.4 & 6.0 & 10.6  \\

          PIFS~\citep{CermelliBMVC21PrototypeBasedIncFewShotSemSegm}  & Pixel & 60.0 &	33.3 &	\underline{42.8} &
         60.5 & \underline{26.4} & \underline{36.8} & 
         42.8 & \underline{15.7} & \underline{23.0} &
         40.9 & \underline{11.1} & \underline{17.5} \\
         \hline
         
         WILSON$\dagger$~\citep{CermelliCVPR22IncrementalLearninginSemSegmfromImageLabels} & Image & 64.1 &	20.5 &	31.1 & 63.3 & 10.2 &	17.6 & 
         45.0 &	\bf{5.8} &	\bf{10.3} &
          \bf{43.6} &	1.9 &	3.6  \\
         
          \ours & Image &	\makecell{\bf{64.4}\\(\up0.5\%)} &	\makecell{\bf{21.3}\\(\up3.9\%)} &	\makecell{\bf{32.0}\\(\up2.9\%)} &	
         \makecell{\bf{63.5}\\(\up0.3\%)} & \makecell{\bf{10.7}\\(\up4.9\%)} & \makecell{\bf{18.3}\\(\up4.0\%)} & 
         \makecell{\bf{45.1}\\(\up0.2\%)} &	\makecell{5.6\\(\down3.4\%)} &	\makecell{10.0\\(\down2.9\%)} &	
         \makecell{43.5\\(\down0.2\%)} &	\makecell{\bf{2.0}\\(\up5.3\%)} &	\makecell{\bf{3.8}\\(\up5.6\%)}  \\
         \bottomrule
    \end{tabular}%
    }
    \caption{\textbf{Few-shot results.} The mIoU (in \%) scores for the \textit{single-step} (2 tasks) incremental few-shot SiS settings on the \vocfs and \cocofs benchmarks, for 5-shot and 2-shot cases. We show the average results over the 4 folds as in \citep{CermelliBMVC21PrototypeBasedIncFewShotSemSegm}.
    For each experiment, columns report performance on the base classes, new classes, and the Harmonic-Mean (HM) of the two scores. The best numbers for the pixel supervised and image-label  supervised methods are highlighted in underline and bold, respectively}
    \label{tab:ifss_full}
\end{table}

\begin{table}[!h]
    \centering
    \small
    \setlength{\tabcolsep}{0.pt}
    \resizebox{\textwidth}{!}{%
    \begin{tabular}{l|c|ccc|ccc|ccc|ccc}
    \toprule
         \multirow{2}{*}{Method} & \multirow{2}{*}{Supervision} &  
         \multicolumn{3}{c|}{\bf{Fold  5-0}} & \multicolumn{3}{c|}{\bf{Fold  5-1}} & \multicolumn{3}{c|}{\bf{Fold  5-2}} & \multicolumn{3}{c}{\bf{Fold  5-3}}  \\
         
          &
          & 1-15 & 16-20 & HM & 1-15 & 16-20 & HM & 1-15 & 16-20 & HM & 1-15 & 16-20 & HM \\ 
         \midrule
         
          FT & Pixel
         & 58.4 & 22.8 & 32.8 & 52.3 & 42.7 & 47.0 & 50.6 & 29.7 & 37.5 & 62.0 & 23.0 & 33.6 \\
         \hline
           
         SPN & Pixel
         &	63.3 &	28.2 &	39.0 &	53.4 &	43.7 &	48.1 &	54.5 &	33.5 &	41.5 &	62.3 &	28.2 &	38.8\\
           
         MiB & Pixel
         &	68.0 &	24.8 &	36.4 &	62.1 &	35.2 &	44.9 &	60.6 &	27.1 &	37.4 &	69.1 &	25.4 &	37.2 \\
         
         PIFS  & Pixel
         &	64.3 &	26.7 &	37.7 &	53.3 &	41.0 &	46.3 &	57.4 &	33.8 &	42.5 &	65.2 &	31.6 &	42.6 \\
         \hline
         
          WILSON$\dagger$ & Image
          &	66.6 &	18.8 &	29.3 &	\bf{60.2} &	22.5 &	32.8 &	61.1 &	21.3 &	31.6 &	68.5 &	19.2 &	30.0 \\
         
          \ours & Image 
          &	\makecell{\bf{66.9}\\(\up0.5\%)} &	\makecell{\bf{19.8}\\(\up5.3\%)} &	\makecell{\bf{30.6}\\(\up4.4\%)} &	\makecell{\bf{60.2}\\(0.0\%)} &	\makecell{\bf{23.0}\\(\up2.2\%)} &	\makecell{\bf{33.3}\\(\up1.5\%)} &	\makecell{\bf{61.4}\\(\up0.5\%)} &	\makecell{\bf{21.7}\\(\up1.9\%)} &	\makecell{\bf{32.1}\\(\up1.6\%)} &	\makecell{\bf{69.0}\\(\up0.7\%)} &	\makecell{\bf{20.5}\\(\up6.8\%)} &	\makecell{\bf{31.6}\\(\up5.3\%)}\\
         \bottomrule
    \end{tabular}%
    }
    \caption{\textbf{5-shot results per fold.} The m-IoU (in \%) scores for the \textit{single-step} (2 tasks) incremental few-shot (\textbf{5-shot}) SIS setting on the \vocfs benchmark.
    HM signifies the harmonic-mean of the base (0-15) and new classes (16-20) mIoU scores. The best numbers for image-label  supervised methods are highlighted in bold}
    \label{tab:ifss-voc-ss-5shot}
\end{table}

\vspace{5mm}

\begin{table}[!h]
    \centering
    \small
    \setlength{\tabcolsep}{0.pt}
    \resizebox{\textwidth}{!}{%
    \begin{tabular}{l|c|ccc|ccc|ccc|ccc}
    \toprule
         \multirow{2}{*}{Method} & \multirow{2}{*}{Supervision} & 
         \multicolumn{3}{c|}{\bf{Fold 5-0}} & \multicolumn{3}{c|}{\bf{Fold 5-1}} & \multicolumn{3}{c|}{\bf{Fold 5-2}} & \multicolumn{3}{c}{\bf{Fold  5-3}}  \\
         
         &
         & 0-15 & 16-20 & HM & 0-15 & 16-20 & HM & 0-15 & 16-20 & HM & 0-15 & 16-20 & HM \\ 
         \midrule
         
         FT & Pixel
         & 61.7 & 12.6 & 20.9 & 57.5 & 31.0 & 40.3 & 54.8 & 20.2 & 29.5 & 62.5 & 15.0 & 24.2 \\
         \hline
         
         DWI & Pixel
         & 68.2 & 15.1 & 24.7 & 60.4 & 30.9 & 40.9 & 60.4 & 17.2 & 26.8 & 70.1 & 16.2 & 26.3\\
         
         ILT & Pixel
         & 68.4 & 16.1	& 26.1 & 58.3 & 33.7 & 42.7 &	61.1 &	25.6 &	36.1 &	68.9 &	17.1 &	27.4 \\
         
         PIFS  & Pixel
         & 64.0 &	18.9 &	29.1 &	53.9 &	36.6 &	43.6 &	58.2 &	26.5 &	36.4 &	65.9 &	23.6 &	34.7 \\
         \hline
         
          WILSON$\dagger$ & Image
         &	\bf{65.7} &	7.7 & 13.8 &	\bf{60.6} &	\bf{14.7} &	\bf{23.7} &	60.0 &	9.4 & 16.3 &	66.8 &	9.0 & 15.9 \\
         
          \ours & Image  
         &	\makecell{\bf{65.7}\\(0.0\%)} &	\makecell{\bf{8.5}\\(\up10.4\%)} & \makecell{\bf{15.1}\\(\up9.4\%)} & \makecell{\bf{60.6}\\(0.0\%)} &	\makecell{14.0\\(\down4.8\%)} &	\makecell{22.7\\(\down4.2\%)} &	\makecell{\bf{60.5}\\(\up0.8\%)} &	\makecell{\bf{9.8}\\(\up4.3\%)} & \makecell{\bf{16.9}\\(\up3.7\%)} & \makecell{\bf{67.1}\\(\up0.4\%)} &	\makecell{\bf{10.6}\\(\up17.8\%)} & \makecell{\bf{18.3}\\(\up15.1\%)} \\
         \bottomrule
    \end{tabular}%
    }
    \caption{\textbf{2-shot results per fold.} The m-IoU (in \%) scores for the \textit{single-step} (2 tasks) incremental few-shot (\textbf{2-shot}) SIS setting on the \vocfs benchmark.
    HM signifies the harmonic-mean of the base (0-15) and new classes (16-20) mIoU scores. The best numbers for image-label  supervised methods are highlighted in bold}
    \label{tab:ifss-voc-ss-2shot}
\end{table}

\vspace{5mm}

\begin{table}[!h]
    \centering
    \small
    \setlength{\tabcolsep}{0.pt}
    \resizebox{\textwidth}{!}{%
    \begin{tabular}{l|c|ccc|ccc|ccc|ccc}
    \toprule
         \multirow{2}{*}{Method} & \multirow{2}{*}{Supervision} 
         & \multicolumn{3}{c|}{{\bf Fold 20-0}} & \multicolumn{3}{c|}{{\bf Fold 20-1}} & \multicolumn{3}{c|}{{\bf Fold 20-2}} & \multicolumn{3}{c}{{\bf Fold 20-3}}  \\
         
          &
          & 0-61 & 61-80 & HM & 0-61 & 61-80 & HM & 0-61 & 61-80 & HM & 0-61 & 61-80 & HM \\ 
         \midrule
         
          FT & Pixel
          & 37.3 & 7.6 & 12.6 & 40.9 & 15.0 & 22.0 & 45.3 & 13.7 & 21.0 & 43.0 & 12.9 & 19.8 \\
         \hline
         
         ILT & Pixel
         & 41.9 & 7.1 & 12.2 & 47.0 & 13.9 & 21.5 & 50.4 & 11.2 & 18.3 & 48.6 & 11.8 & 19.0 \\
         
          PIFS  & Pixel
          & 40.6 & 10.7 & 16.9 & 41.5 & 17.7 & 24.8 & 45.3 & 16.9 & 24.7 & 43.9 & 17.5 & 25.0\\
         \hline
         
         WILSON$\dagger$ & Image
         &	41.1 &	\bf{5.6} &	\bf{9.9} &	\bf{44.4} &	\bf{4.6} &	\bf{8.3} &	\bf{48.5} &	\bf{5.9} &	\bf{10.5} &	46.1 &	\bf{7.1} &	\bf{12.3} \\
         
          \ours & Image 
          &	\makecell{\bf{41.2}\\(\up0.2\%)} &	\makecell{5.5\\(\down0.2\%)} &	\makecell{9.7\\(\down2.0\%)} &	\makecell{\bf{44.4}\\(0.0\%)} &	\makecell{4.3\\(\down6.5\%)} &	\makecell{7.8\\(\down6.0\%)} &	\makecell{48.3\\(\down0.4\%)} &	\makecell{5.8\\(\down1.7\%)} &	\makecell{10.4\\(\down1.0\%)} &	\makecell{\bf{46.3}\\(\up0.4\%)} &	\makecell{6.9\\(\down2.8\%)} &	\makecell{12.0\\(\down2.4\%)} \\
         \bottomrule
         
    \end{tabular}%
    }
    \caption{\textbf{5-shot results per fold.} The m-IoU (in \%) scores for the \textit{single-step} (2 tasks) incremental few-shot (\textbf{5-shot}) SIS setting on the \cocofs benchmark.
    HM signifies the harmonic-mean of the base (0-60) and new classes (61-80) mIoU scores. The best numbers for image-label  supervised methods are highlighted in bold}
    \label{tab:label}
\end{table}

\vspace{5mm}

\begin{table}[!h]
    \centering
    \small
    \setlength{\tabcolsep}{0.1pt}
    \resizebox{\textwidth}{!}{%
    \begin{tabular}{l|c|ccc|ccc|ccc|ccc}
    \toprule
         \multirow{2}{*}{Method} & \multirow{2}{*}{Supervision} 
         & \multicolumn{3}{c|}{\bf{Fold 20-0}} & \multicolumn{3}{c|}{\bf{Fold 20-1}} & \multicolumn{3}{c|}{\bf{Fold 20-2}} & \multicolumn{3}{c}{\bf{Fold 20-3}}  \\
         
          &
          & 0-60 & 61-80 & HM & 0-60 & 61-80 & HM & 0-60 & 61-80 & HM & 0-60 & 61-80 & HM \\ 
         \midrule
         
          FT & Pixel
          & 37.4 & 4.2 & 7.6 & 40.3 & 9.0 & 14.7 & 45.4 & 7.7 & 13.2 & 43.1 & 8.4 & 14.0 \\
         \hline
         
         RT & Pixel
         & 40.6 & 5.5 & 9.7 & 46.8 & 10.5 & 17.2 & 50.8 & 8.1 & 14.0 & 48.5 & 11.1 & 18.1 \\
           
          PIFS  & Pixel
          & 38.6 & 6.8 & 11.6 & 39.4 & 13.1 & 19.7 & 43.5 & 11.4 & 18.1 & 42.2 & 13.1 & 20.0 \\
         \hline
         
          WILSON$\dagger$ & Image
          &	\bf{39.8} &	2.6 &	4.9 &	\bf{42.9} &	\bf{1.4} &	\bf{2.7} &	\bf{46.8} &	1.6 &	3.1 &	44.7 &	1.9 &	3.6 \\
          
         \ours & Image
         &	\makecell{39.7\\(\down0.3\%)} 
         &	\makecell{\bf{2.8}\\(\up7.7\%)} &	\makecell{\bf{5.2}\\(\up6.1\%)} &	\makecell{42.5\\(\down0.9\%)} &	\makecell{\bf{1.4}\\(0.0\%)} &	\makecell{\bf{2.7}\\(0.0\%)} &	\makecell{\bf{46.8}\\(0.0\%)} &	\makecell{\bf{1.7}\\(\up6.3\%)} &	\makecell{\bf{3.3}\\(\up6.5\%)} &	\makecell{\bf{44.9}\\(\up0.5\%)} &	\makecell{\bf{2.1}\\(\up10.5\%)} &	\makecell{\bf{4.0}\\(\up11.1\%)} \\
         \bottomrule
         
    \end{tabular}%
    }
    \caption{\textbf{2-shot results per fold.} The m-IoU (in \%) scores for the \textit{single-step} (2 tasks) incremental few-shot (\textbf{2-shot}) SIS setting on the \cocofs benchmark. HM signifies the harmonic-mean of the base (0-60) and new classes (61-80) mIoU scores. The best numbers for image-label  supervised methods are highlighted in bold}
    \label{tab:ifss-coco-ss-2shot}
\end{table}

\section{Additional details about WILSON}
\label{sec:app-wilson-details}

\subsection{Knowledge distillation losses}
\label{sec:app-wilson-kd}

Here we detail the two knowledge distillation losses used by \wil  and \ours. 
The first one, $\calL_{\text{KDE}}$, -- denoted by  $l_{\text{ENC}}$ in \cite{CermelliCVPR22IncrementalLearninginSemSegmfromImageLabels} -- computes the mean-squared error between the features extracted by the current encoder $E^t$ and those extracted by the old one $E^{t-1}$:
\begin{equation}
\label{eqn:encoder_loss-loss}
    \calL_{\text{KDE}}(\bfx) = \frac{1}{|\calI|} \displaystyle \sum_{i \in \calI} \displaystyle \| e^t_i -  e^{t-1}_i \|, 
\end{equation}
where  $e^{t-1}_i$ and $e^t_i$ are the feature vectors of  the pixel $i$ in the feature maps  $E^t(\bfx)$ and  $E^{t-1}(\bfx)$ respectively.

The second distillation loss $\calL_{\text{KDL}}$ -- denoted by  $l_{\text{LOC}}$ in \cite{CermelliCVPR22IncrementalLearninginSemSegmfromImageLabels} --
encourages consistency between the pixel-wise scores for old classes predicted by the localizer $(E \circ G)^t$ and those predicted by the old model $(E \circ F)^{t-1}$. It is carried out via the following binary cross-entropy loss:
\begin{equation}
\label{eqn:prior-loss}
    \calL_{\text{KDL}}(\bfz, \tilde{\bfy}) = - \frac{1}{|\calY^{t-1}||\calI|} \displaystyle \sum_{i \in \calI} \displaystyle \sum_{c \in \calY^{t-1}} \tilde{y}^c_i \log (\sigma(z^c_i)) + (1 - \tilde{y}^c_i) \log (1 - \sigma(z^c_i)).
\end{equation}

\subsection{Aggregating pixel-level scores}
\label{sec:app-ngwp}

In order to train the localizer with image-level labels,
normalized Global Weighted Pooling (nGWP)~\citep{AraslanovCVPR20SingleStageSemSegmFromImageLabels}  is used where  the channel-wise scores $\bfz$ are aggregated into a one-dimensional output vector $\hat{\bfy}_{\text{nGWP}} \in \R^{|\calY^{t}|}$ as follows:
\begin{align}
\label{eqn:ngwp}
y^c_{\text{nGWP}} = \frac{\sum_{i \in \calI} m^c_i z^c_i}{\epsilon + \sum_{i \in \calI} m^c_i},
\end{align}
with $\bfm=\softmax(\bfz)$
and $\epsilon$ is a small constant preventing  division by zero. Moreover, to penalize the localizer from predicting very small object masks, as in \cite{AraslanovCVPR20SingleStageSemSegmFromImageLabels},  the following  focal penalty term is added:
\begin{equation}
\label{eqn:focal-penalty}
   y^c_{\text{FOC}} = \left(1 - \frac{\sum_{i \in \calI} m^c_i}{|\calI|} \right)^{\gamma} \log \left(\lambda + \frac{\sum_{i \in \calI} m^c_i}{|\calI|} \right),
\end{equation}
where $\gamma$ and $\lambda$ are the hyperparameters. The final score from the localizer is then obtained by summing the scores from Eq.~(\ref{eqn:ngwp}) and Eq.~(\ref{eqn:focal-penalty}) namely  $\hat{\bfy} = \hat{\bfy}_{\text{nGWP}} + \hat{\bfy}_{\text{FOC}}$.

\subsection{The pseudo-supervision scores $\tilde{q}^c$}

The pixel level predictions of the localizer are combined 
 with the old model predictions to generate 
the pseudo-supervision scores $\tilde{q}^c$ as follows. First, the predicted binary segmentation maps  $\hat{\bfq}^{c}$ (hard assignments) are smoothed with the softmax scores:
\begin{align}
\label{eqn:soft-label}
    \bfq^c = \alpha \hat{\bfq}^{c^*} + (1 - \alpha) \bfm^c,
\end{align}
where  $\hat{q}^{c}_i=1$ if $c=\argmax_{k \in \calY^t} m^k_i$ and 0 otherwise. 

Then to get the final values to supervise the update of the segmentation module, for the new classes ($c \in \calC^t$)  the smoothed scores $\bfq^c$ from the localizer are considered, for the old classes the old model is trusted, while concerning the background the two outputs are combined. 
Concretely:
\begin{align}
\label{eqn:pseudo-label}
    \tilde{\bfq}^c = 
    \begin{cases}
        \text{min}(\tilde{\bfy}^c, \bfq^c) & \text{if } c = \text{`}bkg\text{'},\\
        \bfq^c & \text{if } c \in \calC^t, \\
        \tilde{\bfy}^c  & \text{otherwise},
    \end{cases}
\end{align}
where $\tilde{\bfy} = \sigma( (F \circ E)^{t-1}(\bfx))$.

\begin{table}[h]
    \centering
    \setlength{\tabcolsep}{1.7pt}
    \begin{tabular}{lcccccc}
         & \multicolumn{2}{c}{{\bf 15-5 VOC} (2 tasks)} & \multicolumn{2}{c}{{\bf 10-5 VOC} (3 tasks)} & \multicolumn{2}{c}{{\bf 10-2 VOC} (6 tasks)} \\
         \rotatebox{90}{\hspace{0.60cm}Input}& \includegraphics[width=0.15\columnwidth]{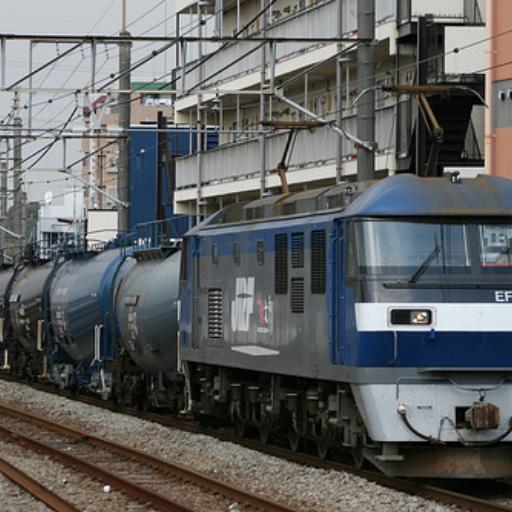} & \includegraphics[width=0.15\columnwidth]{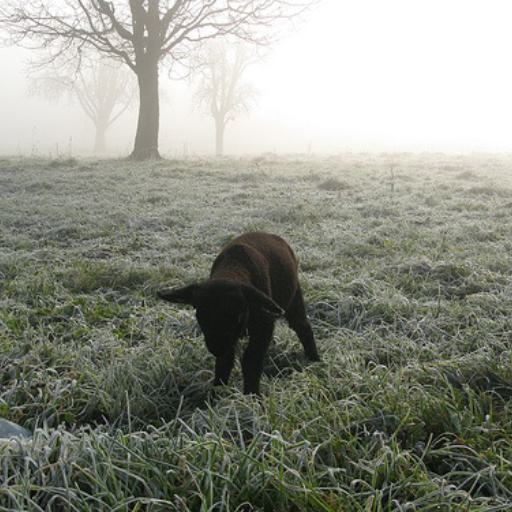} & \includegraphics[width=0.15\columnwidth]{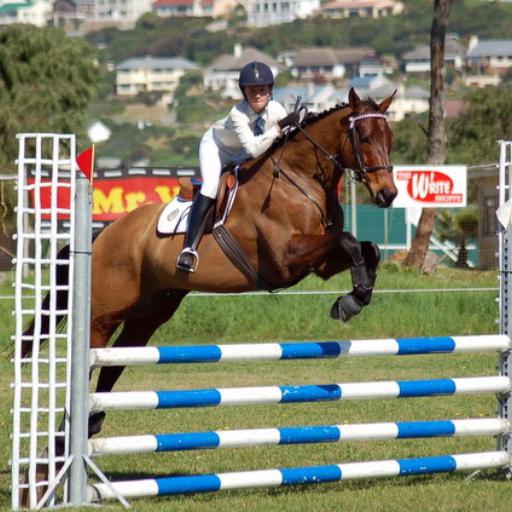} & \includegraphics[width=0.15\columnwidth]{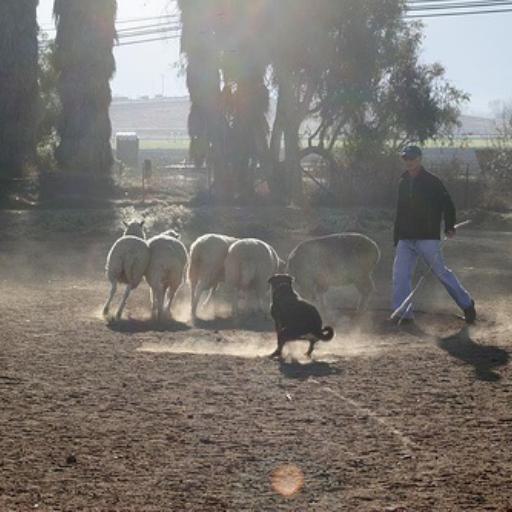} & \includegraphics[width=0.15\columnwidth]{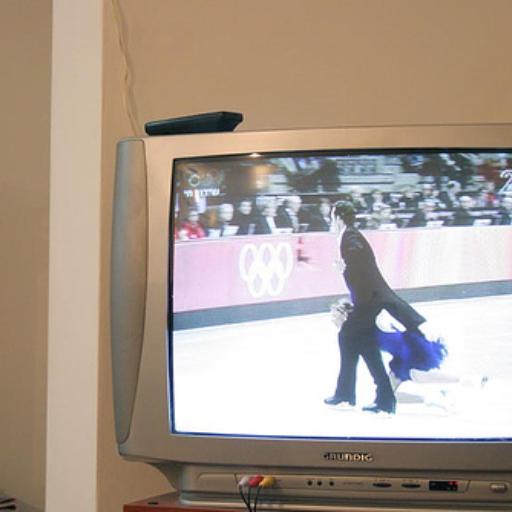} & \includegraphics[width=0.15\columnwidth]{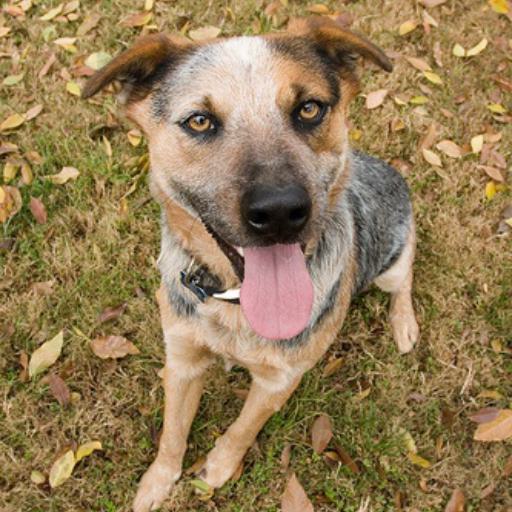}\\
         
         \rotatebox{90}{\hspace{0.35cm}WILSON}& \includegraphics[width=0.15\columnwidth]{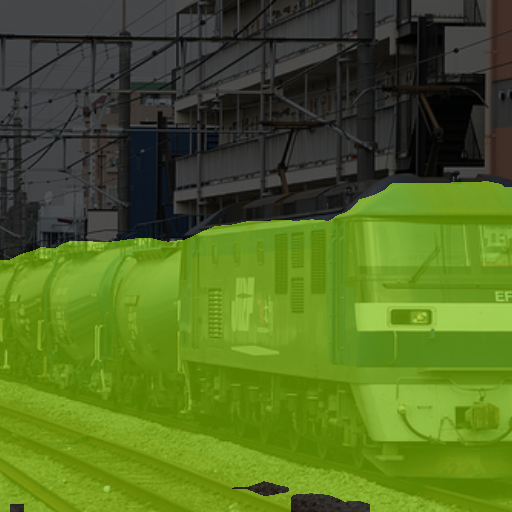} & \includegraphics[width=0.15\columnwidth]{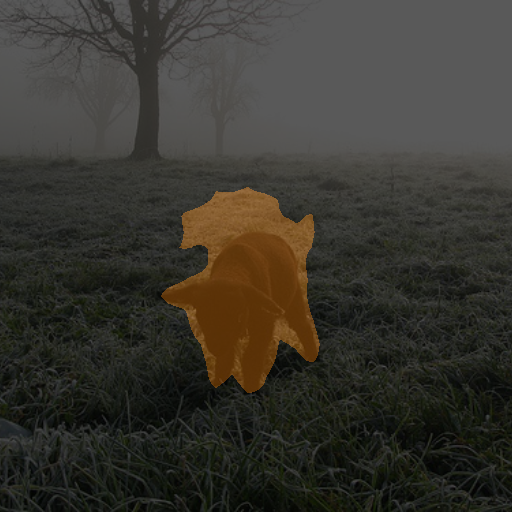} & \includegraphics[width=0.15\columnwidth]{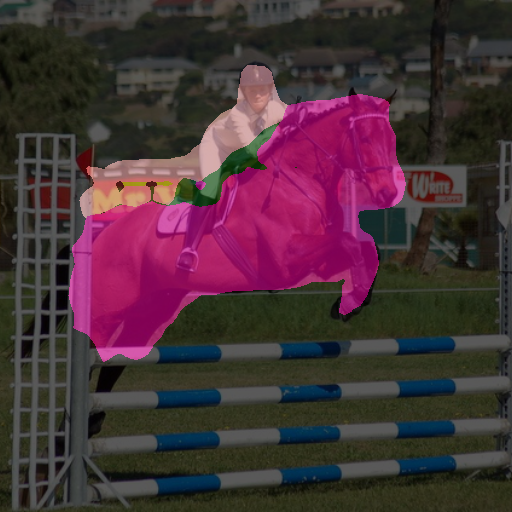} & \includegraphics[width=0.15\columnwidth]{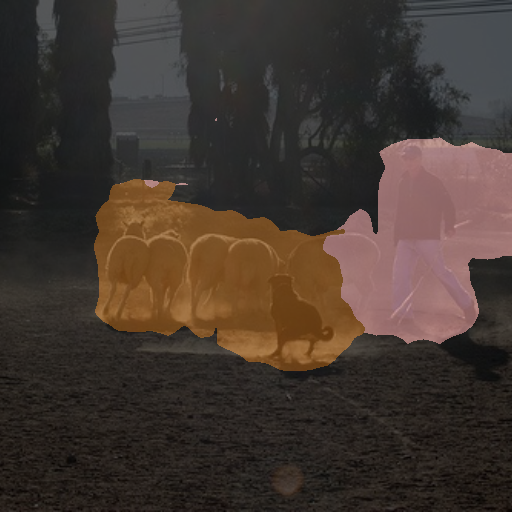} & \includegraphics[width=0.15\columnwidth]{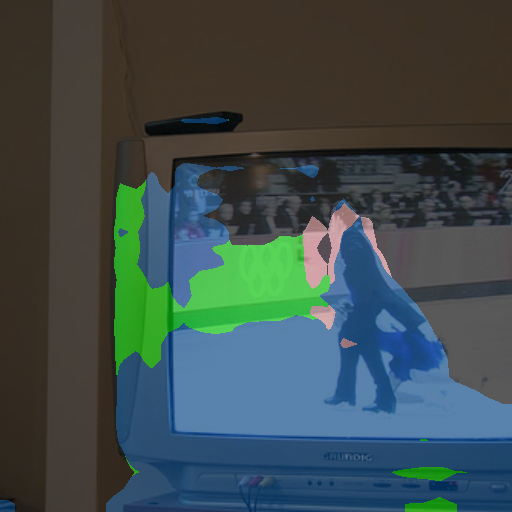} & \includegraphics[width=0.15\columnwidth]{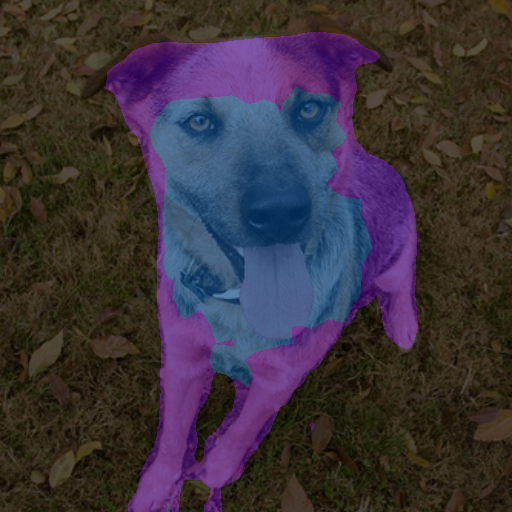}\\ 
         
         \rotatebox{90}{\hspace{0.15cm}\ours (Ours)}& \includegraphics[width=0.15\columnwidth]{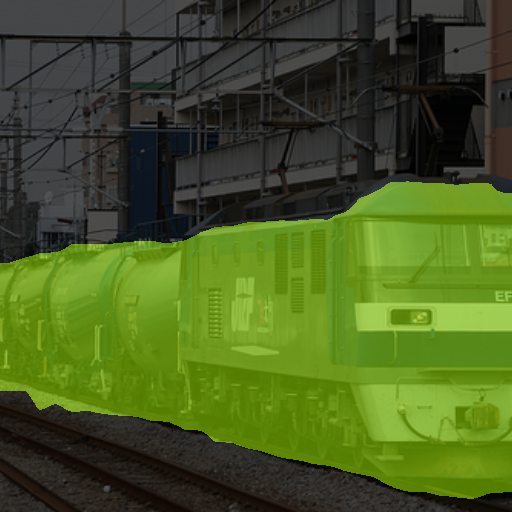} & \includegraphics[width=0.15\columnwidth]{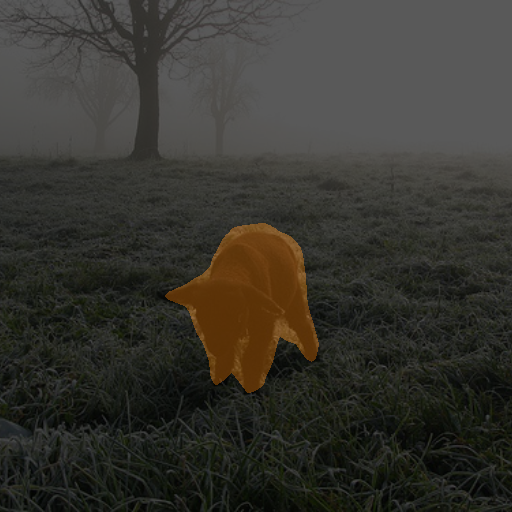} & \includegraphics[width=0.15\columnwidth]{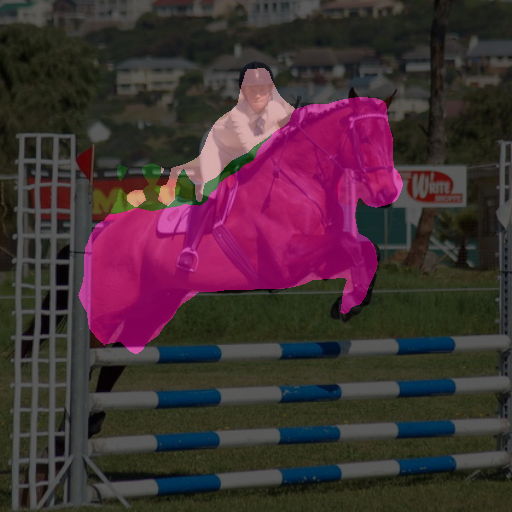} & \includegraphics[width=0.15\columnwidth]{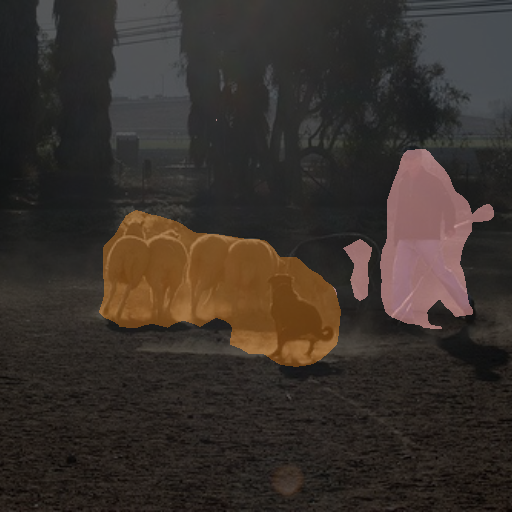} & \includegraphics[width=0.15\columnwidth]{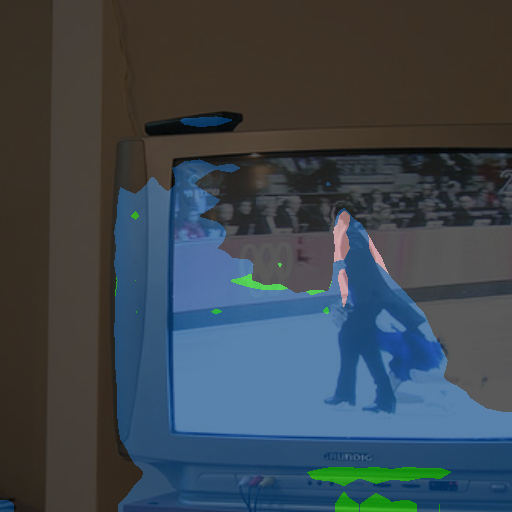} & \includegraphics[width=0.15\columnwidth]{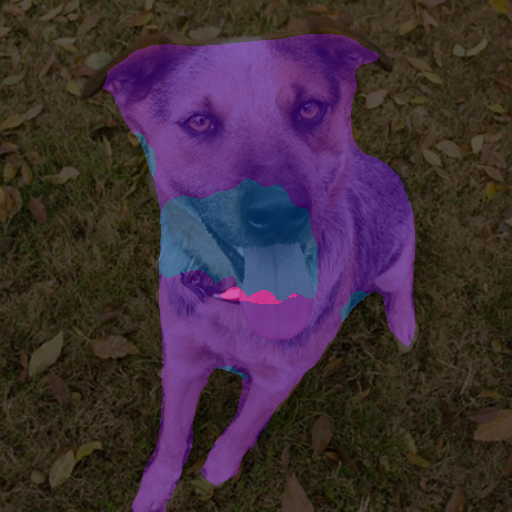}\\
         
         \rotatebox{90}{\hspace{0.05cm}Ground Truth}& \includegraphics[width=0.15\columnwidth]{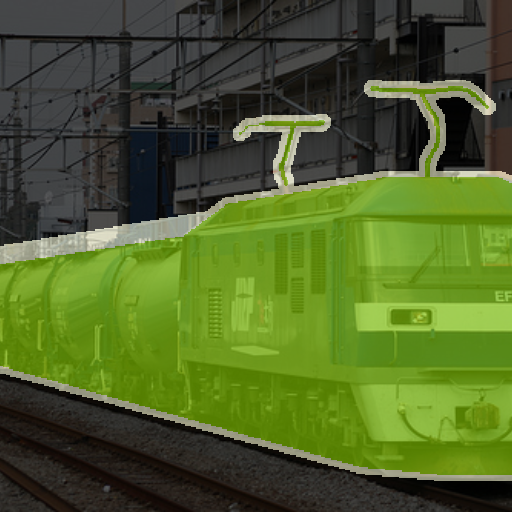} & \includegraphics[width=0.15\columnwidth]{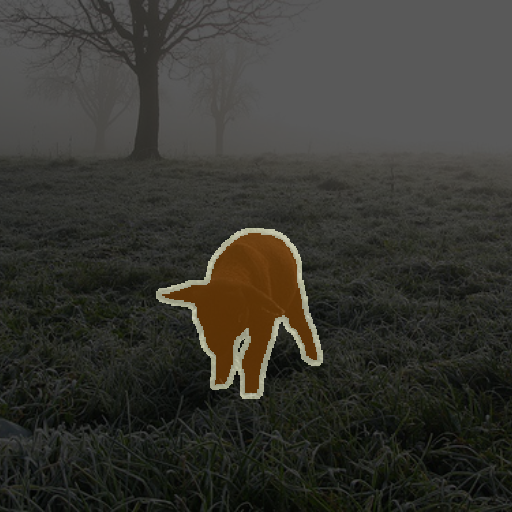} & \includegraphics[width=0.15\columnwidth]{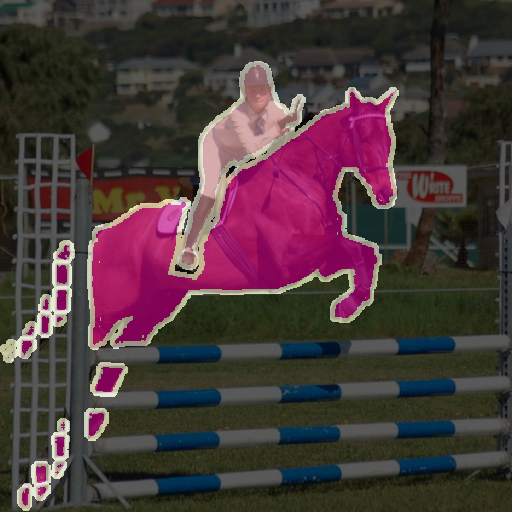} & \includegraphics[width=0.15\columnwidth]{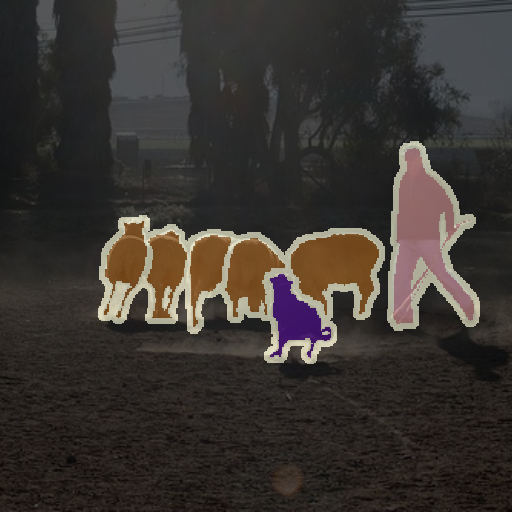} & \includegraphics[width=0.15\columnwidth]{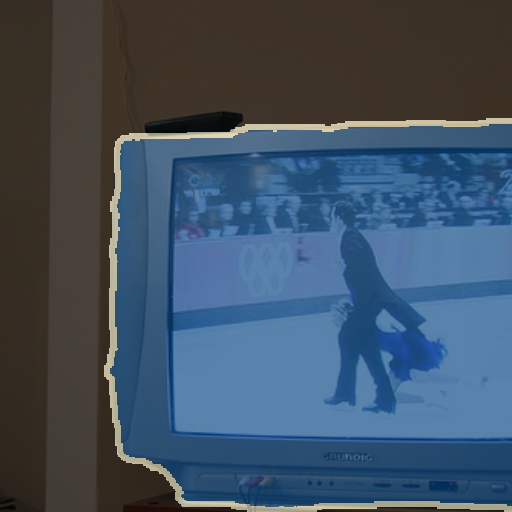} & \includegraphics[width=0.15\columnwidth]{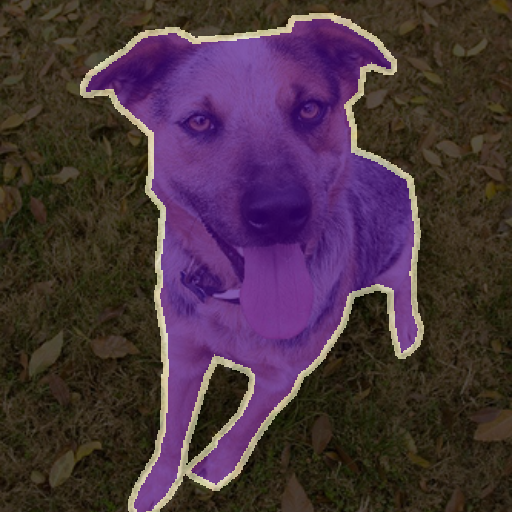}\\
         
    \end{tabular}
    
    \captionof{figure}{Qualitative  results from different  \textit{single-step}  and \textit{multi-step} \textbf{overlap} incremental settings on \voc. 
    The are from the final step of the corresponding settings
    }
    \label{tab:qualitative_visualization_app}
\end{table}

\section{Further qualitative results}
\label{sec:app-qual}

\begin{figure}[h!]
    \centering
    \setlength{\tabcolsep}{1.7pt}
    \begin{tabular}{cccccc}
     \small Input & \small GT & \small $(F\circ E)^{t-1}(\bfx_t)$   & \small $\bf{s}^{\bf{l}_t}$ &  \small \ours   & \small \wil \\

   \includegraphics[width=0.15\columnwidth]{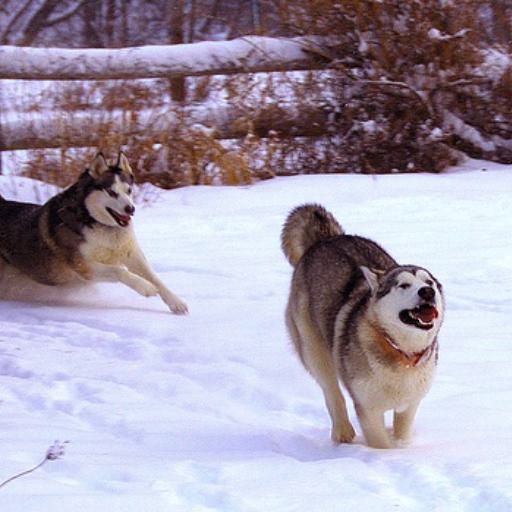} &
       \includegraphics[width=0.15\columnwidth]{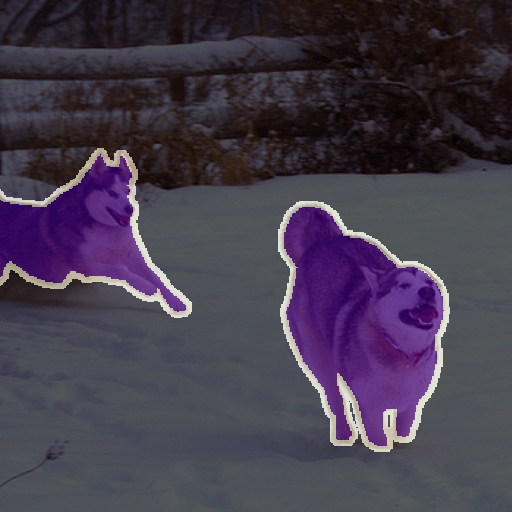} & 
	   \includegraphics[width=0.15\columnwidth]{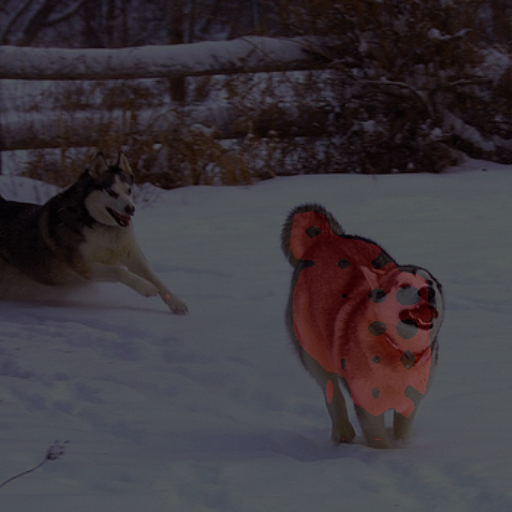} & 
	   \includegraphics[width=0.15\columnwidth]{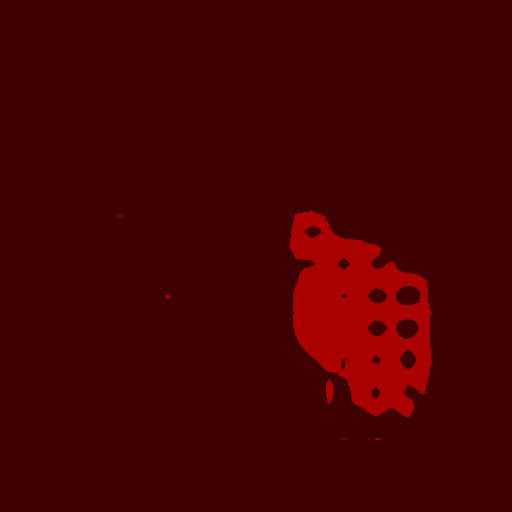} & 
	   \includegraphics[width=0.15\columnwidth]{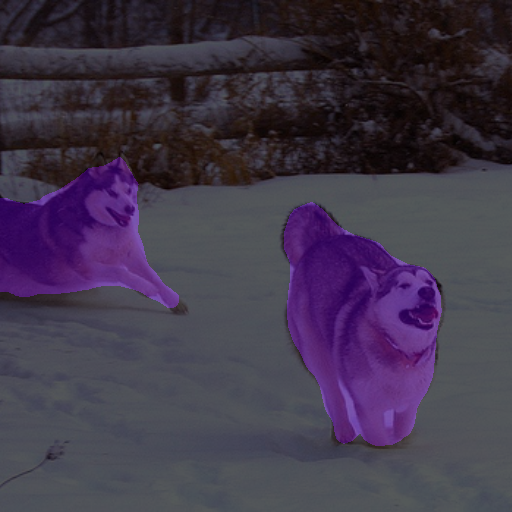} & 
	   \includegraphics[width=0.15\columnwidth]{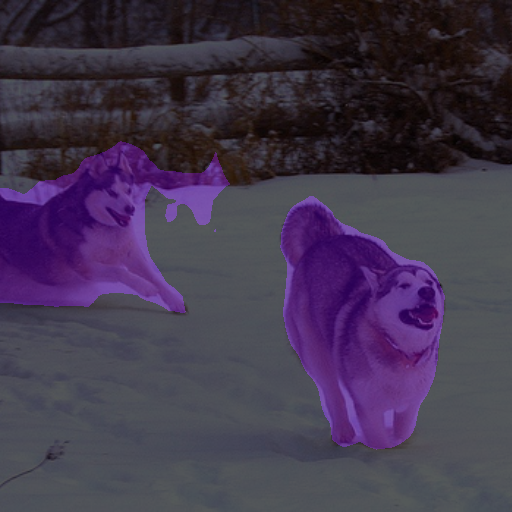} \\

      \includegraphics[width=0.15\columnwidth]{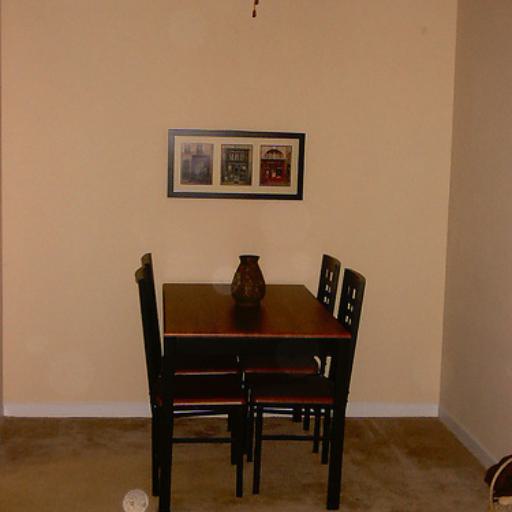} &
       \includegraphics[width=0.15\columnwidth]{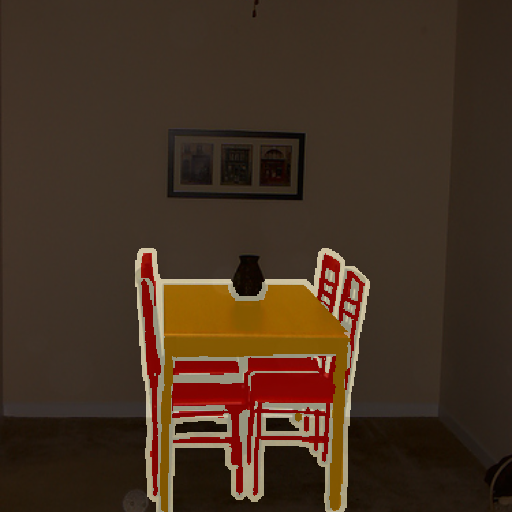} & 
	   \includegraphics[width=0.15\columnwidth]{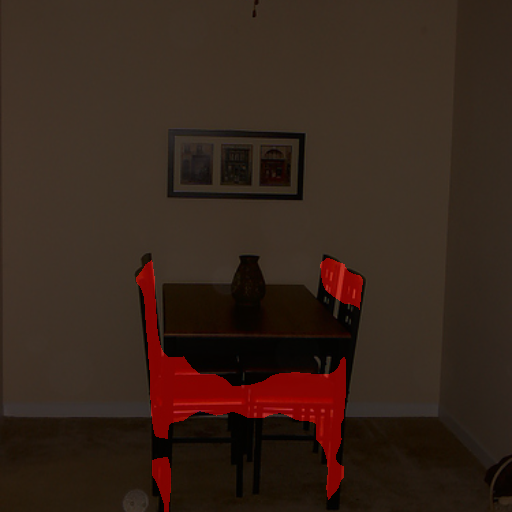} & 
	   \includegraphics[width=0.15\columnwidth]{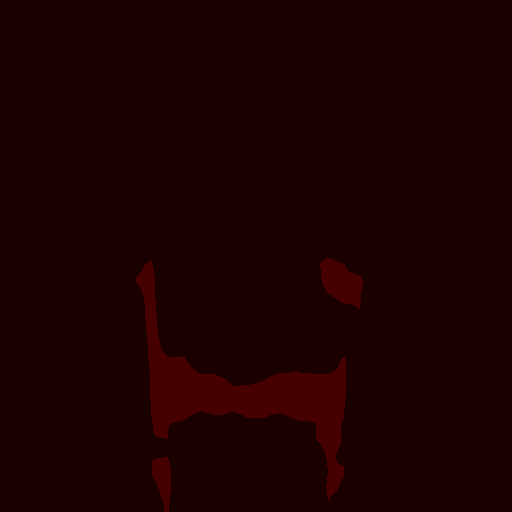} & 
	   \includegraphics[width=0.15\columnwidth]{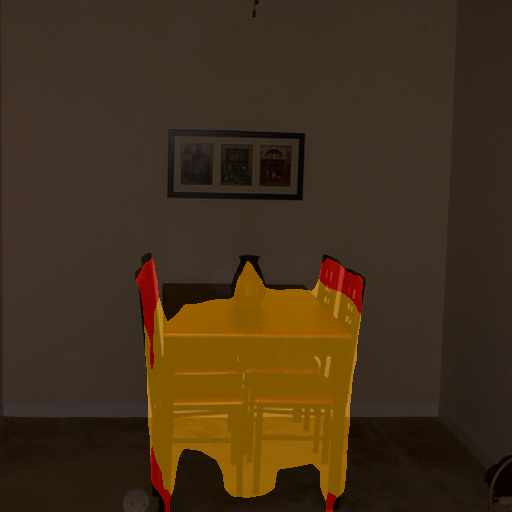} & 
	   \includegraphics[width=0.15\columnwidth]{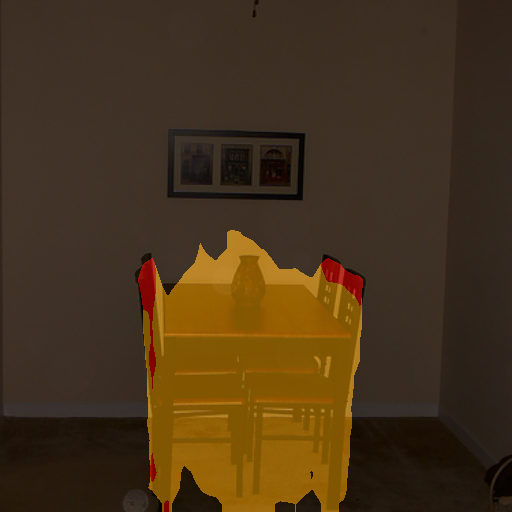} \\

        \includegraphics[width=0.15\columnwidth]{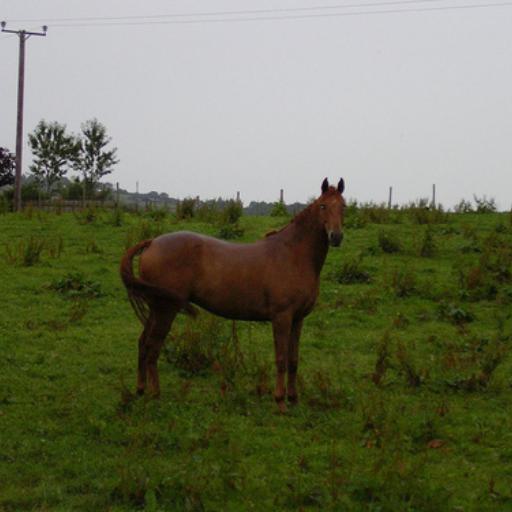} &
       \includegraphics[width=0.15\columnwidth]{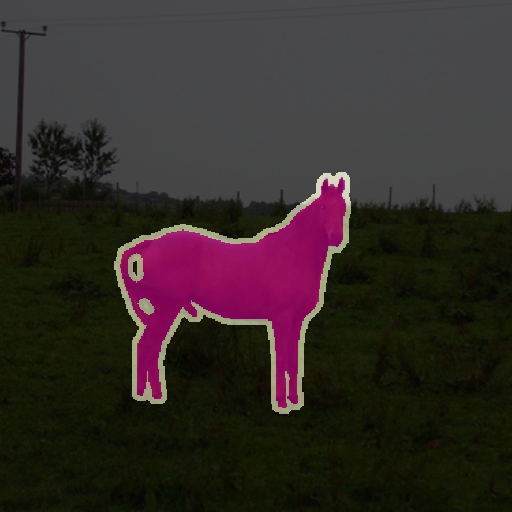} & 
	   \includegraphics[width=0.15\columnwidth]{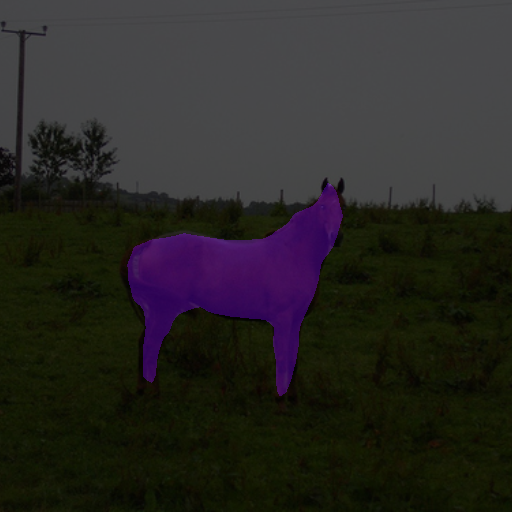} & 
	   \includegraphics[width=0.15\columnwidth]{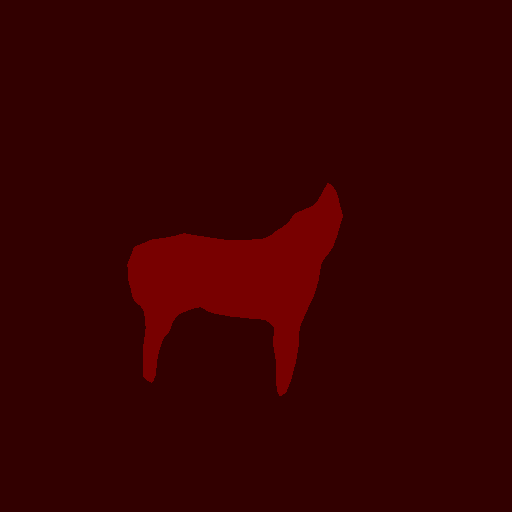} & 
	   \includegraphics[width=0.15\columnwidth]{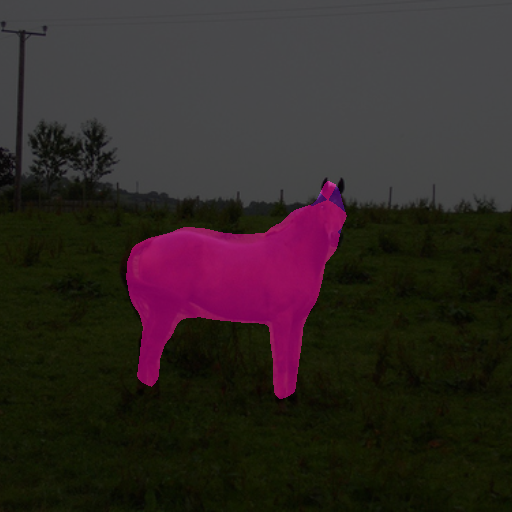} & 
	   \includegraphics[width=0.15\columnwidth]{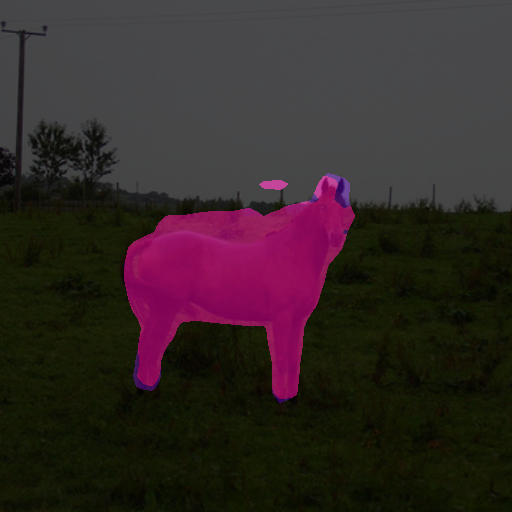} \\
    
      \includegraphics[width=0.15\columnwidth]{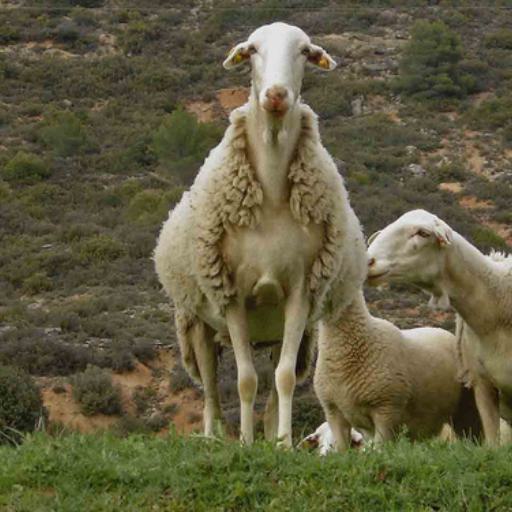} &
       \includegraphics[width=0.15\columnwidth]{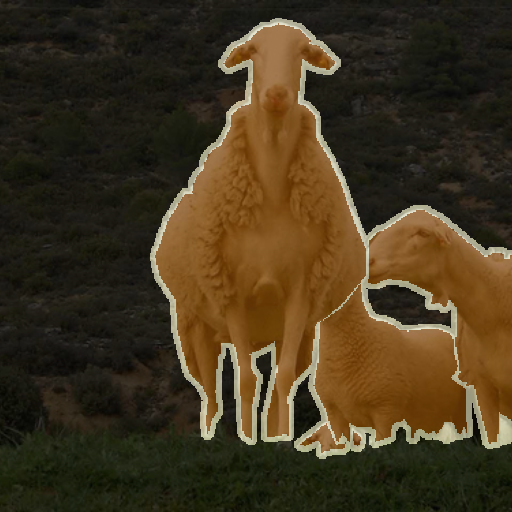} & 
	   \includegraphics[width=0.15\columnwidth]{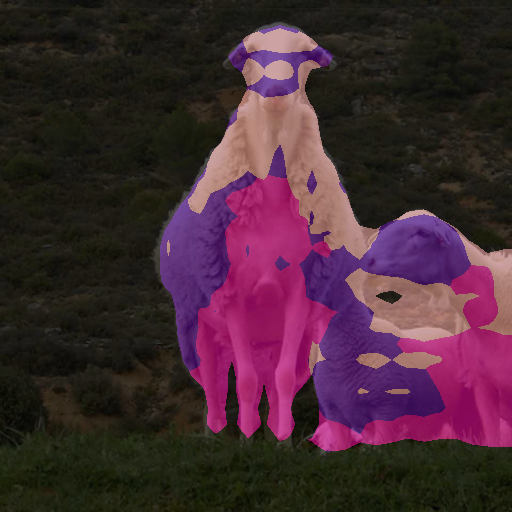} & 
	   \includegraphics[width=0.15\columnwidth]{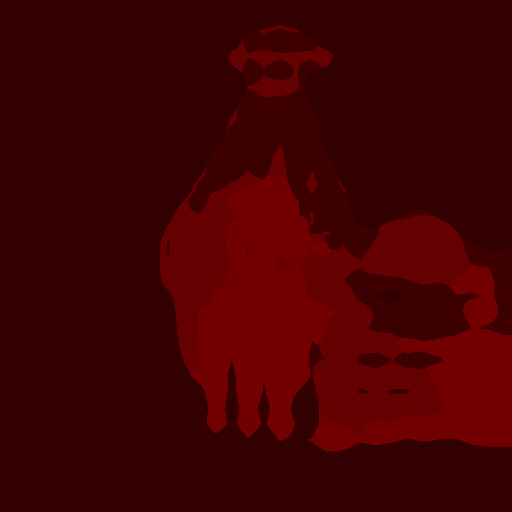} & 
	   \includegraphics[width=0.15\columnwidth]{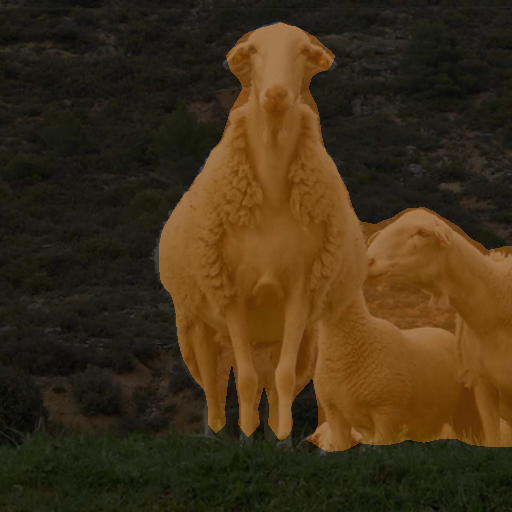} & 
	   \includegraphics[width=0.15\columnwidth]{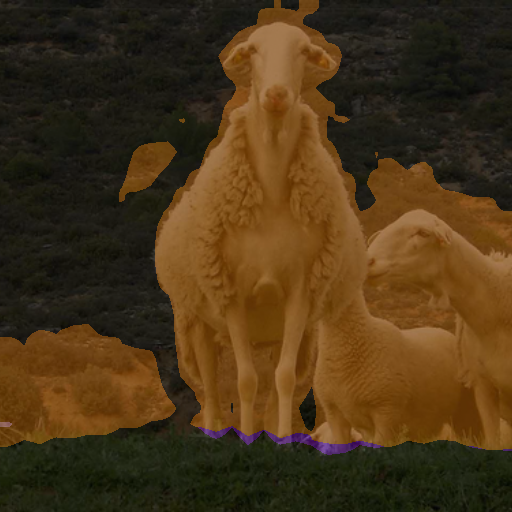} \\
	   
	 \includegraphics[width=0.15\columnwidth]{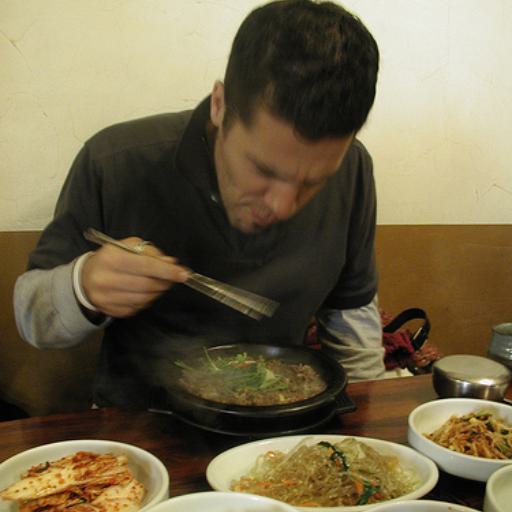} &
       \includegraphics[width=0.15\columnwidth]{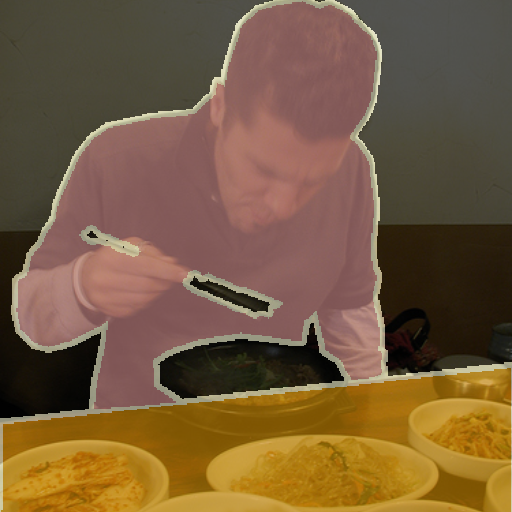} & 
	   \includegraphics[width=0.15\columnwidth]{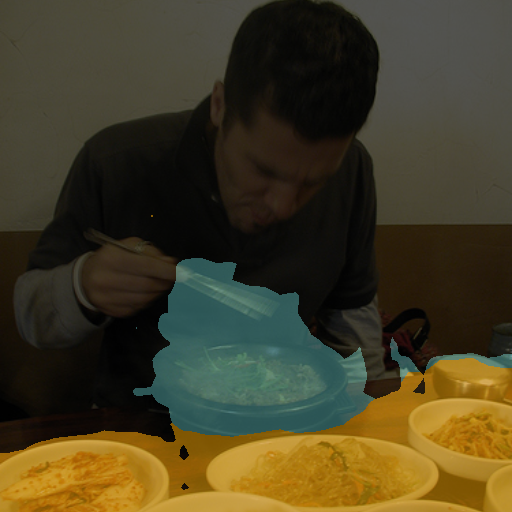} & 
	   \includegraphics[width=0.15\columnwidth]{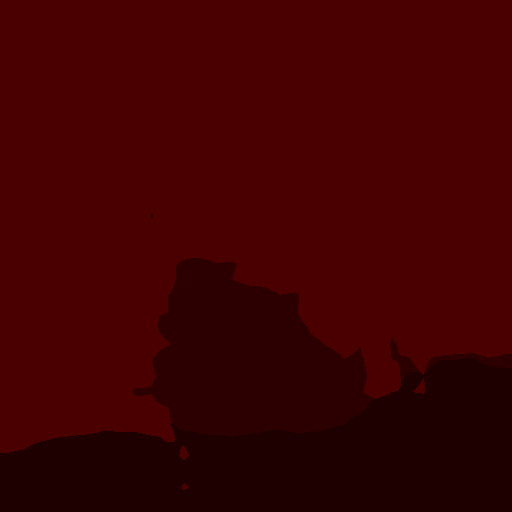} & 
	   \includegraphics[width=0.15\columnwidth]{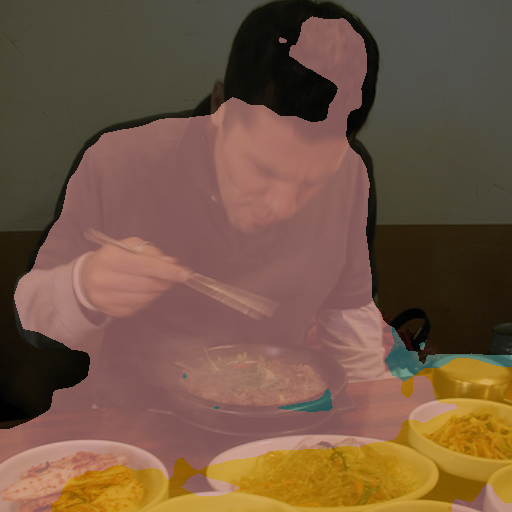} & 
	   \includegraphics[width=0.15\columnwidth]{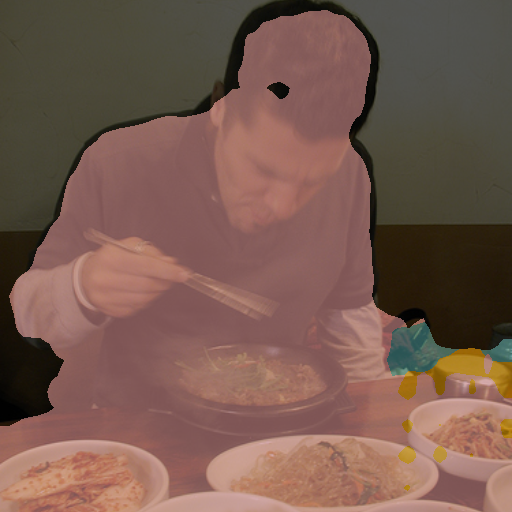} \\

       \includegraphics[width=0.15\columnwidth]{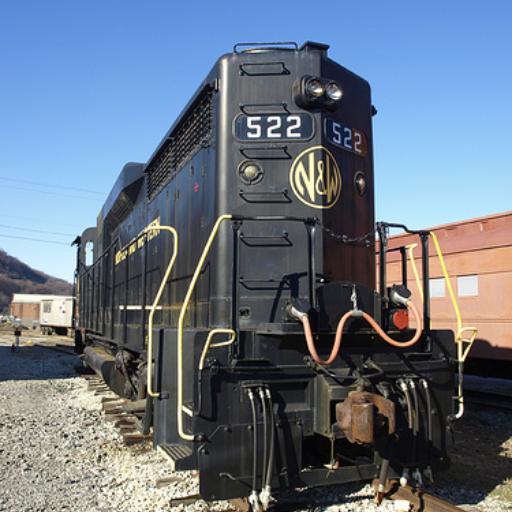} &
	   \includegraphics[width=0.15\columnwidth]{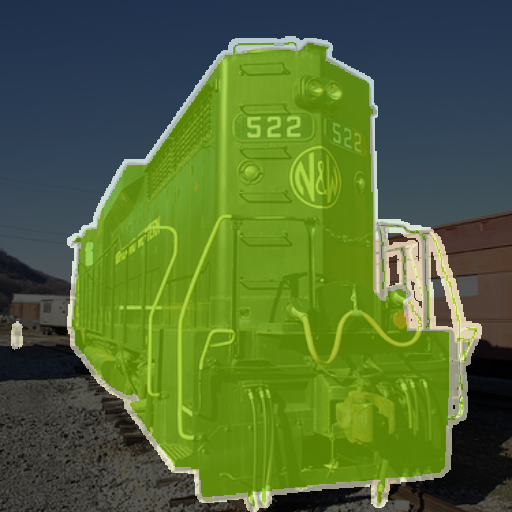} & 
	   \includegraphics[width=0.15\columnwidth]{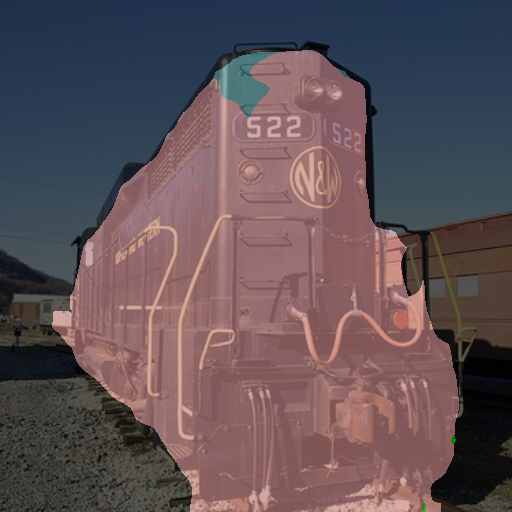} & 
	   \includegraphics[width=0.15\columnwidth]{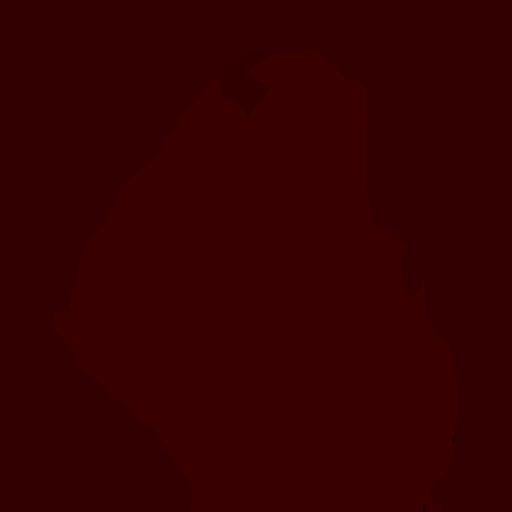} & 
	   \includegraphics[width=0.15\columnwidth]{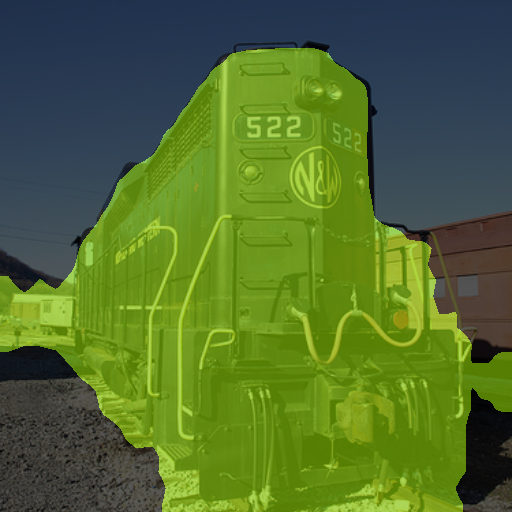} & 
	   \includegraphics[width=0.15\columnwidth]{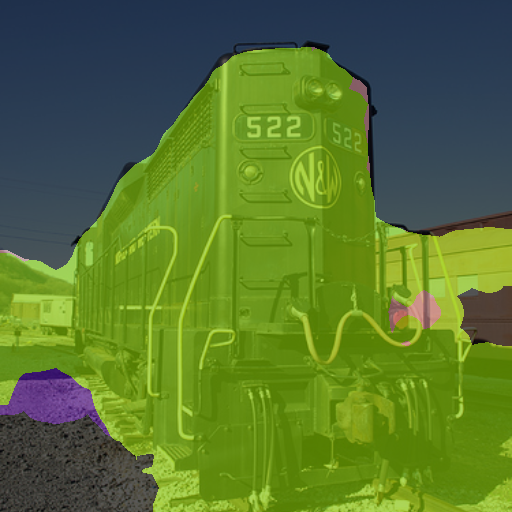} \\
       
       \includegraphics[width=0.15\columnwidth]{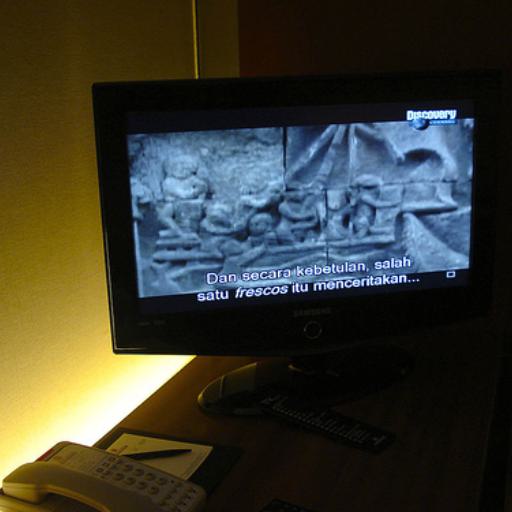} &
	   \includegraphics[width=0.15\columnwidth]{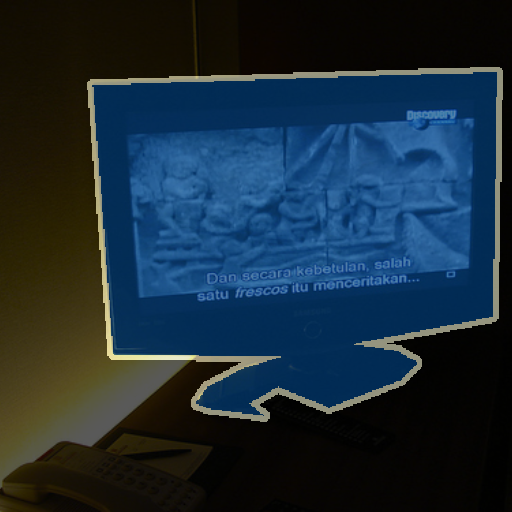} & 
	   \includegraphics[width=0.15\columnwidth]{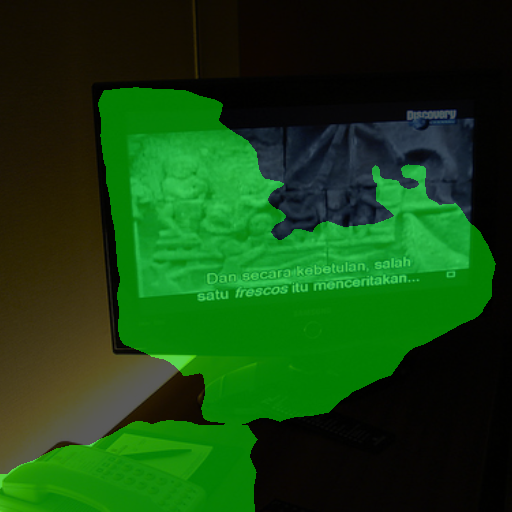} & 
	   \includegraphics[width=0.15\columnwidth]{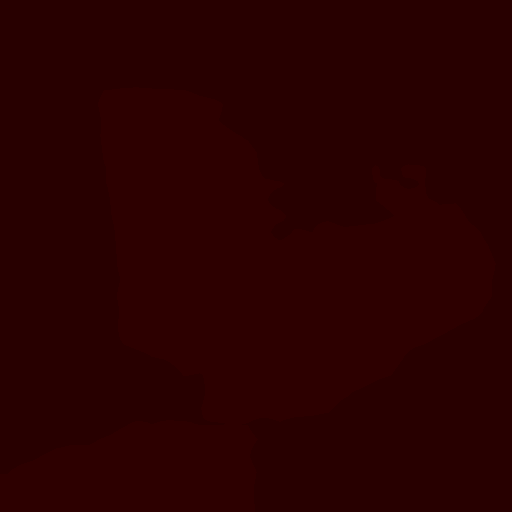} & 
	   \includegraphics[width=0.15\columnwidth]{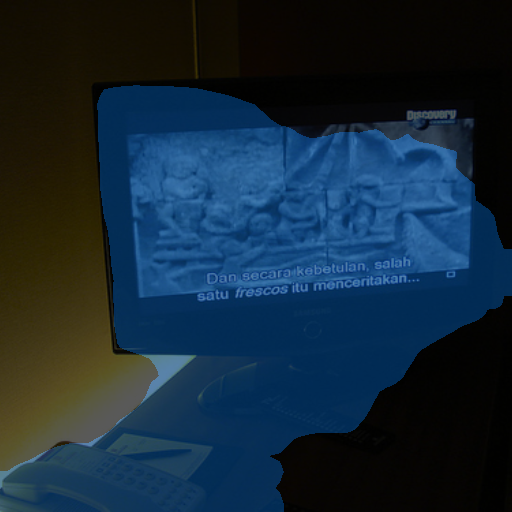} & 
	   \includegraphics[width=0.15\columnwidth]{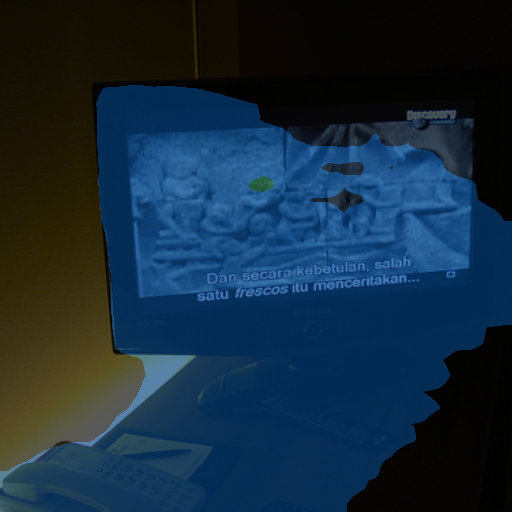} \\
       
           \end{tabular}
    
    \captionof{figure}{\rebuttal{\textbf{Visualizations.} Qualitative figures from  the \textit{multi-step} \textbf{overlap} incremental protocol  on 10-2 VOC. From left to right: 
    input image,   GT  segmentation overlayed, 
    predicted segmentation from old model, 
    semantic similarity map  
    computed between the image  label and  old classes, 
    predicted segmentation  obtained with \ours and with \wil. 
    Semantic similarity maps displayed in OpenCV  colormap HOT
    (low~\includegraphics[width=36pt,height=7pt]{images/viz/colorscale_hot.jpg}~high similarity)}}
    \label{tab:qualitative_visualization2}
\end{figure}

\begin{figure}[h]
    \centering
    \setlength{\tabcolsep}{1.7pt}
    \begin{tabular}{ccccc}
     \small Input & \small GT  & \small $(F\circ E)^{t-1}(\bfx_t)$   
     &  \small \ours   & \small \wil \\

   \includegraphics[width=0.18\columnwidth]{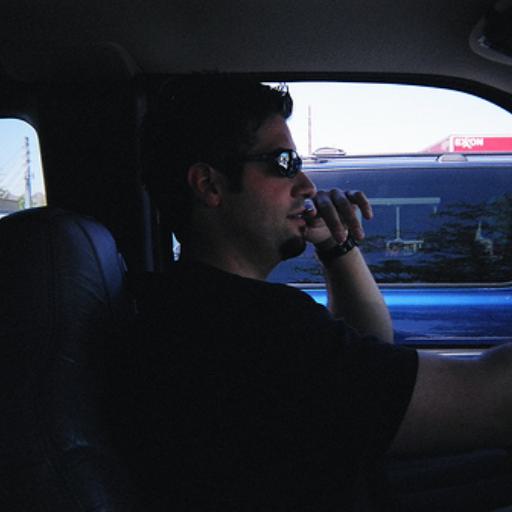} &
       \includegraphics[width=0.18\columnwidth]{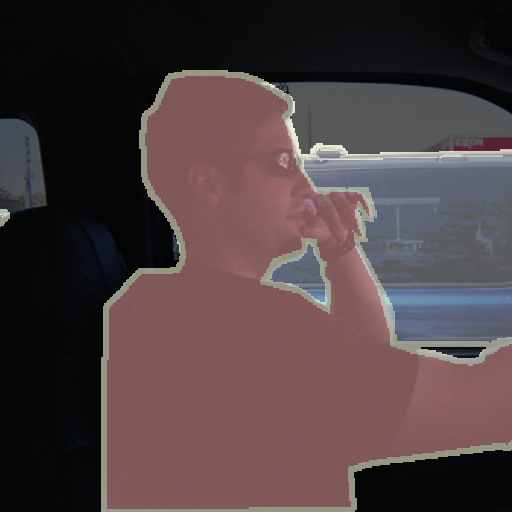} & 
	   \includegraphics[width=0.18\columnwidth]{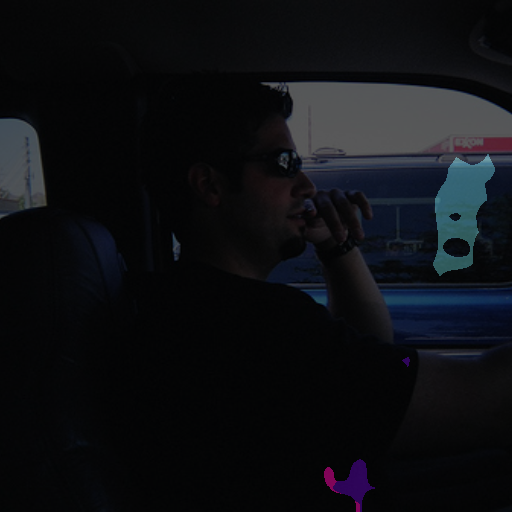} & 
	   \includegraphics[width=0.18\columnwidth]{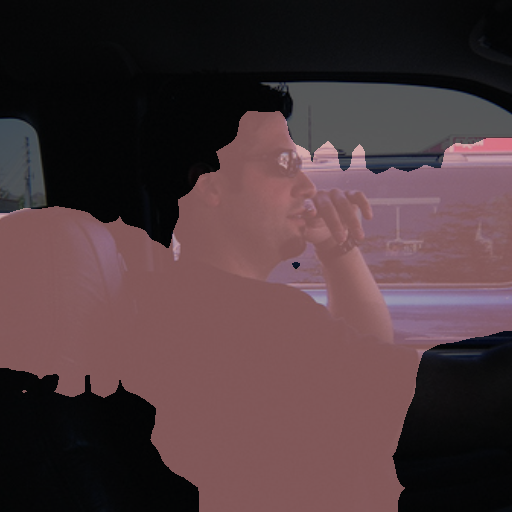} & 
	   \includegraphics[width=0.18\columnwidth]{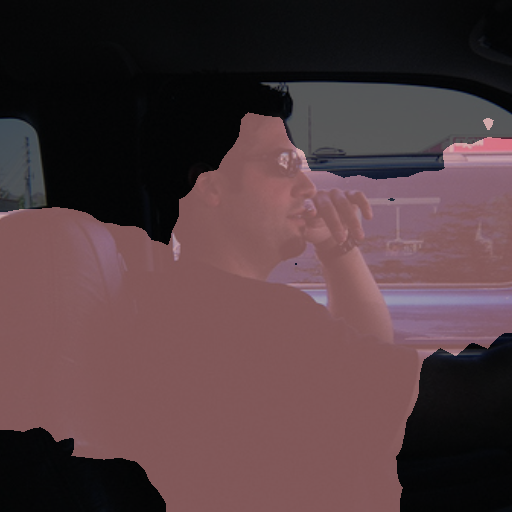} \\
    
    \includegraphics[width=0.18\columnwidth]{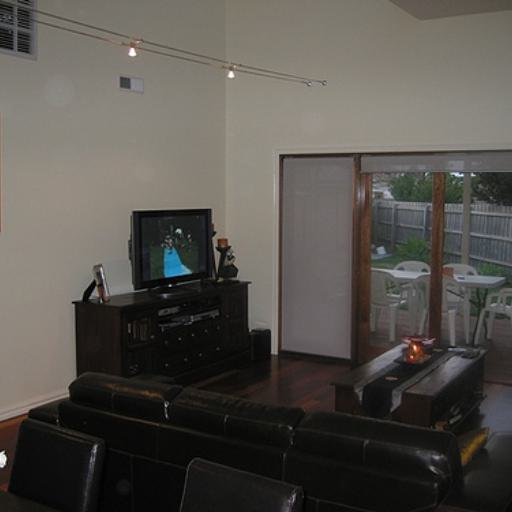} 
    &\includegraphics[width=0.18\columnwidth]{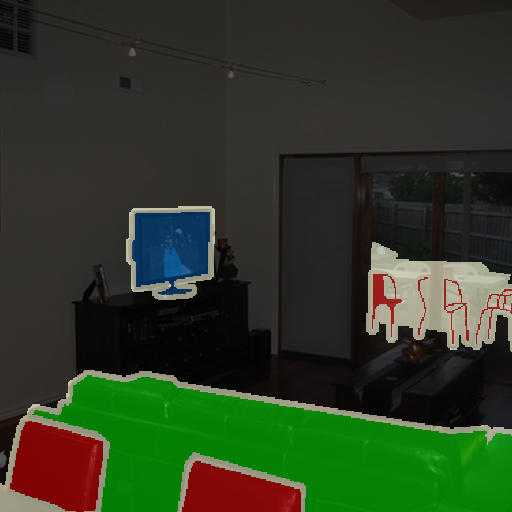} 
	   & \includegraphics[width=0.18\columnwidth]{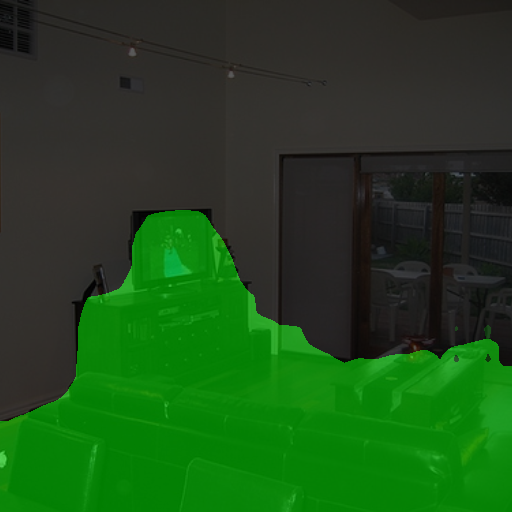} 
	& \includegraphics[width=0.18\columnwidth]{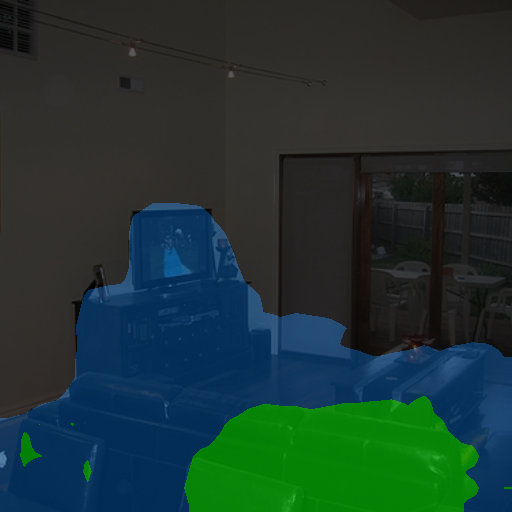} 
	   & \includegraphics[width=0.18\columnwidth]{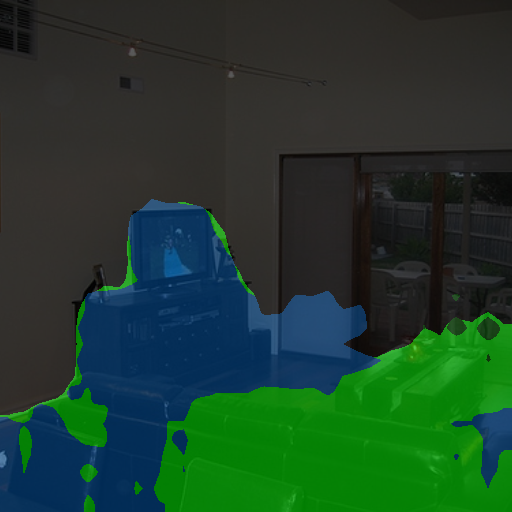} \\

  \includegraphics[width=0.18\columnwidth]{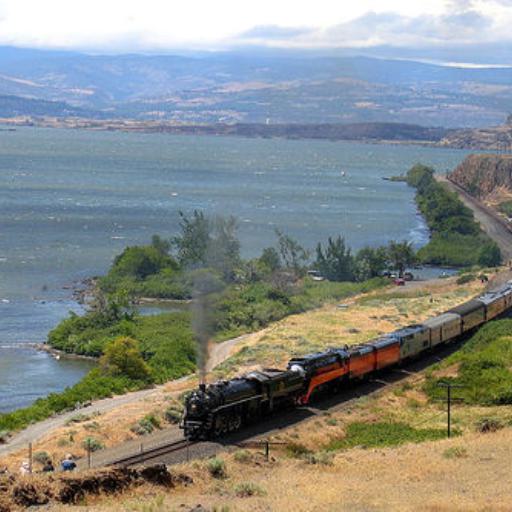} &
       \includegraphics[width=0.18\columnwidth]{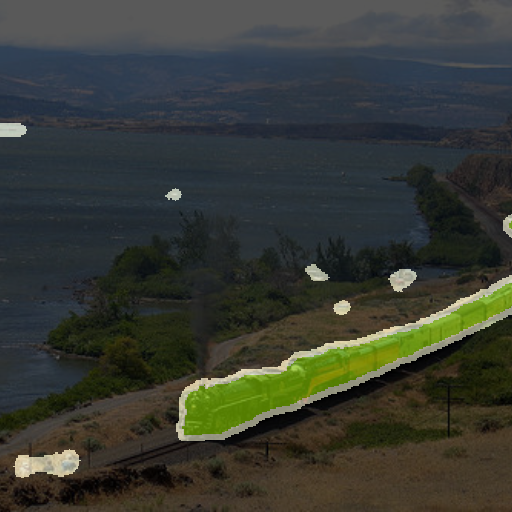} & 
	   \includegraphics[width=0.18\columnwidth]{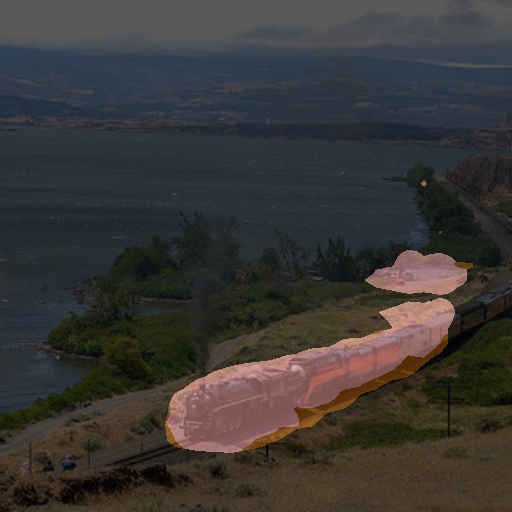} &  
	   \includegraphics[width=0.18\columnwidth]{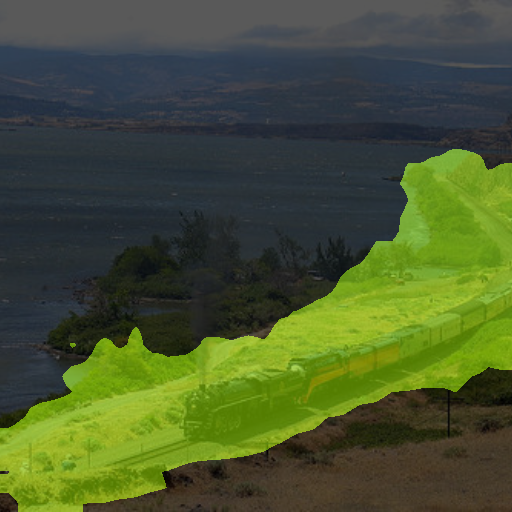} & 
	   \includegraphics[width=0.18\columnwidth]{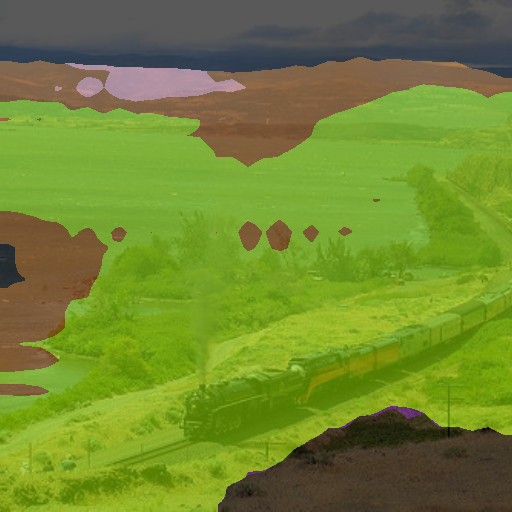} \\
    
        \end{tabular}
	   \captionof{figure}{\rebuttal{\textbf{Failure cases on new classes.} Figures from  the \textit{multi-step} \textbf{overlap} 
	incremental protocol  on 10-2 VOC. From left to right: 
    input image,   GT  segmentation overlayed, 
    predicted segmentation from old model, 
    predicted segmentation  obtained with \ours and with \wil }
    \label{tab:failure_cases_new}}
\end{figure}

\begin{figure}[h]
    \centering
    \setlength{\tabcolsep}{1.7pt}
    \begin{tabular}{ccccc}
     \small Input & \small GT & \small $(F\circ E)^{t-1}(\bfx_t)$   
     &  \small \ours   & \small \wil \\
	   \includegraphics[width=0.18\columnwidth]{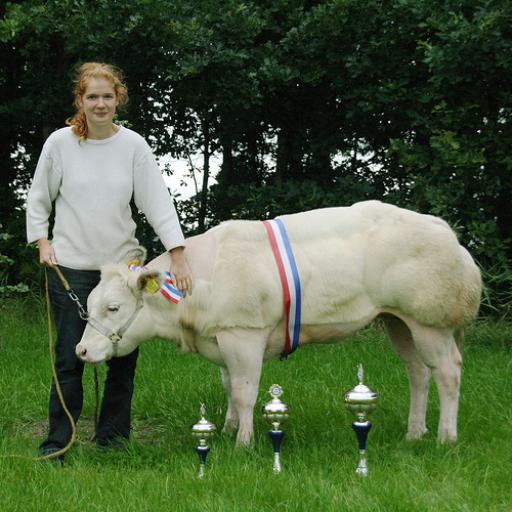} 
	   &\includegraphics[width=0.18\columnwidth]{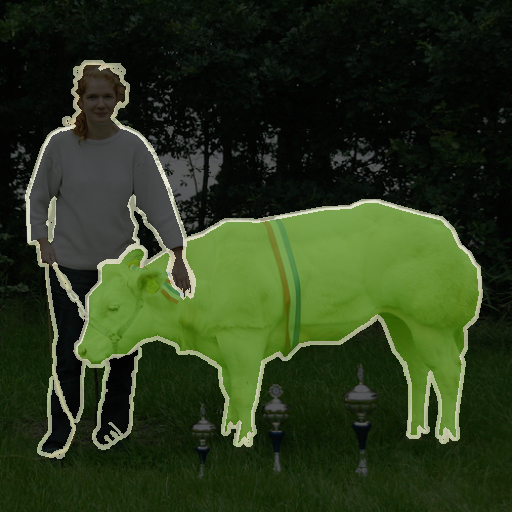} 
	   & \includegraphics[width=0.18\columnwidth]{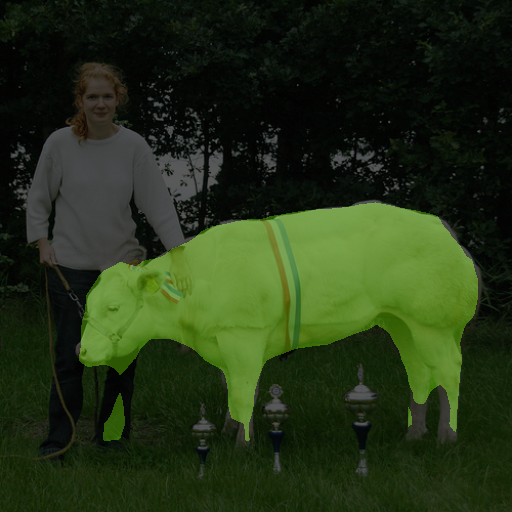} 
	   & \includegraphics[width=0.18\columnwidth]{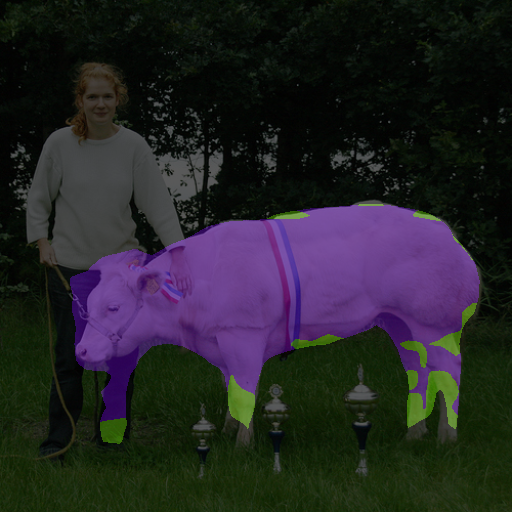} 
	   & \includegraphics[width=0.18\columnwidth]{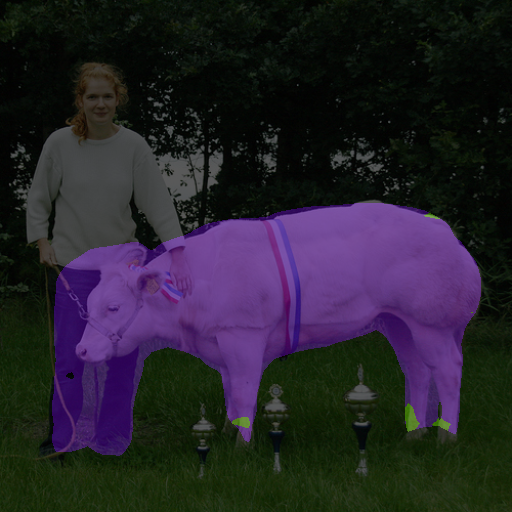} \\

       \includegraphics[width=0.18\columnwidth]{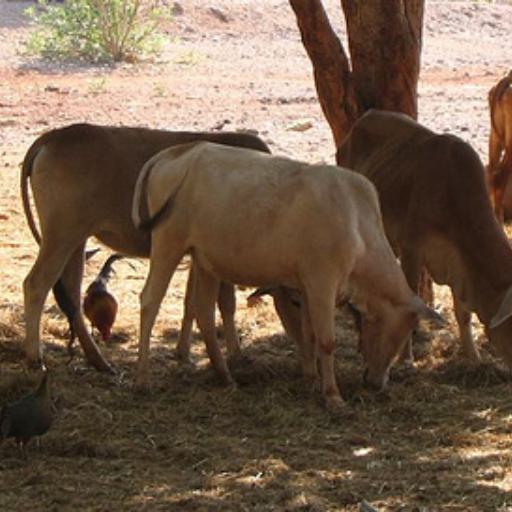} 
	   & \includegraphics[width=0.18\columnwidth]{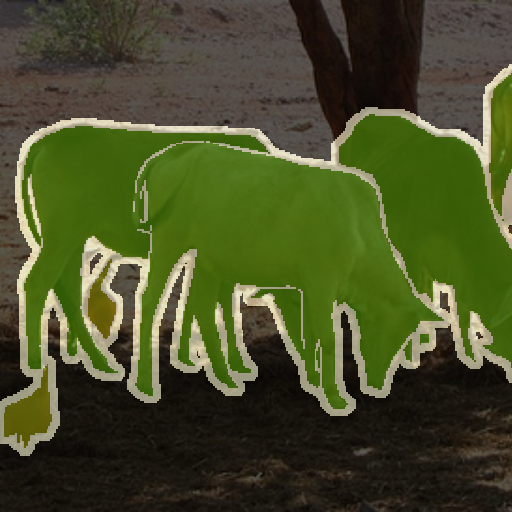} 
	   & \includegraphics[width=0.18\columnwidth]{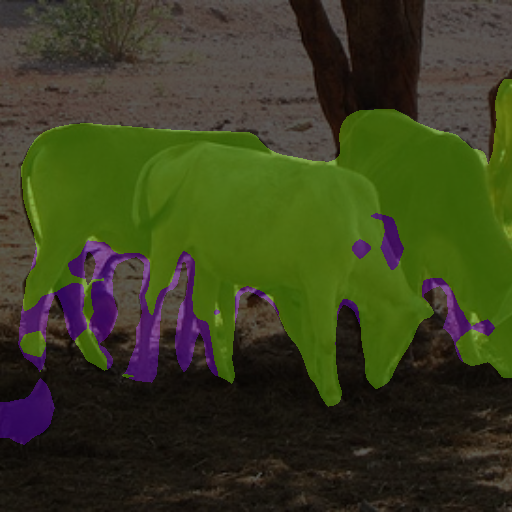} 
	   & \includegraphics[width=0.18\columnwidth]{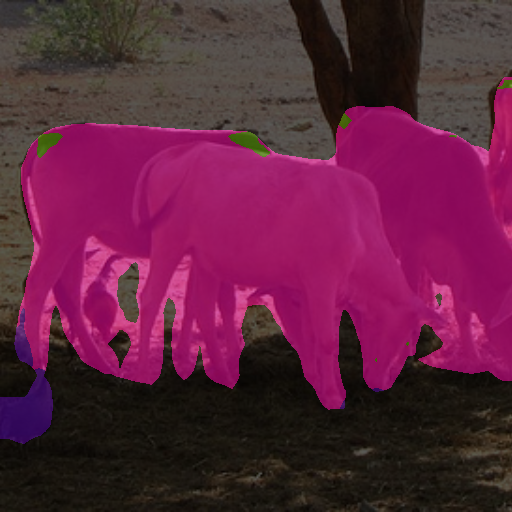} 
	   & \includegraphics[width=0.18\columnwidth]{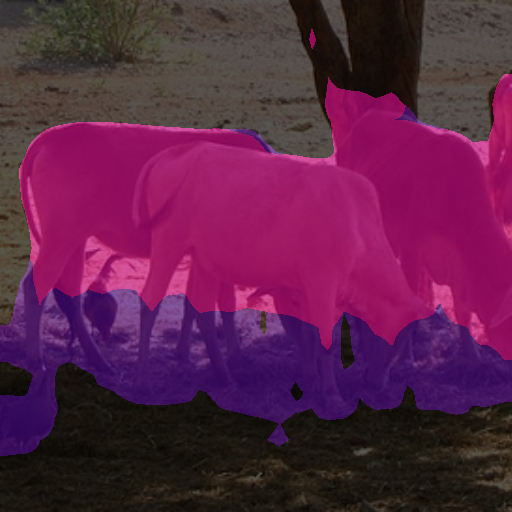} \\

      \includegraphics[width=0.18\columnwidth]{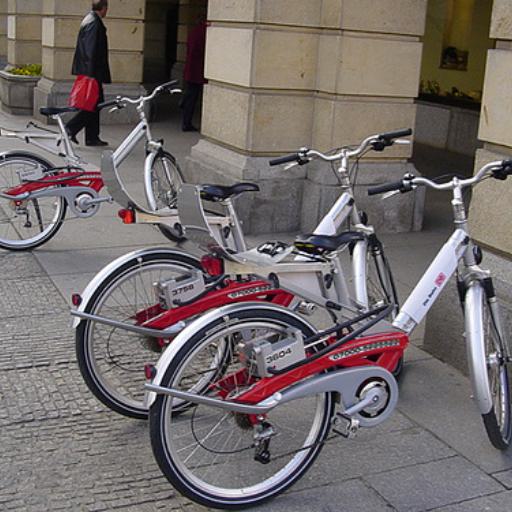} &
       \includegraphics[width=0.18\columnwidth]{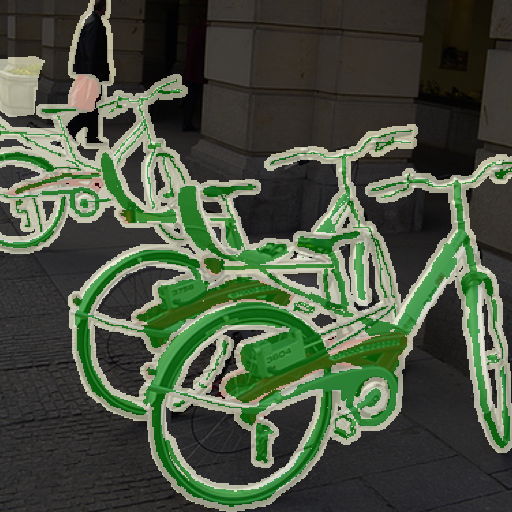} & 
	   \includegraphics[width=0.18\columnwidth]{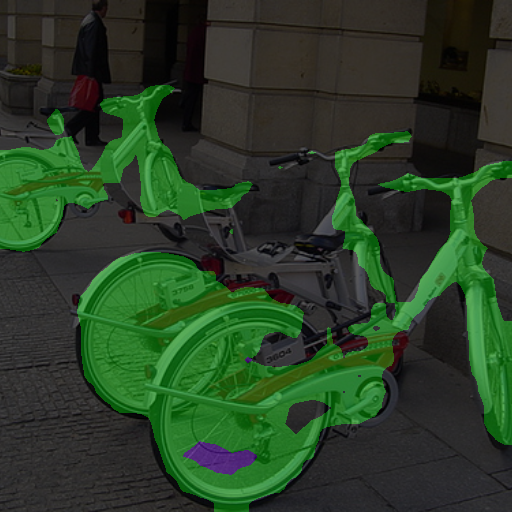} & 
	   \includegraphics[width=0.18\columnwidth]{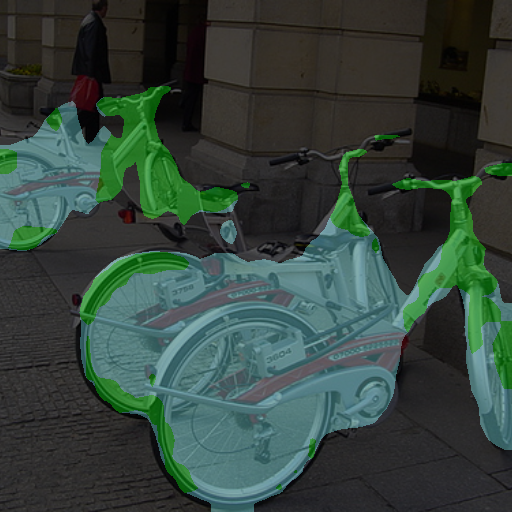} & 
	   \includegraphics[width=0.18\columnwidth]{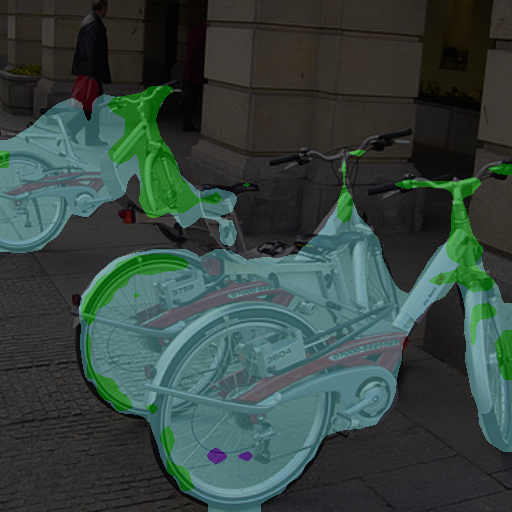} \\

\includegraphics[width=0.18\columnwidth]{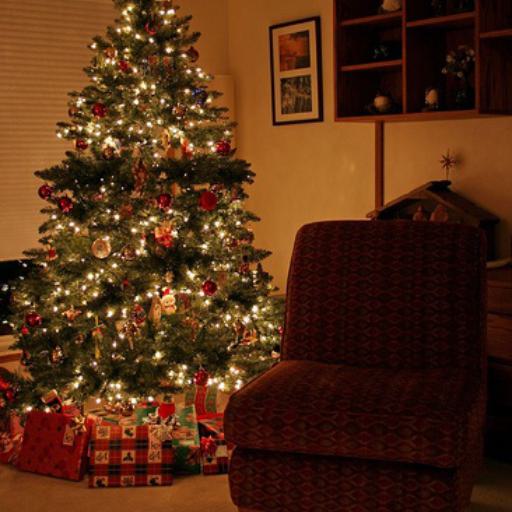} &
       \includegraphics[width=0.18\columnwidth]{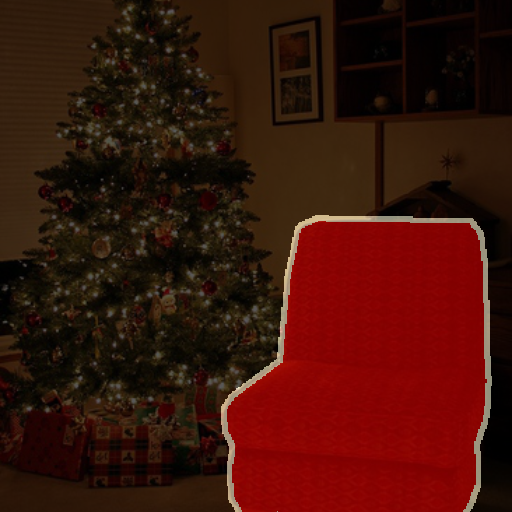} & 
	   \includegraphics[width=0.18\columnwidth]{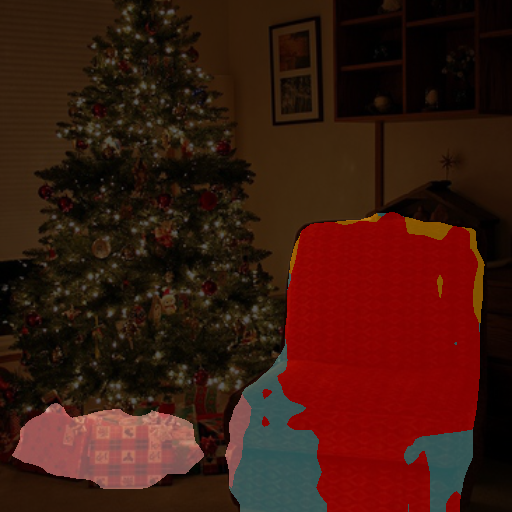} & 
	   \includegraphics[width=0.18\columnwidth]{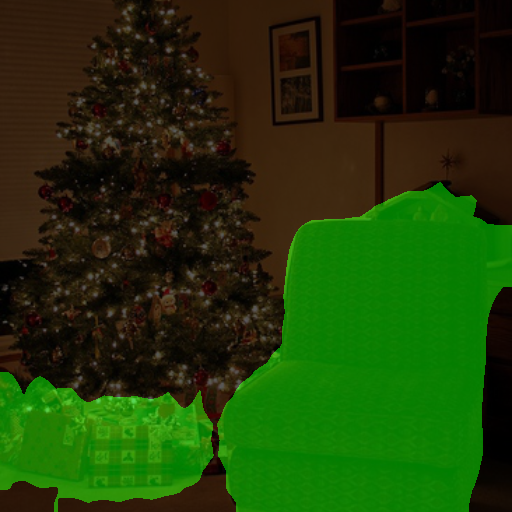} & 
	   \includegraphics[width=0.18\columnwidth]{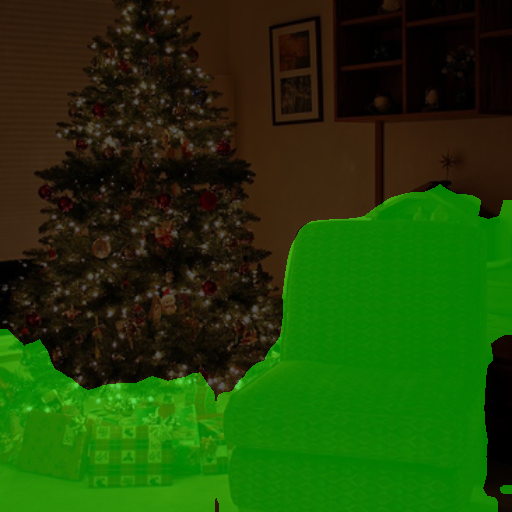} \\
	   
	   \end{tabular}
    
    \captionof{figure}{\rebuttal{\textbf{Failure cases on old classes.} 
    Figures from  the \textit{multi-step} \textbf{overlap} 
	incremental protocol  on 10-2 VOC. From left to right: 
    input image,   GT  segmentation overlayed, 
    predicted segmentation from old model, 
    predicted segmentation  obtained with \ours and with \wil }
    \label{tab:failure_cases_old}}
\end{figure}

We conclude by providing additional qualitative results. In Fig.~\ref{tab:qualitative_visualization_app}, we show further comparison of \ours with \wil on various incremental settings that differ by the number of tasks: 15-5 VOC (2 tasks), 10-5 VOC (3 tasks) and 10-2 VOC (6 tasks). In Fig.~\ref{tab:qualitative_visualization2} we show further examples with the old model prediction and similarity maps between the image label and old classes. 
Finally, in Fig.~\ref{tab:failure_cases_new} we show failure cases for the new class due to lack of semantically similar class, lack of good region detection or low similarity with the predicted class. In Fig.~\ref{tab:failure_cases_old} failure cases are depicted for the old classes where the new class model takes over the old class model (severe forgetting).

From the Fig.~\ref{tab:qualitative_visualization_app} we can see that in the 15-5 VOC setting, the \wil overestimates the ``train'' pixels due to the fact that it uses CAM-like objective under the hood, which suffers from spuriously correlated ``tracks'' in the background -- a general problem among the WSSS methods~\citep{lee2021railroad}. On the other hand, as \ours derives dense pseudo-supervision from previously encountered base class, e.g., ``bus'', which never occurs alongside ``train tracks'', it hinders the CAM-like objective to put mass on the ``train tracks''. This is indeed an interesting property offered by the semantic similarity loss of \ours, which leads to improved segmentation. Similarly, for the other settings we can observe that \ours leads to improved foreground segmentation. Finally, for the harder 10-2 VOC setting, we notice that \wil predicts much of the ``dog'' pixels to be belonging to the class ``tv-monitor'', since the ``tv-monitor'' class is learned in the final task. This happens due to the recency-bias issue described in Sec.~\ref{sec:app-classwise}. While \ours also partially suffers from the same problem, but with lesser severity than \wil.

In the Fig.~\ref{tab:qualitative_visualization2} we provide additional visualizations from the 10-2 VOC setting and highlight the overconfident predictions of the old model on unseen classes. As shown by the $(F \circ  E)^{t-1}({\bf x}_t)$ column in the Fig.~\ref{tab:qualitative_visualization2}, the old model at step $t-1$ predicts the unseen foreground objects to be belonging to the previously learnt classes. This observation is quite contradictory to the conventional knowledge, established in the class-incremental segmentation literature~\citep{CermelliCVPR20ModelingBackgroundIncrementalLearningSemSegm}, that the old model will assign all the unseen classes pixels as the \textit{bkg} due to the \textit{background-shift} issue. As an example, in the first and third rows of the Fig.~\ref{tab:qualitative_visualization2} the old model predicts the ``dog'' and ``horse'' (both unseen) as ``cat'' and ``dog'' (both previously seen), respectively. Our proposed \ours capitalizes on these predictions to obtain denser supervision for free.

Indeed there are also some instances, (see the 5\ths row in the Tab.~\ref{tab:qualitative_visualization2}) where the old model rightfully predicts previously unseen objects (``person'') as the class \textit{bkg}, in-line with background-shift issue. Even in such scenarios, \ours is able to correctly segment the ``person'' object without suppressing the signal from the CAM objective.. In summary, \ours can inherit all the advantages from the \wil framework, and even goes further to help refine its predictions when \wil fails.

Despite the successes shown by \ours, it is far from perfect. We showcase the failure cases on both the new and the old classes in the Tab.~\ref{tab:failure_cases_new} and Tab.~\ref{tab:failure_cases_old}, respectively. In the Tab.~\ref{tab:failure_cases_new} we can observe that both \wil and \ours fail to satisfactorily segment the weakly-labelled new classes. Given the old model predictions are either not present or insufficient, the proposed \ours loss can not guide the model to the right regions of the foreground. Simultaneously, we also observe failure on the old classes, which are demonstrated in the Tab.~\ref{tab:failure_cases_old}. We can see that the base classes ``cow'', ``bicycle'' and ``chair'', etc are mostly segmented as the newly learnt classes, both by \wil and \ours, despite the old model correctly segmenting them. Given that we use the pseudo-labels supervision from the localizer to re-train the main segmentation head, it wipes away previously learned information about the old classes. Note that this phenomenon is not introduced by the \ours loss, and is rather caused due to the pseudo-labelling loss of \wil, as described in \cref{eqn:wilson-final} of the main paper.

\section{Discussion}
\label{sec:app-discuss}

In this section we discuss some of the edge-cases where our proposed RaSP may fail to provide clean pseudo-labels for supervision. In particular, we discuss about two of such cases where ambiguity in pseudo-supervision may arise.

In the \textit{first} case we can imagine a scenario, where the new class (\textit{e.g.}, ``sheep'') co-occurs with an old class (\textit{e.g.}, ``cow'') in the current task image. Due to strong visual similarity, the old model will predict ``sheep'' pixels as ``cow'', whereas the localizer will predict ``sheep'' pixels correctly. Such conflict is introduced by WILSON's design because it needs to make a decision for the pseudo-label of a pixel given the predictions of both the old model and the localizer (see Eq. (7) in~\citep{CermelliCVPR22IncrementalLearninginSemSegmfromImageLabels}). The assumption made in WILSON is that the localizer will predict new classes with far higher confidence than the old model. We do not introduce any additional ambiguity for this given use-case because our proposed RaSP loss creates the semantic similarity maps only for the new classes (``sheep'' in this case) and not for the old class ``cow'' (note the subscript in \cref{eqn:semantic-loss}, where $c \in \mathcal{C}_t$, i.e., new classes). In practice, we observe that the old and new classes co-occurring in the new task images do not happen quite often. If such co-occurrences happen only a few times then the model is able to handle this ambiguity.

In the \textit{second} case, with the introduction of several new classes at once, there is a possibility of confusion in the semantic similarity maps. In detail, the ambiguity will specifically arise when there are more than one new class (\textit{e.g.}, ``horse'' and ``sheep'') in a given new image that has strong visual similarity with an old class (\textit{e.g.}, ``cow''). To recap, we have access to image-level labels only for the new classes, and the old model $(E \circ F)^{t-1}$ predicts the ``horse'' and ``sheep'' pixels as ``cow'' owing to strong visual similarity among them. As a result, the estimated semantic similarity maps for the ``horse'' and ``sheep'' channels will be simultaneously high for the pixel locations where these two objects are present. This is not ideal because it can drive the model to misclassify ``horse'' as ``sheep'' and vice versa. However, such co-occurences do not happen quite often, which is evident from only a small drop in performance (-0.7\%) in the 60-20 COCO-to-VOC. With enough data the model will eventually learn to ignore these noisy pseudo-labels coming from a small fraction of images.

As a future work we plan to reduce the ambiguities introduced by WILSON and RaSP, and as a consequence provide cleaner supervisory signal to the model.

\end{document}